\documentclass[twoside,11pt]{article}

\usepackage{blindtext}

\usepackage{jmlr2e}

\usepackage{bm} 
\usepackage{bbold} 

\usepackage{mario-def}

\usepackage{dsfont} 
\usepackage{amsmath}
\usepackage{graphicx}      
\usepackage[makeroom]{cancel}
\usepackage{float}         
\usepackage{caption} 
\usepackage{subcaption}    

\usepackage{booktabs}
\usepackage{multirow}
\usepackage{siunitx}

\usepackage{color}

\usepackage{lastpage}
\jmlrheading{23}{2022}{1-\pageref{LastPage}}{1/21; Revised 5/22}{9/22}{21-0000}{Author One and Author Two}

\ShortHeadings{Mixtures of Transparent Local Models}{Diaby and Duchesne and Marchand}
\firstpageno{1}

\begin{document}

\title{Mixtures of Transparent Local Models}

\author{\name Niffa Cheick Oumar Diaby \email niffa-cheick-oumar.diaby.1@ulaval.ca \\
       \addr Department of Computer Science and Software Engineering\\
       Laval University\\
       Québec, QC G1V0A6, CANADA
       \AND
       \name Thierry Duchesne \email thierry.duchesne@mat.ulaval.ca \\
       \addr Department of Mathematics and Statistics\\
       Laval University\\
       Québec, QC G1V0A6, CANADA
       \AND
       \name Mario Marchand \email mario.marchand@ift.ulaval.ca \\
       \addr Department of Computer Science and Software Engineering\\
       Laval University\\
       Québec, QC G1V0A6, CANADA}

\editor{My editor}

\maketitle

\begin{abstract}

The predominance of machine learning models in many spheres of human activity has led to a growing demand for their transparency. The transparency of models makes it possible to discern some factors, such as security or non-discrimination. In this paper, we propose a \textit{mixture of transparent local models} as an alternative solution for designing interpretable (or transparent) models. Our approach is designed for the situations where a simple and transparent function is suitable for modeling the label of instances in some localities/regions of the input space, but may change abruptly as we move from one locality to another. Consequently, the proposed algorithm is to learn both the transparent labeling function and the locality of the input space where the labeling function achieves a small risk in its assigned locality. By using a new multi-predictor (and multi-locality) loss function, we established rigorous PAC-Bayesian risk bounds for the case of binary linear classification problem and that of linear regression. In both cases, synthetic data sets were used to illustrate how the learning algorithms work. The results obtained from real data sets highlight the competitiveness of our approach compared to other existing methods as well as certain opaque models.
\end{abstract}


\section{Introduction} 
Interpretable machine learning models are important for fairness testing, causality assessment and reliability or robustness screening. This is especially true in high-stakes sectors such as finance, transportation and health care. Consequently, several state laws have emerged to regulate the use of machine learning models and enforce some level of interpretability and transparancy such as Law $25$ of the Province of Quebec~\citep{LOI_25_QC_2021}, Article $22$ of the General Data Protection Regulation (GDPR) of the European Union~\citep{EU_GDPR_2016}, and other similar laws in the United States~\citep{US_GDPR_2018}.

Following these growing attentions, two main strategies have been developed to address transparency and explainability issues.

The first is the use of intrinsic interpretable model design techniques. Intrinsic interpretable models are prediction models whose decision-making process is understandable by humans without the need of any additional tool or algorithm. Example of models that belong to this family are generalized linear models (GLMs) (linear regression, log-linear model, etc.)~\citep{nelder1972generalized, mccullagh1989generalized}, generalised additive models (GAMs)~\citep{hastie1990generalized}, decision trees~\citep{breiman1984classification, quinlan1986induction}, case-based reasoning~\citep{kolodner1992introduction, smyth2007case}, rule sets~\citep{mccormick2011hierarchical, yang2022truly} and rule lists~\citep{angelino2018learning, pellegrina2024scalable}. However, while the way in which transparent models make decisions is clear and fully understood by humans, the complexity of these types of models is usually restricted to improve their interpretability.

The second solution is to resort to methods for explaining complex models, also called black box models. These methods usually rely on transparent models to explain locally or globally non-transparent models. They include, in particular, instruments such as TREPAN \citep{craven1995extracting}, LIME \citep{ribeiro2016should}, SHAP \citep{lundberg2017unified}, BETA \citep{lakkaraju2017interpretable}, Anchors \citep{ribeiro2018anchors} or MUSE \citep{lakkaraju2019faithful}. Unfortunately, the explanations of these models in post-hoc fashion have proved to be unreliable and highly manipulatable \citep{aivodji2019fairwashing, dimanov2020you, aivodji2021characterizing}.

In this paper, we therefore propose to focus on alternative solutions for designing interpretable models. Our solution is based on the use of multiple model techniques to design a \textit{mixture of transparent local models}. Succinctly, a mixture of transparent local models divides the problem space into several subspaces in order to learn interpretable predictors (transparent local models) in the different regions. Instead of designing a single complex model, which is potentially difficult to interpret, it might be more interesting to build simple and interpretable local models around certain points of interest. A point of interest can be an instance for which we want to predict the label based on similar instances present in the training samples. 

Before moving into the theory of \textit{mixture of transparent local models}, let us highlight why this approach could be better than learning a complex global model. Let $\Dcal$ be some unknown data-generating distribution over $\Xcal \times \Ycal$, where $\Xcal$ denotes the input space and $\Ycal$ the label space. Similarly, $\Dcal^{'}$ is a data-generating distribution over $\Xcal^{'} \times \Ycal^{'}$, where $\Xcal^{'}$ and $ \Ycal^{'}$ designate, respectively, the input space and the label space. Consider two sets of training samples $S \eqdef \{(\xb_{i}, y_{i})\}_{i=1}^{m}$ and $S^{'} \eqdef \{(\xb_{i}^{'}, y_{i}^{'})\}_{i=1}^{m}$ of size $m$, where each example $(\xb_{i}, y_{i}) \in \Xcal \times \Ycal$ and $(\xb_{i}^{'}, y_{i}^{'}) \in \Xcal^{'} \times \Ycal^{'}$. Suppose we would like to build a predictive model on each data set $S$ and $S^{'}$, having access to a class of polynomial functions $\mathcal{H}_{k}$ of order $k$. 

\begin{figure}[H]
   \begin{subfigure}[b]{0.48\textwidth}
        \includegraphics[scale=0.115]{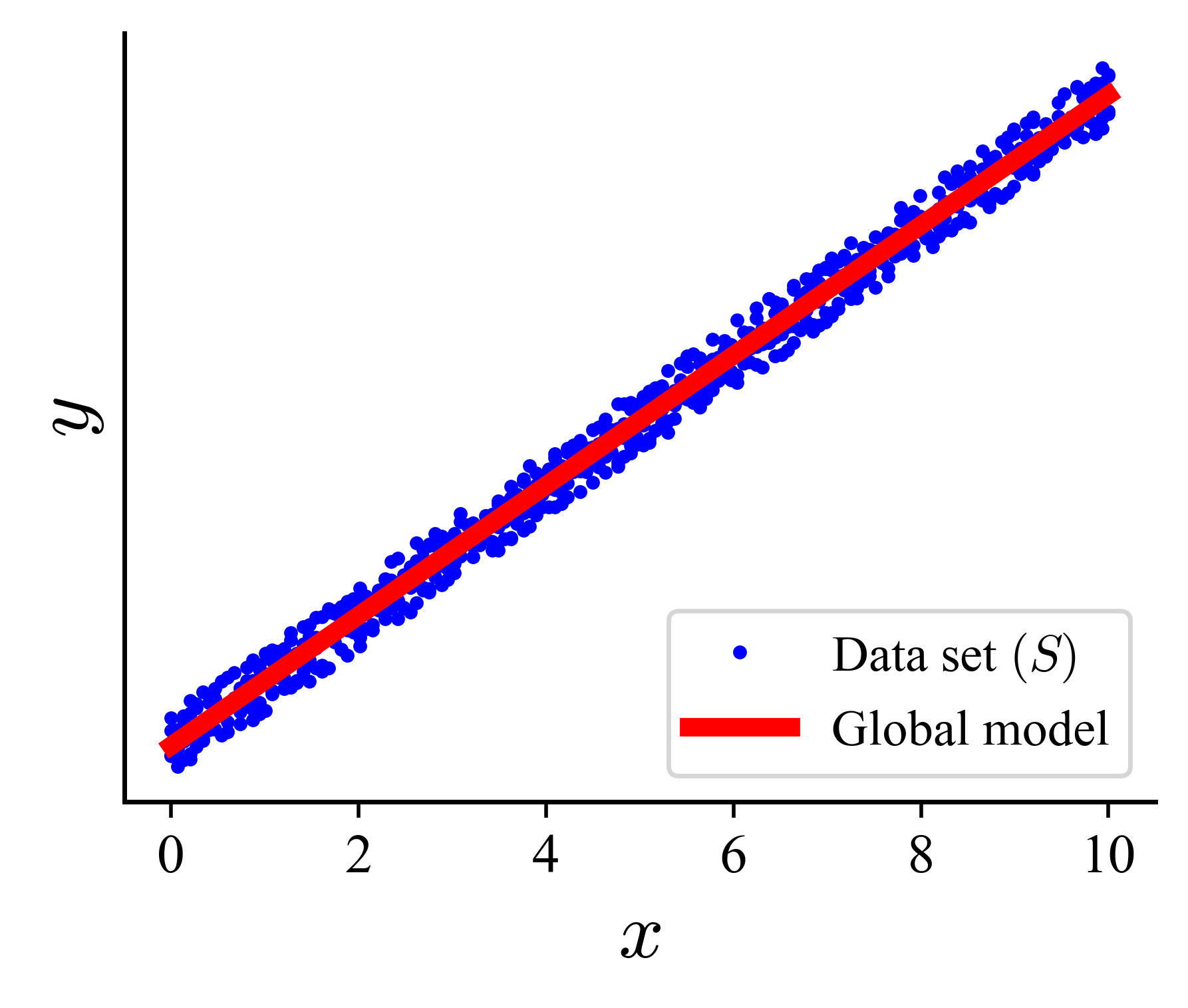}
        \caption{Simple global model.}
    \end{subfigure}
    \hfill
    \begin{subfigure}[b]{0.48\textwidth}
        \includegraphics[scale=0.115]{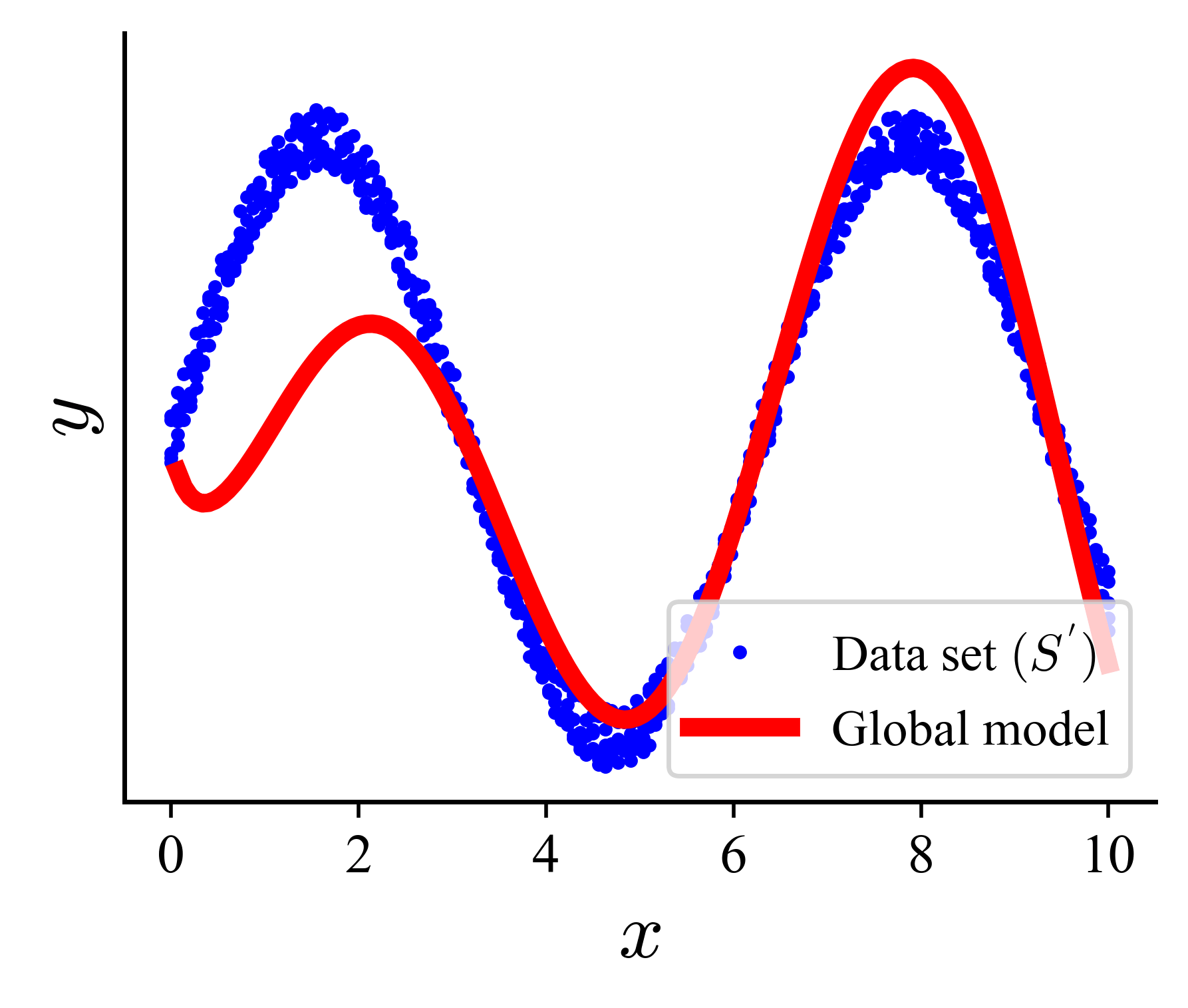}
        \caption{Complex global model.}
    \end{subfigure}
    \caption{Example of global models.}
    \label{fig:motivation}
\end{figure}

The global model learned from the data set $S$, is simpler and has an acceptable level of performance. However, the global model learned from the data set $S^{'}$ is much more complex, requiring a polynomial of high order $k$. Thus, Figure \ref{fig:motivation} illustrates that, \textit{on certain tasks, learning a global model might force it to be very complex and therefore difficult to interpret}.

\begin{figure}[H]
    \centering
    \includegraphics[scale=0.115]{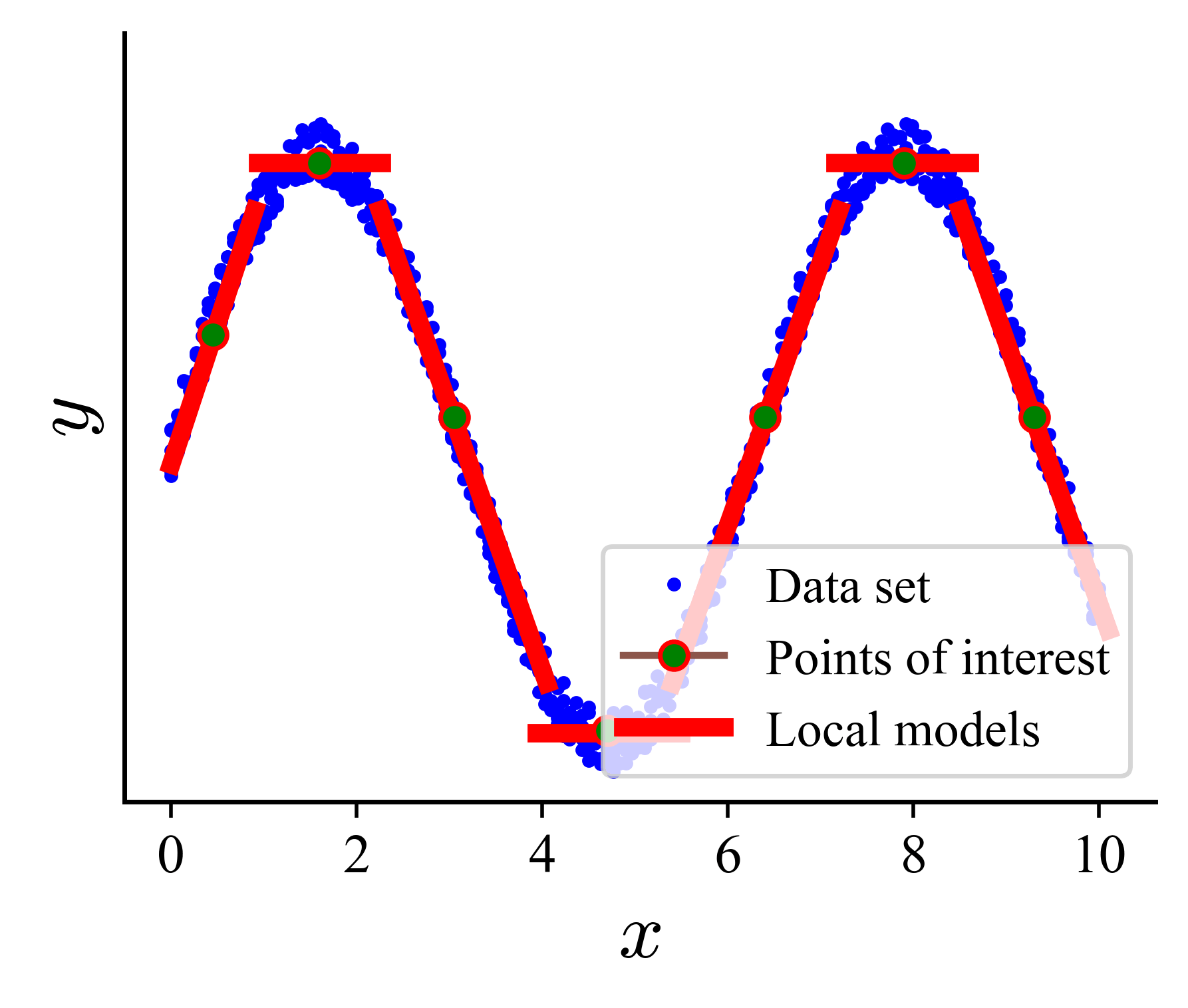}
    \caption{Example of mixture of transparent local models.}
    \label{fig:motivation_end}
\end{figure}

Now, let us reformulate the predictive modelling problem on the data set $S {'}$ as seven sub-problems. Each sub-problem consists of learning a simple transparent local model on a part of the data set situated in a specific locality, particularly around the vicinity of a point of interest, as shown in Figure \ref{fig:motivation_end}. 

In this case, we find for each sub problem, that a reasonable level of performance could be achieved with a polynomial of degree $0$ or $1$. Therefore, by combining these seven simple local models, we could gain an additional gain on the performance level assessed across all examples in the data set $S {'}$. In other words, \textit{a low risk could be achieved with the use of local models that can be interpreted}.

\textbf{Related Work:} 
Multiple model techniques in interpretable machine learning involve combining two or more models to increase performance gain or interpretability. Here are some works based on this approach.

\citet{wang2015trading} proposed Oblique Treed Sparse Additive Models (OT-SpAMs) which divide the input space into regions with sparse oblique tree splitting and assign local sparse additive experts to individual regions. Other works are Hybrid Interpretable Models \citep{pan2020interpretable, wang2021hybrid, ferry2023learning}, which are systems that involve the cooperation of a transparent model and a complex black box. At inference time, any input of the hybrid model is assigned to either its interpretable or complex component based on a gating mechanism. \citet{vogel2024meta} proposed The League of Experts. They adapt a particular case of ensemble learning, in which each ensemble member is associated with a specific part of the input space. More specifically, they introduce an algorithm that adaptively learns a partition of the input space in order to assign one specialized ensemble member—the expert—to each subset of the partition. 

Our work differs from \citet{wang2015trading} in that the division of the input space is carried out around certain points of interest instead of using decision trees or some other mechanism of partitionning the input space without overlap. It is clear that our proposal is totally different from Hybrid Interpretable Models, since it doesn't necessitate the collaboration of a transparent model and a complex black box. As for \citet{vogel2024meta}, they divide the input space around some points of interest without overlaps. In contrast we do not use a partition and tolerate overlaps. Finally, we also study the case where the points of interest are given by the user (such as the typical clients of an insurance company) as well as the case where the points of interest must be learned. 

\section{PAC-Bayesian Machine Learning}
In this section, we present the PAC-Bayesian setting that provides guarantees (in the form of risk bounds) and the associated learning algorithm that optimizes these risk bounds and that will be used subsequently to construct mixtures of transparent local models.

\subsection{Basic Notions of Machine Learning}
Consider a data-generating distribution $\Dcal$ on $\Xcal\times\Ycal$ where $\Xcal$ is the instance (or input) space and where $\Ycal$ is the label space. We assume that each training and testing example $(\xb,y)\in \Xcal\times\Ycal$ is generated according to $\Dcal$ so that a training sample $S \eqdef \{(\xb_{i}, y_{i})\}_{i=1}^{m}$ of $m$ examples is drawn according to $\Dcal^{m}$. We also consider learning algorithms that select a predictor among a class $\Hcal$ of available functions $h: \Xcal\to\Ycal$. Moreover, we are given a loss function $\ell: \Ycal\times\Ycal \to \Reals_+$ such that $\ell(h(\xb),y)$ gives the loss incurred by $h$ on example $(\xb, y)$. Given a training sample $S$, the goal of the learning algorithm is to find a predictor $h \in \mathcal{H}$ that minimizes the true risk $L_\Dcal(h)$ defined as the expected loss according to $\Dcal$:
\[
    L_{\Dcal}(h) \eqdef \underset{(\xb, y) \sim \Dcal}{\EE} \LB \ell(h(\xb), y) \RB\, \textrm{.}
\]
This quantity is usually impossible to calculate since $\Dcal$ is unknown. So, instead, the learning algorithm uses the empirical risk $L_S(h)$ of $h$ evaluated on $S$, defined by
\[
    L_{S}(h) \eqdef \frac{1}{m} \sum_{i=1}^{m} \ell(h(\xb_{i}), y_{i})\, \textrm{.}
\]
Note that in this paper, the input space $\Xcal$ represents the feature space.

\subsection{PAC-Bayesian Risk Bounds}
The PAC-Bayes approach, introduced by \citet{mcallester1998some}, is a tool to produce probably approximately correct (PAC) guarantees for learning algorithms in the form of tight risk bounds. In this framework, given a loss function $\ell$ and a class $\Hcal$ of predictors, a learning algorithm $A_{\Hcal}$ receives, as input, a sample $S$ of $m$ examples and a \textit{prior} distribution $P$ on $\Hcal$ and returns a \textit{posterior} distribution $Q$ on $\Hcal$.

The effectiveness of training is captured by the accuracy of the predictors $h$ drawn according to $Q$. More formally, the loss incurred over an example $(\xb, y)$ is defined as  
\[
    \ell(Q, (\xb, y)) \eqdef \underset{h \sim Q}{\EE} \LB \ell(h(\xb), y) \RB\, \textrm{.}
\]
The true risk is meanwhile expressed by
\[
    L_{\Dcal}(Q) \eqdef \underset{(\xb, y) \sim \Dcal}{\EE} \LB \ell(Q, (\xb, y)) \RB = \underset{h \sim Q}{\EE} L_{\Dcal}(h)\, \textrm{.}
\]
The empirical risk evaluated on the sample $S$ is defined as
\[
    L_{S}(Q) \eqdef \frac{1}{m} \sum_{i=1}^{m} \ell(Q, (\xb_{i}, y_{i})) = \underset{h \sim Q}{\EE} L_{S}(h)\, \textrm{.}
\]
As the PAC-Bayes framework returns a stochastic predictor called the \textit{Gibbs predictor}, the risks $L_{\Dcal}(Q)$ and $L_{S}(Q)$ are often called, respectively, the true risk of Gibbs and the empirical risk of Gibbs. The PAC generalization bounds established via PAC-Bayes approach depend on the Kullback-Leibler divergence between the \textit{posterior} $Q$ and the \textit{prior} $P$.

\begin{definition}\label{def:kl}
    The Kullback-Leibler (KL) divergence is a measure of dissimilarity between two probability distributions. Let $p(h)$ and $q(h)$ be the probability densities of $P$ and $Q$ at point $h \in \mathcal{H}$. Note that $Q$ is absolutely continuous with respect to $P$, denoted by $Q \ll P$ if and only if 
    \[
        \forall h \in \Hcal: p(h) = 0 \implies q(h) = 0\, \textrm{.}
    \]
    For any distributions $Q$ and $P$ on $\Hcal$, the KL divergence between $Q$ and $P$, denoted \newline $KL(Q||P)$, is defined as
    \[
        KL(Q||P) \eqdef
        \begin{cases}
          \underset{h \sim Q}{\EE} \ln \LB \frac{q(h)}{p(h)} \RB & \text{if $Q \ll P$}\\
          \infty & \text{otherwise.}
        \end{cases}
    \]
\end{definition}
The following theorem can be considered as a ``core" of the PAC-Bayes approach, since many PAC-Bayes generalization bounds can be retrieved through this theorem. 

\begin{theorem}{\rm\citep{MM-LN-24}}
    \label{general_theorem}
    Consider some distribution $\Dcal$ over the space $\Xcal \times \Ycal$, a class of predictors $\Hcal: \Xcal \rightarrow \Ycal$, and a prior distribution $P$ on $\Hcal$. Let $\phi : \Reals_{+} \rightarrow \Reals_{+}$, and $f : \Hcal \times (\Xcal \times \Ycal)^{m} \rightarrow \Reals$ be mesurable functions satisfaying 
    \begin{equation}
        \forall \lambda > 0, \forall h \in \Hcal : \; \underset{S \sim \Dcal^{m}}{\EE} e^{\lambda f(h, S)} \leq \phi(\lambda)\, \textrm{.}
    \end{equation}
    Then, for any $\lambda > 0$ and for any $\delta \in (0, 1)$, with probability at least $1 - \delta$ over the draws of $S \sim \Dcal^{m}$, we have
    \begin{equation}
        \forall Q \ll P:\ \underset{h \sim Q}{\EE} f(h, S) \leq \frac{1}{\lambda}\LB KL(Q||P) + \ln \frac{\phi(\lambda)}{\delta} \RB\, \textrm{.}
    \end{equation}
\end{theorem}

As a consequence of this theorem, if we choose $f(h, S) = L_{\Dcal}(h) - L_{S}(h)$, the previous upper bound on $\underset{h \sim Q}{\mathds{E}} f(h, S)$ automatically gives an upper bound on $L_{\Dcal}(Q) - L_{S}(Q)$. The same observation applies to  $f(h, S) =  L_{S}(h) - L_{\Dcal}(h)$. 

Throughout this paper, we will assume the loss function $\ell$ is $\sg$-sub Gaussian, which implies that we have 
\begin{equation}
    \forall h \in \Hcal, \forall \lambda\in\Reals, \; \underset{S \sim \Dcal^{m}}{\EE} e^{\lambda \LB L_{\Dcal}(h) - L_{S}(h)\RB} \leq e^{\frac{\lambda^{2} \sg^{2}}{2m}} = \phi(\lambda)\, \textrm{.}
\end{equation}

By using the union bound on the upper bounds of $L_{\Dcal}(Q) - L_{S}(Q)$ and $L_{S}(Q) - L_{\Dcal}(Q)$, we obtain the following corollary, which is similar to a result of \citet{NIPS2016_84d2004b}.

\begin{corollary}
    \label{absolute_bound}
    For any distribution $\Dcal$ on $\Xcal \times \Ycal$, any distribution $P$ on $\Hcal$, any $\sg$-sub Gaussian loss function $\ell$, any $\lambda > 0$, and any $\delta \in (0, 1)$, with probability at least $1 - \delta$ over the draws of $S \sim \Dcal^{m}$, we have
    \[
        \forall Q \ll P:\ |L_{\Dcal}(Q) - L_{S}(Q)| \leq \frac{1}{\lambda}\LB KL(Q||P) + \ln \frac{2}{\delta} \RB + \frac{\lambda \sg^{2}}{2m}\, \textrm{.}
    \]
\end{corollary}

\subsection{The PAC-Bayesian Learning Algorithm}
According to the Corollary \ref{absolute_bound}, given a training set $S$, a \textit{prior} distribution $P$ on a class $\Hcal$ of predictors, and $\lambda > 0$, the learning algorithm that optimizes the PAC-Bayesian guarantee on $L_{\Dcal}(Q)$ is given by 
\[
    \mathbf{PB}_{\Hcal}^{\lambda}(S) \eqdef \underset{Q \ll P}{\mathrm{argmin}} 
    \Biggl[ L_{S}(Q) + \frac{1}{\lambda} KL(Q||P) \Biggr]\, \textrm{.}
\]
Thus, the learning algorithm $\mathbf{PB}_{\Hcal}^{\lambda}(S)$ aims to minimize the empirical risk $L_{S}(Q)$ and the complexity term $\frac{1}{\lambda} KL(Q||P)$. In the following, we will propose such learning algorithms for parametrized \textit{posterior} distributions on mixtures of interpretable local models. 

\section{Basic Elements for Mixtures of Transparent Local Models}
Our approach is based on the fact that the label of instances that are close to some instance or point of interest should be predictable by a simple model. Consequently, we will try to train simple models only on the instances that are within some distance to a reference point. For this task, we will make use of a vicinity function which assigns a value of $1$ to the instances that are within some distance to a reference point and $0$ to all other instances. Hence, the simple models will be trained on instances located within a ball with respect to a metric which should be compatible with our notion of similarity or dissimilarity among pair of instances. This is to be contrasted with other approaches that try to find a partition of the input space, such as the one obtained by a decision tree or a tiling of Voronoïd cells, which are not balls defined with respect to some metric compatible with a notion of similarity between instances.

\subsection{Vicinity Function}
Given some metric $d: \Xcal\times\Xcal \to \Reals_+$, the measure of vicinity between a point of interest $\cb$ and some given instance $\xb$ is captured by the vicinity function $K(\cb, \xb, \be)$ defined as
\begin{equation}\label{eq:hard_treshold_function}
  K(\cb, \xb, \be) \eqdef
    \begin{cases}
      1 & \text{if $d(\cb, \xb) \leq \be$}\\
      0 & \text{otherwise.}
    \end{cases}       
\end{equation}
where $\be \in \Reals_{+}$ denotes the locality parameter. Then, the complement of $K(\cb, \xb, \be)$, denoted $\bar{K}(\cb, \xb, \be)$, is defined as  
\begin{equation}\label{eq:complement_of_hard_treshold_function}
    \bar{K}(\cb, \xb, \be) \eqdef 1 - K(\cb, \xb, \be)\, \textrm{.}
\end{equation}
As is often the case, we consider that the locality parameter $\beta$ belongs to a set $\Tilde{\Bcal} \subseteq \mathds{R}_{+}$. Thus, the set $\Tilde{\mathcal{B}}$ parameterizes the set of vicinity functions available to us.

\subsection{A Loss Function Defined on a Set of Models and Localities} 
Consider $n$ spaces of points of interest $\{ \Ccal_{i} \}_{i=1}^{n}$ and let $\bm{c} \eqdef (\cb_{1}, \dots, \cb_{n}) \in \Ccal_{1} \times \dots \times \Ccal_{n}$ denote the vector containing $n$ points of interest. Then, consider $n+1$ classes of predictors $\Hcal_{1}, \cdots, \Hcal_{n}$, $\Hcal_{ext}$, and some loss function $\ell(h(\xb),y)$ that quantify the loss incured by $h$ on example $(\xb,y)$. Let $\bm{h}$ denote the vector containing the $n$ predictors $(h_{1}, \cdots, h_{n}) \in \Hcal_{1} \times \cdots \times \Hcal_{n}$. 

Consider $K(\cb, \xb, \be)$ a vicinity function and $\be \in \Tilde{\Bcal}$. Let us denote by $\boldsymbol{\be}$ the vector containing $(\be_{1},  \cdots, \be_{n}) \in \Tilde{\Bcal}_{1} \times \cdots \times \Tilde{\Bcal}_{n}$. Thus, for all $(\bm{c}, \bm{h}, \boldsymbol{\be}, h_{ext}) \in \{ \Ccal_{i} \times \Hcal_{i} \times \Tilde{\Bcal}_{i} \}_{i=1}^{n} \times \Hcal_{ext}$, we propose the following loss function:
\begin{equation}\label{eq:general_loss_expression}
    \ell\Bigl[\bm{c}, \bm{h}, \boldsymbol{\be}, h_{ext}, (\xb, y) \Bigr] \eqdef \sum_{i=1}^{n} \Bigl[ \ell(h_{i}(\xb), y) K(\cb_{i}, \xb, \be_{i}) \Bigr] + \ell(h_{ext}(\xb), y)  \prod_{i=1}^{n} \bar{K}(\cb_{i}, \xb, \be_{i})\, \textrm{.}
\end{equation}
Three scenarios are taken into account when we formulate the loss as in \eqref{eq:general_loss_expression}: 
\begin{itemize}
    \item The first scenario is the case where the instance $\xb$ only falls into one locality $\be_{r}$ with $1 \leq r \leq n$, which means $K(\cb_{i}, \xb, \be_{i}) = 1$ when $i=r$ and $K(\cb_{i}, \xb, \be_{i}) = 0$ for $i \neq r$.  In this case, the loss depends only of the output of predictor $h_r$.
    \item The second scenario takes care of the situation where the instance $\xb$ doesn't belong to any locality, which means $\forall \; i \in [n], K(\cb_{i}, \xb, \be_{i}) = 0 \Leftrightarrow \prod_{i=1}^{n} \bar{K}(\cb_{i}, \xb, \be_{i}) = 1$. In this case the loss depends only on the output of $h_{ext}$.
    \item The last scenario concerns the case where the instance $\xb$ falls into several localities at the same time. In this circumstance, the loss depends on the output of all these predictors and we have zero loss only if each of them predicts the correct label $y$.
\end{itemize}

Consequently, this (uncommon) multi-predictor loss function penalizes contradictory predictions in overlapping neighborhoods given by the vicinity functions. However, this loss does not penalize overlapping neighborhoods when all the corresponding predictors are accurate in their neighborhoods. Finally, note that nothing in the current model specifies what to predict when a test instance falls into several (overlapping) neighborhoods with contradictory predictions. We feel that in such ambiguous cases, it is mostly important to specify the ambiguity and leave the decision of what to predict to the user.

In the PAC-Bayesian learning setting, we have to provide on each parameter class $\mathcal{C}_{i}$, $\mathcal{H}_{i}$, $\Tilde{\mathcal{B}}_{i}$ and $\mathcal{H}_{ext}$, some \textit{prior} $P$ and \textit{posterior} $Q$ distributions. For \textit{priors}, we have, $\forall \; i \in [n]$: $\{ \mathbf{c}_{i}, h_{i}, \beta_{i} \} \sim \{ P_{\mathcal{C}_{i}}, P_{\mathcal{H}_{i}} , P_{\Tilde{\mathcal{B}}_{i}} \}$ and $h_{ext} \sim P_{\mathcal{H}_{ext}}$. With \textit{posteriors}, we have, $\forall \; i \in [n]$: $\{ \mathbf{c}_{i}, h_{i}, \beta_{i} \} \sim \{ Q_{\mathcal{C}_{i}}, Q_{\mathcal{H}_{i}} , Q_{\Tilde{\mathcal{B}}_{i}} \}$ and $h_{ext} \sim Q_{\mathcal{H}_{ext}}$. All of these elements allow us to define the loss
\begin{equation}\label{eq:perte_Q_exp_generale}
    \ell\Bigl[ Q, (\mathbf{x}, y) \Bigr] \eqdef \underset{Q}{\mathds{E}} \left\{ \ell\Bigl[\bm{c}, \bm{h}, \boldsymbol{\beta}, h_{ext}, (\mathbf{x}, y) \Bigr] \right\}\, ,
\end{equation}
where $Q \eqdef \left(\{ Q_{\mathcal{C}_{i}}, Q_{\mathcal{H}_{i}} , Q_{\Tilde{\mathcal{B}}_{i}} \}_{i=1}^{n}, Q_{\mathcal{H}_{ext}}\right)$.

The same loss can also be applied to the prior $P \eqdef \left(\{ P_{\mathcal{C}_{i}}, P_{\mathcal{H}_{i}} , P_{\Tilde{\mathcal{B}}_{i}} \}_{i=1}^{n}, P_{\mathcal{H}_{ext}}\right)$.

\subsection{Risk Function}
From the loss function~\eqref{eq:perte_Q_exp_generale}, for all $(\bm{c}, \bm{h}, \boldsymbol{\beta}, h_{ext}) \in \{ \mathcal{C}_{i} \times \mathcal{H}_{i} \times \Tilde{\mathcal{B}}_{i} \}_{i=1}^{n} \times \mathcal{H}_{ext}$, the goal is to find the vector $\bm{c}$ of $n$ points of interest, the vector $\bm{h}$ of predictors, the vector $\boldsymbol{\beta}$ of locality parameters, and the external predictor $h_{ext}$ which minimize the risk $L_{ \mathcal{D}}(Q)$ defined as
\begin{equation}\label{eq:risque_généralisation_forme_espérance}
    L_{\mathcal{D}}(Q) \eqdef \underset{(\mathbf{x},y)\sim\mathcal{D}}{\mathds{E}} \Biggl\{ \ell\Bigl[ Q, (\mathbf{x}, y) \Bigr] \Biggr\}\; \textrm{.}
\end{equation}
Then, the empirical risk is given by
\begin{equation}\label{eq:risque_généralisation_forme_empirique}
        L_{S}(Q) \eqdef \frac{1}{m} \sum_{j=1}^{m} \Biggl\{ \ell\Bigl[ Q, (\mathbf{x}_{j}, y_{j}) \Bigr] \Biggr\}\; \textrm{.}
\end{equation}

\subsection{The Kullback-Leibler (KL) Divergence Between Q and P}
In this paper, we consider\footnote{A priori, we have no reason to consider the more complicated case where there would be association/correlation between $\left\{\mathbf{c}_{i}, h_{i}, \beta_{i} \right\}_{i=1}^{n}, h_{ext}$.} that the \textit{priors} and \textit{posteriors} respectively assigned to variables $\left\{\mathbf{c}_{i}, h_{i}, \beta_{i} \right\}_{i=1}^{n}$ and $h_{ext}$ are such that all these variables are independent. The global $KL$-divergence thus becomes a sum of $KL$-divergences for each random variable, i.e., we have
\begin{equation}\label{eq:forme_kl}
        KL(Q || P) = \sum_{i=1}^{n} \Bigl[ KL\left(Q_{\mathcal{C}_{i}} || P_{\mathcal{C}_{i}}\right) + KL\left(Q_{\mathcal{H}_{i}} || P_{\mathcal{H}_{i}}\right) + KL\left(Q_{\Tilde{\mathcal{B}}_{i}} || P_{\Tilde{\mathcal{B}}_{i}}\right) \Bigr] + KL\left(Q_{\mathcal{H}_{ext}} || P_{\mathcal{H}_{ext}}\right) \textrm{.}
\end{equation}
It is important to note that this global $KL$-divergence will ensure that the expectations and variances of \textit{posteriors} do not deviate too much from the corresponding values of the \textit{priors}. Hence, combining the expression of $KL$ from~\eqref{eq:forme_kl} with that of empirical risk $L_{\mathcal{S}}(Q)$, the PAC-Bayesian learning algorithm $\mathbf{PB_{\left\{\mathcal{C}_{i} \times \mathcal{H}_{i} \times \Tilde{\mathcal{B}}_{i} \right\}_{i=1}^{n} \times \mathcal{H}_{ext}}^{\lambda}}(S)$ which optimizes a PAC-Bayesian bound on $L_{\mathcal{D}}(Q)$ is given by
\begin{equation}\label{eq:pac_bayesian_bound_algo}
    \mathbf{PB_{\left\{\mathcal{C}_{i} \times \mathcal{H}_{i} \times \Tilde{\mathcal{B}}_{i} \right\}_{i=1}^{n} \times \mathcal{H}_{ext}}^{\lambda}}(S) \eqdef \underset{Q \ll P}{\mathrm{argmin}} 
    \Biggl[ L_{\mathcal{S}}(Q) + \frac{1}{\lambda} KL(Q||P) \Biggr] \textrm{.}
\end{equation}
Equation~\eqref{eq:pac_bayesian_bound_algo} will return some quite accurate \textit{posterior} $Q$ over all variables. Then, we will construct deterministic predictors using the means of the \textit{posterior} $Q$, which means that the variables $\LC \cb_{i}, h_{i}, \be_{i} \RC_{i=1}^{n}$ and $h_{ext}$ will be estimated by, respectively, the expectations $\LC \EE_{Q_{\Ccal_{i}}} \LB \cb_{i} \RB, \EE_{Q_{\Hcal_{i}}} \LB h_{i} \RB, \EE_{Q_{\Tilde{\Bcal}_{i}}} \LB \be_{i} \RB \RC_{i=1}^{n}$, and $\EE_{Q_{\Hcal_{ext}}} \LB h_{ext} \RB$. Let us now define the hypothesis classes $\left\{ \mathcal{H}_{i} \right\}_{i=1}^{n} \times \mathcal{H}_{ext}$ and $\left\{\mathcal{C}_{i} \times \Tilde{\mathcal{B}}_{i} \right\}_{i=1}^{n}$, and their corresponding \textit{priors} and \textit{posteriors} in order to obtain closed forms and easy-to-compute expressions for $L_{\mathcal{S}}(Q)$ and $KL(Q || P)$.

\subsection{Hypothesis Classes}
We work with the linear hypothesis classes as transparent models. So, each predictor $h_{i} \in \left\{ \mathcal{H}_{i} \right\}_{i=1}^{n}$ (respectively $h_{ext} \in \mathcal{H}_{ext}$) will be specified by their weights $\mathbf{v}_{i}$ (respectively $\mathbf{v}_{ext}$). Thus, the hypothesis classes $\left\{ \mathcal{H}_{i} \right\}_{i=1}^{n}$ and $\mathcal{H}_{ext}$ will be designated respectively by the spaces of the weights $\left\{ \mathcal{W}_{i} \right\}_{i=1}^{n}$ and $\mathcal{W}_{ext}$.

Additionally, we will use $n+1$ bias classes $\left\{\mathcal{B}_{i} \right\}_{i=1}^{n}$ and $\mathcal{B}_{ext}$ where each bias $b_{i} \in \left\{\mathcal{B}_{i} \right\}_{i=1}^{n}$ (respectively $b_{ext} \in \mathcal{B}_{ext}$) will be used jointly with a vector of weights $\mathbf{v}_{i} \in \mathcal{W}_{i}$ (respectively $b_{ext} \in \mathcal{B}_{ext}$ will be used jointly with the weight vector $\mathbf{v}_{ext} \in \mathcal{W}_{ext}$).

Now, we choose $\mathds{R}^{d}$ for each weight space $\mathcal{W}_{i}$ and $\mathcal{W}_{ext}$, and for each space $\mathcal{C}_{i}$ of point of interest. We also use $\mathcal{B}_{i} = \mathds{R}$ and $\mathcal{B}_{ext} = \mathds{R}$ for the $n+1$ classes of biases, and ${\Tilde{\mathcal{B}}}_{i} = \mathds{R}_{+}$ for the $n$ classes of vicinity function parameters.

\subsection{Classes of Distributions}
For each space of point of interest $\mathcal{C}_{i} = \mathds{R}^{d}$, we choose a multivariate Gaussian distribution $\mathcal{G}_{\mathcal{C}_{i}} \eqdef \mathcal{N}\left(\cb_{i_{0}}, \mathbf{I}_{i} \varep_{i}^{2}\right)$ with the expectation $\cb_{i_{0}} \in \mathds{R}^{d}$, the identity matrix $\mathbf{I}_{i} \in \mathds{R}^{d \times d}$ and the standard deviation $\varep_{i} \in \mathds{R}_{+}$.

For each weight space $\mathcal{W}_{i} = \mathds{R}^{d}$, we also use a multivariate Gaussian distribution $\mathcal{G}_{\mathcal{W}_{i}} \eqdef \mathcal{N}\left(\mathbf{w}_{i}, \mathbf{I}_{i} \rho_{i}^{2}\right)$ with expectation $\mathbf{w}_{i} \in \mathds{R}^{d}$ and standard deviation $\rho_{i} \in \mathds{R}_{+}$. Similarly, for the space $\mathcal{W}_{ext} = \mathds{R}^{d}$, we have a  distribution $\mathcal{G}_{\mathcal{W}_{ext}} \eqdef \mathcal{N}\left(\mathbf{w}_{ext}, \mathbf{I}_{ext} \rho_{ext}^{2}\right)$.

For each class of biases $\mathcal{B}_{i} = \mathds{R}$, we choose a univariate Gaussian distribution $\mathcal{G}_{\mathcal{B}_{i}} \eqdef \mathcal{N}\left(\mu_{i}, \sigma_{i}^{2}\right)$ with expectation $\mu_{i} \in \mathds{R}$ and standard deviation $\sigma_{i} \in \mathds{R}_{+}$. Similarly, for the bias space $\mathcal{B}_{ext} = \mathds{R}$, we use a univariate Gaussian distribution $\mathcal{G}_{\mathcal{B}_{ext}} \eqdef \mathcal{N}\left(\mu_{ext}, \sigma_{ext}^{2}\right)$.

Finally, for each set of vicinity function parameters ${\Tilde{\mathcal{B}}}_{i} = \mathds{R}_{+}$, we choose a Gamma distribution $\mathcal{G}_{{\Tilde{\mathcal{B}}}_{i}} \eqdef \Gamma\left(k_{i}, \tau_{i}\right)$ with shape $k_{i} \in \mathds{R}_{+}^{*}$ and rate $\tau_{i} \in \mathds{R}_{+}^{*}$. Recall that the probability density function of a Gamma distribution $\Gamma\left(k, \tau\right)$ is given by
\[
    q_{k, \tau} (\beta) = \frac{\beta^{k-1} \tau^{k}}{\Gamma(k)} e^{-\tau\beta}, \; \beta \in \mathds{R}_{+}\, \textrm{,}
\]
where $\Gamma(\cdot)$ denotes the Euler Gamma function defined as
\[
    \forall \; x > 0, \quad \Gamma(x) = \int_{0}^{+\infty} t^{x-1} e^{-t} dt\, \textrm{.}
\]
Furthermore, according to \citet{duchi2007derivations, soch2016kullback}, the KL divergence between two multivariate Gaussian distributions in 
$\mathds{R}^{d}$, denoted by $KL\left[\mathcal{G}_{\mathcal{W}_{1}} || \mathcal{G}_{\mathcal{W}_{2}} \right] \eqdef KL\left[\mathcal{N}\left(\mathbf{w}_{1}, \mathbf{I}_{1} \rho_{1}^{2}\right) || \mathcal{N}\left(\mathbf{w}_{2}, \mathbf{I}_{2} \rho_{2}^{2}\right) \right]$, is given by
\begin{equation}\label{eq:kl_between_two_gaussians}
    KL\left[\mathcal{G}_{\mathcal{W}_{1}} || \mathcal{G}_{\mathcal{W}_{2}} \right] = \frac{1}{2}\Biggl[ \ln{\left\{ \left(\frac{\rho_{2}^{2}}{\rho_{1}^{2}}\right)^{d} \right\}} - d + d \left(\frac{\rho_{1}^{2}}{\rho_{2}^{2}}\right) + \frac{1}{\rho_{2}^{2}}(\mathbf{w}_{2} - \mathbf{w}_{1})^{T}(\mathbf{w}_{2} - \mathbf{w}_{1}) \Biggr]\, \textrm{,}
\end{equation} 
and that between two Gamma distributions, denoted by $KL\left[\mathcal{G}_{{\Tilde{\mathcal{B}}}_{1}} || \mathcal{G}_{{\Tilde{\mathcal{B}}}_{2}} \right] \eqdef$ \newline $ KL\left[\Gamma\left(k_{1}, \tau_{1}\right) || \Gamma\left(k_{2}, \tau_{2}\right) \right]$, is given by \begin{equation}\label{eq:kl_between_two_gamma}
    KL\left[\mathcal{G}_{{\Tilde{\mathcal{B}}}_{1}} || \mathcal{G}_{{\Tilde{\mathcal{B}}}_{2}} \right] = ( k_{1} - k_{2} )\psi(k_{1}) + \ln\left(\frac{\Gamma(k_{2})}{\Gamma(k_{1})}\right) + k_{2}\ln\left(\frac{\tau_{1}}{\tau_{2}}\right) + k_{1}\frac{\tau_{2} - \tau_{1}}{\tau_{1}}\, \textrm{,}
\end{equation}
where $\psi$ designates the Digamma function defined as
\[
    \forall \; x > 0, \quad \psi(x) \eqdef \frac{d}{dx}\ln[\Gamma(x)] = \frac{\Gamma^{'}(x)}{\Gamma(x)}\, \textrm{.}
\]

\begin{remark}
    The simplicity and interpretability of the internal and external predictors of our mixture of transparent models are the focus of this paper. However, it is possible to design, using exactly the same approach, a mixture of opaque local models. To do this, just consider some feature space $\phi: \Xcal \rightarrow \mathds{R}^{p}$ for some, possibly very large, feature-space dimension $p$. So instead of $\xb$ and $\cb$, we would have $\phi(\xb)$ and $\phi(\cb)$. The set of predictors to consider would be such that their output on any input $\xb$ would be given by some function of $\langle \mathbf{w}, \phi(\xb)\rangle$. Thanks to the representer theorem, we then have that $\wb = \sum_{j=1}^m \alpha_j \phi(\xb_j)$ for some $(\alpha_1, \ldots,\alpha_m)$ and, consequently, $\langle \mathbf{w}, \phi(\xb)\rangle = \sum_{j=1}^m \alpha_j K(\xb_j,\xb)$ where $K(\xb_j,\xb) = \langle\phi(\xb_j),\phi(\xb)\rangle$ is the kernel associated to the feature space. As for the geometry of the localities, it would also be determined by the choice of the metric $d(\cb, \xb)$ where $\|\phi(\cb),\phi(\xb)\|$ is one possible choice. In that last case $d(\cb, \xb)$ is also expressible using kernels. This implies that much of what we do allows the use of kernels at the cost of a substantial reduction in interpretability, but we not investigate this research direction in this paper.
\end{remark}

\section{Mixtures of Transparent Local Models with Known Points of Interest}
\label{section_given_gn}
In this section, we suppose that we know each point of interest  beforehand, in the vicinity of which we want to construct a transparent local model. Thus, starting from the preliminary notions presented above, here, the points of interest do not constitute unknown variables. Therefore, only the $n+1$ predictors $\LC \vb_{i} \RC_{i=1}^{n}$ and $\vb_{ext}$, $n+1$ biases $\LC b_{i} \RC_{i=1}^{n}$ and $b_{ext}$, and the $n$ locality parameters $\LC \be_{i} \RC_{i=1}^{n}$ have to be sought. For that purpose, we will consider the case of binary linear classification and linear regression.

\subsection{Binary Linear Classification}
\label{Binary_Linear_clf_gn}
Since it is often advised to standardize data to ensure that all features of the input space are on the same scale, the natural choice for the \textit{priors} of variables $\LC \vb_{i}, b_{i} \RC_{i=1}^{n}$, $\vb_{ext}$ and $b_{ext}$ would be isotropic standard Gaussian distributions because they are generic weakly informative priors. As for their \textit{posteriors}, let’s also choose reduced isotropic Gaussian distributions but centered on some vectors of non-zero weight.

For positive variables $\LC \be_{i} \RC_{i=1}^{n}$, let’s choose as \textit{prior}, a Gamma distribution $\Gm\LP k^{'},  \tau^{'}  \RP$ with shape $k^{'}=2$ and rate $\tau^{'}=0.1$ as recommended by \citet{chung2013nondegenerate}. This \textit{prior} will particularly keep the mode away from 0 but still allows it to be arbitrarily close if the likelihood indicates that based on the data. With their \textit{posteriors}, let’s also use a Gamma distribution $\Gm\LP k, \tau \RP$ where $k$ and $\tau$ will be learned from the training data.

Note that if the appropriate range values for the different variables $\LC \vb_{i}, b_{i}, \be_{i} \RC_{i=1}^{n}$, $\vb_{ext}$ and $b_{ext}$ were known beforehand, that information could have been included in their \textit{prior} distributions. But, unfortunately, this is generally not the case. 

Table \ref{tab:priors_and_posteriors_clf_gn} shows all of the chosen \textit{priors} $P$ and \textit{posteriors} $Q$. Specifically, $P$ denotes the set of prior distributions $\Bigl( \LC P_{\mathbf{0}, \Ib_{i}}, P_{0, 1}, P_{k'_{i},\tau'_{i}} \RC_{i=1}^{n}$, $P_{\mathbf{0}, \Ib_{ext}}$, $P_{0, 1} \Bigr)$, and $Q$ denotes the set of posterior distributions $\Bigl( \LC Q_{\wb_{i}, \Ib_{i}}, Q_{\mu_{i}, \sg_{i}^{2}}, Q_{k_{i},\tau_{i}} \RC_{i=1}^{n}$, $Q_{\wb_{ext}, \Ib_{ext}}$, $Q_{\mu_{ext}, \sg_{ext}^{2}} \Bigr)$. 

\begin{table}[H]
\centering
\begin{tabular}{l|l}
\toprule
\multicolumn{1}{c|}{\textit{Priors}} & \multicolumn{1}{c}{\textit{Posteriors}} \\
\midrule
\multirow{2}{*}{\text{$\vb_{i} \sim P_{\mathbf{0}, \Ib_{i}} \eqdef \Ncal\LP \mathbf{0}, \Ib_{i} \RP$, $\forall \; i \in [n]$}} & \multirow{2}{*}{\text{$\vb_{i} \sim Q_{\wb_{i}, \Ib_{i}} \eqdef \Ncal\LP \wb_{i}, \Ib_{i} \RP$, $\forall \; i \in [n]$}} \\ \\
\multirow{2}{*}{\text{$b_{i} \sim P_{0, 1} \eqdef \Ncal\LP 0, 1 \RP$, $\forall \; i \in [n]$}} & \multirow{2}{*}{\text{$b_{i} \sim Q_{\mu_{i}, \sg_{i}^{2}} \eqdef \Ncal\LP\mu_{i}, \sg_{i}^{2} \RP$, $\forall \; i \in [n]$}} \\ \\
\multirow{2}{*}{\text{$\be_{i} \sim P_{k_{i}^{'}, \tau_{i}^{'}} \eqdef \Gm\LP 2, 0.1 \RP$, $\forall \; i \in [n]$}} & \multirow{2}{*}{\text{$\be_{i} \sim Q_{k_{i}, \tau_{i}} \eqdef \Gm\LP k_{i}, \tau_{i} \RP$, $\forall \; i \in [n]$}} \\ \\
\multirow{2}{*}{\text{$\vb_{ext} \sim P_{\mathbf{0}, \Ib_{ext}} \eqdef \Ncal\LP \mathbf{0}, \Ib_{ext} \RP$}} & \multirow{2}{*}{\text{$\vb_{ext} \sim Q_{\wb_{ext}, \Ib_{ext}} \eqdef \Ncal\LP \wb_{ext}, \Ib_{ext} \RP$}} \\ \\
\multirow{2}{*}{\text{$b_{ext} \sim P_{0, 1} \eqdef \Ncal\LP 0, 1 \RP$}} & \multirow{2}{*}{\text{$b_{ext} \sim Q_{\mu_{ext}, \sg_{ext}^{2}} \eqdef \Ncal\LP \mu_{ext}, \sg_{ext}^{2} \RP$}} \\ \\
\bottomrule
\end{tabular}
\caption{Prior and posterior distributions for binary linear classification task.}
\label{tab:priors_and_posteriors_clf_gn}
\end{table}

We thus have all the necessary ingredients to calculate the KL divergence between \textit{posteriors} and \textit{priors}, denoted $KL_{cl}\LP Q || P \RP$. Plugging~\eqref{eq:kl_between_two_gaussians} and \eqref{eq:kl_between_two_gamma} into~\eqref{eq:forme_kl} yields
\begin{equation}\label{eq:kl_clf_gn_final}
    \begin{aligned}
        KL_{cl}\LP Q || P \RP &= - \LB \sum_{i=1}^{n} \ln\LP \sg_{i} \RP + \ln\LP \sg_{ext} \RP \RB + \frac{1}{2} \LB \sum_{i=1}^{n} \Bigl( \sg_{i}^{2} - 1 \Bigr) + \sg_{ext}^{2} - 1 \RB \\ 
        & \quad + \frac{1}{2} \LB \sum_{i=1}^{n} \Bigl( \LN \wb_{i} \RN^{2} + \mu_{i}^{2} \Bigr) + \LN \wb_{ext} \RN^{2} + \mu_{ext}^{2} \RB \\ 
        & \quad + \sum_{i=1}^{n} \LB ( k_{i} - k_{i}^{'} )\psi(k_{i}) + \ln\LP \frac{\Gm(k_{i}^{'})}{\Gm(k_{i})} \RP + k_{i}^{'}\ln\LP \frac{\tau_{i}}{\tau_{i}^{'}} \RP + k_{i}\frac{\tau_{i}^{'} - \tau_{i}}{\tau_{i}} \RB \textrm{.} 
    \end{aligned}
\end{equation}

Let us now use a different notation for the loss function which is expressed in terms of the parameters of the predictors. Considering that $h$ is parameterized by $\vb$ and $b$, instead of $\ell(h(\mathbf{x}), y)$ we will use
\[
    \ell\Bigl( \vb, b, (\xb, y) \Bigr) \eqdef \mathds{1}\LB y \Bigl( \langle \vb, \xb \rangle + b \Bigr) \leq 0 \RB \textrm{,}
\]
where $\mathds{1}$ denotes the indicator function defined by
\[
    \mathds{1}[\text{a}] =
        \begin{cases}
          1 & \text{if a is true}\\
          0 & \text{otherwise.}
        \end{cases}       
\]
Thus, from~\eqref{eq:general_loss_expression} we have the loss function $\ell : \left\{\mathcal{C}_{i} \times \mathcal{W}_{i} \times \mathcal{B}_{i} \times \Tilde{\mathcal{B}}_{i} \right\}_{i=1}^{n} \times \mathcal{W}_{ext} \times \mathcal{B}_{ext}$, $\mathcal{X} \times \mathcal{Y} \rightarrow \mathds{R}_{+}$ such that 
\[
    \begin{aligned}
        \ell\Bigl[ \Bigl(\bm{c}, \bm{v}, \bm{b}, \bbeta, \vb_{ext}, b_{ext} \Bigr), (\xb, y) \Bigr] &\eqdef \sum_{i=1}^{n} \LB \ell\Bigl( \vb_{i}, b_{i}, (\xb, y) \Bigr) K(\cb_{i}, \xb, \be_{i}) \RB \\ 
        & \quad \quad \quad \quad \quad + \ell\Bigl( \vb_{ext}, b_{ext}, (\xb, y) \Bigr) \prod_{i=1}^{n} \bar{K}(\cb_{i}, \xb, \be_{i}) \textrm{.}
    \end{aligned}
\]
For the binary classification case, we will use $y \in \Ycal = \{-1, +1\}$. Then, by definition, the expression of loss $\ell\Bigl[ Q, (\xb, y) \Bigr]$ is given by
\[
    \begin{aligned}
        \ell\Bigl[ Q, (\xb, y) \Bigr] &\eqdef \underset{Q}{\EE} \LC \ell\Bigl[ \Bigl(\bm{c}, \bm{v}, \bm{b}, \bbeta, \vb_{ext}, b_{ext} \Bigr), (\xb, y) \Bigr] \RC \\ 
        &= \underset{Q}{\EE} \LC \sum_{i=1}^{n} \Biggl[ \mathds{1}\LB y \Bigl( \langle \vb_{i}, \xb \rangle + b_{i} \Bigr) \leq 0 \RB K(\cb_{i}, \xb, \be_{i}) \Biggr] \RC \\
        & \quad \quad \quad \quad \quad + \underset{Q}{\EE} \LC \mathds{1}\LB y \Bigl( \langle \vb_{ext}, \xb \rangle + b_{ext} \Bigr) \leq 0 \RB \prod_{i=1}^{n} \bar{K}(\cb_{i}, \xb, \be_{i}) \RC \textrm{.} 
    \end{aligned}
\]

Let's consider $\bpsi \eqdef y \mathbf{x}$ and the loss $\ell\Bigl[ Q, (\bpsi, \mathbf{x}, y) \Bigr]$ defined as
\begin{equation}\label{eq:perte_Q_exp_generale_clf_gn_def}
    \begin{aligned}
        \ell\Bigl[ Q, (\bpsi, \xb, y) \Bigr] &\eqdef \underset{Q}{\EE} \LC \sum_{i=1}^{n} \Biggl[ \mathds{1}\LB \Bigl( \langle \vb_{i}, \bpsi \rangle + y b_{i} \Bigr) \leq 0 \RB K(\cb_{i}, \xb, \be_{i}) \Biggr] \RC \\
        & \quad \quad \quad \quad \quad + \underset{Q}{\EE} \LC \mathds{1}\LB \Bigl( \langle \vb_{ext}, \bpsi \rangle + y b_{ext} \Bigr) \leq 0 \RB \prod_{i=1}^{n} \bar{K}(\cb_{i}, \xb, \be_{i}) \RC \textrm{.}
    \end{aligned}
\end{equation}
Thus, the empirical risk $L_{S}(Q)$ defined in~\eqref{eq:risque_généralisation_forme_empirique} becomes
\begin{equation}\label{eq:nouveau_risque_généralisation_forme_empirique}
    L_{S}(Q) = \frac{1}{m} \sum_{j=1}^{m} \ell\Bigl[ Q, (\bpsi_{j}, \xb_{j}, y_{j}) \Bigr] \textrm{.} 
\end{equation}
Since $L_{S}(Q)$ depends now on $\ell\Bigl[ Q, (\bpsi, \xb, y) \Bigr]$, let's find the expression of $\ell\Bigl[ Q, (\bpsi, \xb, y) \Bigr]$ to deduce that of $L_{S}(Q)$. So, let us now consider the following quantity
\begin{equation}\label{eq:theta}
     \Theta_{k, \tau, d(\cb, \xb)} \eqdef \int_{d(\cb, \xb)}^{+\infty} \frac{\beta^{k-1} \tau^{k}}{\Gamma(k)} e^{-\tau\beta}d\beta \textrm{.}
\end{equation}
Here, $\Theta_{k, \tau, d(\cb, \xb)}$ is just the survival function of Gamma distribution which means it is the probability that the variate $\beta$ takes a value greater than $d(\cb, \xb)$. Thus, as detailed in Appendix~\ref{app:Binary_Linear_clf_gn}, the loss $\ell\Bigl[ Q, (\bpsi, \xb, y) \Bigr]$ can be expressed as
\begin{equation}\label{eq:perte_Q_exp_generale_clf_gn}
    \begin{aligned}
        \ell\Bigl[ Q, (\bpsi, \xb, y) \Bigr] &= \sum_{i=1}^{n} \LB \Theta_{k_{i}, \tau_{i}, d(\cb_{i}, \xb)} \times \LC 1 - \Phi\LP \frac{y (\langle \wb_{i}, \xb \rangle + \mu_{i})}{\sqrt{\sg_{i}^{2} + \LN \xb \RN^{2}}} \RP \RC \RB \\ 
        & \quad \quad \quad + \prod_{i=1}^{n} \LB 1 - \Theta_{k_{i}, \tau_{i}, d(\cb_{i}, \xb)} \RB \times \LC 1 - \Phi\LP \frac{y (\langle \wb_{ext}, \xb \rangle + \mu_{ext})}{\sqrt{\sg_{ext}^{2} + \LN \xb \RN^{2}}} \RP \RC \textrm{,}
    \end{aligned}
\end{equation}
where $\Theta_{k, \tau, d(\mathbf{c}, \mathbf{x})}$ is given by~\eqref{eq:theta} and $\Phi$ is the cumulative distribution function (CDF) of the standard normal distribution. 

Here, one can see that $\ell\Bigl[ Q, (\bpsi, \xb, y) \Bigr]$ is a non-convex function of the predictors' and vicinity function's parameters, and decreases to zero when both the locality parameters with their corresponding predictors predict the label of the instance $\xb$ accurately. Consequently, we remark that without having any variance parameters for the proposed Gaussian \textit{posteriors} related to the predictors $\LC \vb_{i} \RC_{i=1}^{n}$ and $\vb_{ext}$, we still obtain a loss $\ell\Bigl[ Q, (\bpsi, \xb, y) \Bigr]$ which can vanish. Thus, using variance parameters for the Gaussian \textit{posteriors} associated with predictors $\LC \vb_{i} \RC_{i=1}^{n}$ and $\vb_{ext}$, is not required here.

The closed-form expression of $\ell\Bigl[ Q, (\bpsi, \xb, y) \Bigr]$ of~\eqref{eq:perte_Q_exp_generale_clf_gn} gives us the closed-form expression of the true risk $L_{\mathcal{D}}(Q)$ and the empirical risk $L_{S}(Q)$ given by~\eqref{eq:risque_généralisation_forme_espérance} and~\eqref{eq:nouveau_risque_généralisation_forme_empirique}. Since we also have a closed-form expression of the $KL$ divergence, we can then perform efficiently the PAC-Bayesian learning algorithm $\mathbf{PB_{\left\{\mathcal{C}_{i} \times \mathcal{W}_{i} \times \mathcal{B}_{i} \times \Tilde{\mathcal{B}}_{i} \right\}_{i=1}^{n} \times \mathcal{W}_{ext} \times \mathcal{B}_{ext}}^{\lambda}}(S)$, denoted $\mathbf{PB^{\lambda}}(S)$, which now becomes
\begin{equation}\label{eq:clf_gn_bound_minimize_form}
    \mathbf{PB}^\lambda(S) = \underset{\LC \wb_{i}, \mu_{i}, \sg_{i}, k_{i}, \tau_{i} \RC_{i=1}^{n}, \wb_{ext}, \mu_{ext}, \sg_{ext}} {\mathrm{argmin}} L_{S}(Q) + \frac{1}{\lambda} KL_{cl}(Q||P)\; \textrm{.}
\end{equation}
However, the minimization objective is a non convex function of $\left\{ \mathbf{w}_{i}, \mu_{i}, \sg_{i}, k_{i}, \tau_{i} \right\}_{i=1}^{n}$ and can thus have multiple local minima. In practice, we need several random restarts of the learning algorithm $\mathbf{PB}^\lambda(S)$ to be able to find a good solution.

\subsection{Linear Regression} \label{Linear_Regression_reg_gn}
The steps followed in Subsection \ref{Binary_Linear_clf_gn} to derive a closed-form expression for the loss function $\ell\Bigl[ Q, (\bpsi, \xb, y) \Bigr]$ and, consequently, for the empirical risk of $Q$ and the $KL$ divergence, are identical to those used here. Therefore, we will present only the key results in order to avoid redundancy.

In the linear regression case, we keep the same \textit{prior} distributions over all the variables $\LC \vb_{i}, b_{i}, \be_{i} \RC_{i=1}^{n}$, $\vb_{ext}$ and $b_{ext}$ as in the previous subsection. But, for \textit{posteriors}, we add some variance parameters on the Gaussian distributions related to the predictors $\LC \vb_{i} \RC_{i=1}^{n}$ and $\vb_{ext}$ as displayed in Table~\ref{tab:priors_and_posteriors_reg_gn}.

\begin{table}[H]
\centering
\begin{tabular}{l|l}
\toprule
\multicolumn{1}{c|}{\textit{Priors}} & \multicolumn{1}{c}{\textit{Posteriors}} \\
\midrule
\multirow{2}{*}{\text{$\mathbf{v}_{i} \sim P_{\mathbf{0}, \mathbf{I}_{i}} \eqdef \mathcal{N}\left(\mathbf{0}, \mathbf{I}_{i} \right)$, $\forall \; i \in [n]$}} & \multirow{2}{*}{\text{$\mathbf{v}_{i} \sim Q_{\mathbf{w}_{i}, \mathbf{I}_{i} \rho_{i}^{2}} \eqdef  \mathcal{N}\left(\mathbf{w}_{i}, \mathbf{I}_{i} \rho_{i}^{2} \right)$, $\forall \; i \in [n]$}} \\ \\
\multirow{2}{*}{\text{$b_{i} \sim P_{0, 1} \eqdef \mathcal{N}\left(0, 1 \right)$, $\forall \; i \in [n]$}} & \multirow{2}{*}{\text{$b_{i} \sim Q_{\mu_{i}, \sigma_{i}^{2}} \eqdef \mathcal{N}\left(\mu_{i}, \sigma_{i}^{2} \right)$, $\forall \; i \in [n]$}} \\ \\
\multirow{2}{*}{\text{$\beta_{i} \sim P_{k_{i}^{'}, \tau_{i}^{'}} \eqdef \Gamma\left(2, 0.1 \right)$, $\forall \; i \in [n]$}} & \multirow{2}{*}{\text{$\beta_{i} \sim Q_{k_{i}, \tau_{i}} \eqdef \Gamma\left(k_{i}, \tau_{i} \right)$, $\forall \; i \in [n]$}} \\ \\
\multirow{2}{*}{\text{$\mathbf{v}_{ext} \sim P_{\mathbf{0}, \mathbf{I}_{ext}} \eqdef \mathcal{N}\left(\mathbf{0}, \mathbf{I}_{ext} \right)$}} & \multirow{2}{*}{\text{$\mathbf{v}_{ext} \sim Q_{\mathbf{w}_{ext}, \mathbf{I}_{ext} \rho_{ext}^{2}} \eqdef \mathcal{N}\left(\mathbf{w}_{ext}, \mathbf{I}_{ext} \rho_{ext}^{2} \right)$}} \\ \\
\multirow{2}{*}{\text{$b_{ext} \sim P_{0, 1} \eqdef \mathcal{N}\left(0, 1 \right)$}} & \multirow{2}{*}{\text{$b_{ext} \sim Q_{\mu_{ext}, \sigma_{ext}^{2}} \eqdef \mathcal{N}\left(\mu_{ext}, \sigma_{ext}^{2} \right)$}} \\ \\
\bottomrule
\end{tabular}
\caption{Prior and posterior distributions for linear regression task.}
\label{tab:priors_and_posteriors_reg_gn}
\end{table}

Similarly to~\eqref{eq:kl_clf_gn_final}, we obtain for $KL_{reg}\left(Q || P\right)$ the following expression:
\begin{equation}\label{eq:kl_reg_gn_final}
    \begin{aligned}
        KL_{reg}\left(Q || P\right) &= - \left[ \sum_{i=1}^{n} \ln{\Bigl(\rho_{i}^{d} \times \sigma_{i} \Bigr)} + \ln{\Bigl(\rho_{ext}^{d} \times \sigma_{ext} \Bigr)} \right] + \frac{1}{2} \Biggl[ \sum_{i=1}^{n} \Bigl( d \rho_{i}^{2} + \sigma_{i}^{2} - d - 1 \Bigr) \\ 
        & \quad \quad + d \rho_{ext}^{2}  + \sigma_{ext}^{2} - d - 1 \Biggr] + \frac{1}{2} \Biggl[ \sum_{i=1}^{n} \Bigl( \left \| \mathbf{w}_{i} \right \|^{2} + {\mu_{i}}^{2} \Bigr) +  \left \| \mathbf{w}_{ext} \right \|^{2} + {\mu_{ext}}^{2} \Biggr] \\ 
        & \quad \quad + \sum_{i=1}^{n} \Biggl[ ( k_{i} - k_{i}^{'} )\psi(k_{i}) + \ln\left(\frac{\Gamma(k_{i}^{'})}{\Gamma(k_{i})}\right) + k_{i}^{'}\ln\left(\frac{\tau_{i}}{\tau_{i}^{'}}\right) + k_{i}\frac{\tau_{i}^{'} - \tau_{i}}{\tau_{i}} \Biggr]\, \textrm{.} 
    \end{aligned}
\end{equation}

As in the previous subsection, we will express $\ell(h(\mathbf{x}), y)$ as a function of the parameters of $h$ and write it as $\ell\Bigl( \mathbf{v}, b, (\mathbf{x}, y) \Bigr)$. As is often the case for linear regression, we will use the square loss:
\[
    \ell\Bigl( \mathbf{v}, b, (\mathbf{x}, y) \Bigr) \eqdef \Bigl( \langle \mathbf{v}, \mathbf{x} \rangle + b - y \Bigr)^{2} \textrm{,}
\]
where $y \in \mathcal{Y} = \mathds{R}$ for the linear regression case.

Therefore, from~\eqref{eq:general_loss_expression}, the multi-predictor loss function $\ell : \Bigl\{ \Ccal_{i} \times \Wcal_{i} \times \Bcal_{i} \times \Tilde{\Bcal}_{i} \Bigr\}_{i=1}^{n} \times \Wcal_{ext} \times \Bcal_{ext}$, $\Xcal \times \Ycal \rightarrow \mathds{R}_{+}$ is now given by
\[
    \begin{aligned}
        \ell\Bigl[ \Bigl(\bm{c}, \bm{v}, \bm{b}, \bbeta, \vb_{ext}, b_{ext} \Bigr), (\xb, y) \Bigr] &\eqdef \sum_{i=1}^{n} \LB \ell\Bigl( \vb_{i}, b_{i}, (\xb, y) \Bigr) K(\cb_{i}, \xb, \be_{i}) \RB \\ 
        & \quad \quad \quad \quad \quad + \ell\Bigl( \vb_{ext}, b_{ext}, (\xb, y) \Bigr) \prod_{i=1}^{n} \bar{K}(\cb_{i}, \xb, \be_{i}) \textrm{.}
    \end{aligned}
\]
Consequently, as shown in Appendix \ref{app:Linear_Regression_reg_gn}, the loss $\ell\Bigl[ Q, (\xb, y) \Bigr]$ of posterior $Q$ on example $(\xb,y)$ is given by
\begin{equation}\label{eq:perte_Q_exp_generale_reg_gn}
    \begin{aligned}
        \ell\Bigl[ Q, (\xb, y) \Bigr] &\eqdef \underset{Q}{\EE} \LC \ell\Bigl[ \Bigl(\bm{c}, \bm{v}, \bm{b}, \bbeta, \vb_{ext}, b_{ext} \Bigr), (\xb, y) \Bigr] \RC \\ 
        &= \sum_{i=1}^{n} \Biggl[ \Theta_{k_{i}, \tau_{i}, d(\cb_{i}, \xb)} \times \LC \LN \xb \RN^{2} \rho_{i}^{2} + \sg_{i}^{2} + \Bigl( \langle \wb_{i}, \xb \rangle + \mu_{i} - y \Bigr)^{2} \RC \Biggr] \\
        & \quad \quad \quad + \prod_{i=1}^{n} \LB 1 - \Theta_{k_{i}, \tau_{i}, d(\cb_{i}, \xb)} \RB \\
        & \quad \quad \quad \quad \quad \quad \quad \quad \times \LC \LN \xb \RN^{2} \rho_{ext}^{2} + \sg_{ext}^{2} + \Bigl( \langle \wb_{ext}, \xb \rangle + \mu_{ext} - y \Bigr)^{2} \RC \textrm{,} 
    \end{aligned}
\end{equation}
where $\Theta_{k, \tau, d(\cb, \xb)}$ is given by~\eqref{eq:theta}.

The closed-form expression of $\ell\Bigl[ Q, (\xb, y) \Bigr]$ of~\eqref{eq:perte_Q_exp_generale_reg_gn} gives us the closed-form expression of the true risk $L_{\mathcal{D}}(Q)$ and the empirical risk $L_{S}(Q)$ given by~\eqref{eq:risque_généralisation_forme_espérance} and~\eqref{eq:risque_généralisation_forme_empirique}. Since we also have a closed-form expression of the $KL$ divergence, we can then perform efficiently the PAC-Bayesian learning algorithm, which now becomes
\begin{equation}\label{eq:reg_gn_bound_minimize_form}
    \mathbf{PB^{\lambda}}(S) = \underset{\left\{ \mathbf{w}_{i}, \rho_{i}, \mu_{i}, \sigma_{i}, k_{i}, \tau_{i} \right\}_{i=1}^{n}, \mathbf{w}_{ext}, \rho_{ext}, \mu_{ext}, \sigma_{ext}}{\mathrm{argmin}} L_{S}(Q) + \frac{1}{\lambda} KL_{reg}(Q||P) \; \textrm{.}
\end{equation}
As in the previous subsection, the minimization objective is non convex in the parameters. Hence several restarts at randomly selected positions are needed to explore the parameter space with some success.

\section{Mixtures of Transparent Local Models with Unknown Points of Interest}
\label{section_search_gn}
This section extends the previous section work to tackle the situation where the $n$ points of interest are unknown, but the value of $n$ is known. Therefore, we have to seek all variables $\LC \cb_{i}, \vb_{i}, b_{i}, \be_{i} \RC_{i=1}^{n}$, $\vb_{ext}$ and $b_{ext}$ in order to design the mixtures of transparent local models.

\subsection{Binary Linear Classification}\label{Binary_Linear_clf_gn_search}
We keep the same \textit{priors} and \textit{posteriors} over the variables $\LC \vb_{i}, b_{i}, \be_{i} \RC_{i=1}^{n}$, $\vb_{ext}$ and $b_{ext}$ as in Subsection~\ref{Binary_Linear_clf_gn}. However, for the unknown points of interest $\LC \cb_{i} \RC_{i=1}^{n}$ we add the following \textit{prior} and \textit{posterior} distributions :
\begin{itemize}
    \item \textit{Priors} : $\cb_{i} \sim P_{\mathbf{0}, \mathbf{I}_{i}} \eqdef \Ncal\LP \mathbf{0}, \Ib_{i} \RP$, $\forall \; i \in [n]$
    \item \textit{Posteriors} : $\cb_{i} \sim Q_{\cb_{i_{0}}, \mathbf{I}_{i} \varep_{i}^{2}} \eqdef \Ncal\LP \cb_{i_{0}}, \Ib_{i} \varep_{i}^{2} \RP$, $\forall \; i \in [n]$.
\end{itemize}
In this scenario, the global $KL$-divergence, denoted $KL_{cl}^{'}\left(Q || P\right)$, just becomes 
\begin{equation}\label{eq:kl_clf_search_gn_final}
    KL_{cl}^{'}\left(Q || P\right) = KL_{cl}\left(Q || P\right) - \sum_{i=1}^{n} \ln{\Bigl( \varep_{i}^{d} \Bigr)} + \frac{1}{2} \sum_{i=1}^{n} \Bigl( d \varep_{i}^{2} - d \Bigr) + \frac{1}{2} \sum_{i=1}^{n} \left \| \cb_{i_{0}} \right \|^{2} \textrm{,}
\end{equation}
where $KL_{cl}\left(Q || P\right)$ is given by~\eqref{eq:kl_clf_gn_final}.

Now, let's come back to~\eqref{eq:perte_Q_exp_generale_clf_gn_def} in Subsection~\ref{Binary_Linear_clf_gn}, the loss function $\ell\Bigl[ Q, (\bpsi, \xb, y) \Bigr]$ of the posterior $Q$ is given by
\begin{align}
    \begin{split}
        \ell\Bigl[ Q, (\bpsi, \xb, y) \Bigr] &\eqdef \underset{Q}{\EE} \LC \sum_{i=1}^{n} \Biggl[ \mathds{1}\LB \Bigl( \langle \vb_{i}, \bpsi \rangle + y b_{i} \Bigr) \leq 0 \RB K(\cb_{i}, \xb, \be_{i}) \Biggr] \RC \\
        & \quad \quad \quad \quad \quad + \underset{Q}{\EE} \LC \mathds{1}\LB \Bigl( \langle \vb_{ext}, \bpsi \rangle + y b_{ext} \Bigr) \leq 0 \RB \prod_{i=1}^{n} \bar{K}(\cb_{i}, \xb, \be_{i}) \RC 
    \end{split} \\ 
    \begin{split}\label{eq:perte_Q_exp_generale_clf_gn_search}
        &= \sum_{i=1}^{n} \Biggl[ \underset{Q}{\EE} \LC \mathds{1}\LB \Bigl( \langle \vb_{i}, \bpsi \rangle + y b_{i} \Bigr) \leq 0 \RB K(\cb_{i}, \xb, \be_{i}) \RC \Biggr] \\
        & \quad \quad \quad \quad \quad + \underset{Q}{\EE} \LC \mathds{1}\LB \Bigl( \langle \vb_{ext}, \bpsi \rangle + y b_{ext} \Bigr) \leq 0 \RB \prod_{i=1}^{n} \bar{K}(\cb_{i}, \xb, \be_{i}) \RC
    \end{split} \\ 
    \begin{split}\label{eq:perte_Q_exp_generale_clf_gn_search_1}
        &= \sum_{i=1}^{n} \Biggl[ \underset{(\vb_{i}, b_{i}) \sim \LC Q_{\wb_{i}, \Ib_{i}} \times Q_{\mu_{i}, \sg_{i}^{2}} \RC}{\EE} \mathds{1}\LB \Bigl( \langle \vb_{i}, \bpsi \rangle + y b_{i} \Bigr) \leq 0 \RB \\ 
        & \quad \quad \quad \quad \quad \quad \quad \quad \quad \quad \quad \quad \quad \times \underset{ (\cb_{i}, \be_{i}) \sim \LC Q_{\cb_{i_{0}}, \Ib_{i} \varep_{i}^{2}} \times Q_{k_{i}, \tau_{i}} \RC }{\EE} K(\cb_{i}, \xb, \be_{i}) \Biggr] \\
        & \quad + \underset{(\vb_{ext}, b_{ext}) \sim \LC Q_{\wb_{ext}, \Ib_{ext}} \times Q_{\mu_{ext}, \sg_{ext}^{2}} \RC}{\EE} \mathds{1}\LB \Bigl( \langle \vb_{ext}, \bpsi \rangle + y b_{ext} \Bigr) \leq 0 \RB \\ 
        & \quad \quad \quad \quad \quad \quad \quad \quad \quad \quad \quad \times \prod_{i=1}^{n} \LB \underset{(\cb_{i}, \be_{i}) \sim \LC Q_{\cb_{i_{0}}, \Ib_{i} \varep_{i}^{2}} \times Q_{k_{i}, \tau_{i}} \RC}{\EE} \bar{K}(\cb_{i}, \xb, \be_{i}) \RB  \textrm{.} 
    \end{split} 
\end{align}
Equation~\eqref{eq:perte_Q_exp_generale_clf_gn_search_1} comes from the fact that the predictors' parameters $\LC \vb_{i}, b_{i} \RC_{i=1}^{n}$ and their corresponding points of interest $\LC \cb_{i} \RC_{i=1}^{n}$ and locality parameters $\LC \be_{i} \RC_{i=1}^{n}$ are all independent. As demonstrated in Appendix \ref{app:Binary_Linear_clf_gn}, Equations \eqref{eq:A_i_clf_end} and \eqref{eq:A_ext_clf} yield
\[
    \underset{(\vb_{i}, b_{i}) \sim \LC Q_{\wb_{i}, \Ib_{i}} \times Q_{\mu_{i}, \sg_{i}^{2}} \RC}{\EE} \mathds{1}\LB \Bigl( \langle \vb_{i}, \bpsi \rangle + y b_{i} \Bigr) \leq 0 \RB = 1 - \Phi\LP \frac{y (\langle \wb_{i}, \xb \rangle + \mu_{i})}{\sqrt{\sg_{i}^{2} + \LN \xb \RN^{2}}} \RP \textrm{and}
\]
\[
    \underset{(\vb_{ext}, b_{ext}) \sim \LC Q_{\wb_{ext}, \Ib_{ext}} \times Q_{\mu_{ext}, \sg_{ext}^{2}} \RC}{\EE} \mathds{1}\LB \Bigl( \langle \vb_{ext}, \bpsi \rangle + y b_{ext} \Bigr) \leq 0 \RB = 1 - \Phi\LP \frac{y (\langle \wb_{ext}, \xb \rangle + \mu_{ext})}{\sqrt{\sg_{ext}^{2} + \LN \xb \RN^{2}}} \RP \textrm{.}
\]
Thus, Equation~\eqref{eq:perte_Q_exp_generale_clf_gn_search_1} can be rewritten as
\begin{equation}\label{eq:perte_Q_exp_generale_clf_gn_search_2}
    \begin{aligned}
        \ell\Bigl[ Q, (\bpsi, \xb, y) \Bigr] &= \sum_{i=1}^{n} \LB \LC 1 - \Phi\LP \frac{y (\langle \wb_{i}, \xb \rangle + \mu_{i})}{\sqrt{\sg_{i}^{2} + \|\xb\|^{2}}} \RP \RC \times \underset{ (\cb_{i}, \be_{i}) \sim \LC Q_{\cb_{i_{0}}, \Ib_{i} \varep_{i}^{2}} \times Q_{k_{i}, \tau_{i}} \RC }{\EE} K(\cb_{i}, \xb, \be_{i}) \RB \\
        & \quad + \LC 1 - \Phi\LP \frac{y (\langle \wb_{ext}, \xb \rangle + \mu_{ext})}{\sqrt{\sg_{ext}^{2} + \|\xb\|^{2}}} \RP \RC \\ 
        & \quad \quad \quad \quad \quad \quad \quad \quad \quad \quad \quad \times \prod_{i=1}^{n} \LB \underset{(\cb_{i}, \be_{i}) \sim \LC Q_{\cb_{i_{0}}, \Ib_{i} \varep_{i}^{2}} \times Q_{k_{i}, \tau_{i}} \RC}{\EE} \bar{K}(\cb_{i}, \xb, \be_{i}) \RB \textrm{.}
    \end{aligned}
\end{equation}
Note that, applying the expectation on~\eqref{eq:complement_of_hard_treshold_function} yields
\[
    \underset{(\cb_{i}, \be_{i}) \sim \LC Q_{\cb_{i_{0}}, \Ib_{i} \varep_{i}^{2}} \times Q_{k_{i}, \tau_{i}} \RC}{\EE} \bar{K}(\cb_{i}, \xb, \be_{i}) = 1 - \underset{ (\cb_{i}, \be_{i}) \sim \LC Q_{\cb_{i_{0}}, \Ib_{i} \varep_{i}^{2}} \times Q_{k_{i}, \tau_{i}} \RC }{\EE} K(\cb_{i}, \xb, \be_{i}) \textrm{.}
\]
Then, we have
\begin{align}
    \underset{ (\cb_{i}, \be_{i}) \sim \LC Q_{\cb_{i_{0}}, \Ib_{i} \varep_{i}^{2}} \times Q_{k_{i}, \tau_{i}} \RC }{\EE} K(\cb_{i}, \xb, \be_{i}) &= \underset{ \be_{i} \sim Q_{k_{i}, \tau_{i}} }{\EE} \LC \underset{ \cb_{i} \sim Q_{\cb_{i_{0}}, \Ib_{i} \varep_{i}^{2}} }{\EE} K(\cb_{i}, \xb, \be_{i}) \RC \label{eq:expectation_over_hard_treshold_function_1} \\ 
    &= \underset{ \be_{i} \sim Q_{k_{i}, \tau_{i}} }{\EE} \LC \underset{ \cb_{i} \sim \Ncal\LP \cb_{i_{0}}, \Ib_{i} \varep_{i}^{2} \RP }{\EE} \mathds{1}\Bigl[ d(\cb_{i}, \xb) \leq \be_{i} \Bigr] \RC \label{eq:expectation_over_hard_treshold_function_2} \\ 
    &= \underset{ \be_{i} \sim Q_{k_{i}, \tau_{i}} }{\EE} \LC \underset{ \cb_{i} \sim \Ncal\LP \cb_{i_{0}}, \Ib_{i} \varep_{i}^{2} \RP }{\EE} \mathds{1}\Bigl[ \LN \cb_{i} - \xb \RN \leq \be_{i} \Bigr] \RC  \label{eq:expectation_over_hard_treshold_function_3} \\ 
    &= \underset{ \be_{i} \sim Q_{k_{i}, \tau_{i}} }{\EE} \LC \underset{ \ub \sim \Ncal\LP \cb_{i_{0}} - \xb, \Ib_{i} \varep_{i}^{2} \RP }{\EE} \mathds{1}\Bigl[ \LN \ub \RN \leq \be_{i} \Bigr] \RC \label{eq:expectation_over_hard_treshold_function_4} \\ 
    &= \underset{ \be_{i} \sim Q_{k_{i}, \tau_{i}} }{\EE} \LC \underset{ \ub \sim \Ncal\LP \cb_{i_{0}} - \xb, \Ib_{i} \varep_{i}^{2} \RP }{\EE} \mathds{1}\LB \frac{ \LN \ub \RN^{2}}{\varep_{i}^{2}} \leq \frac{\be_{i}^{2}}{\varep_{i}^{2}} \RB \RC \label{eq:expectation_over_hard_treshold_function_5} \\ 
    &= \underset{ \be_{i} \sim Q_{k_{i}, \tau_{i}} }{\EE} \LC \underset{ \ub \sim \Ncal\LP \cb_{i_{0}} - \xb, \Ib_{i} \varep_{i}^{2} \RP }{\EE} \mathds{1}\LB \sum_{r=1}^{d} \frac{u_{r}^{2}}{\varep_{i}^{2}} \leq \frac{\be_{i}^{2}}{\varep_{i}^{2}} \RB \RC \label{eq:expectation_over_hard_treshold_function_6} \\ 
    &= \underset{ \be_{i} \sim Q_{k_{i}, \tau_{i}} }{\EE} \LC P \LP \frac{\be_{i}^{2}}{\varep_{i}^{2}}; d, \frac{\LN \cb_{i_{0}} - \xb \RN^{2}}{\varep_{i}^{2}} \RP \RC \label{eq:expectation_over_hard_treshold_function_7} \\ 
    &\eqdef \Up_{d, \LN \cb_{i_{0}} - \xb \RN, \varep_{i}, k_{i}, \tau_{i}} \label{eq:expectation_over_hard_treshold_function_8} \textrm{,} 
\end{align}
where $P \LP \frac{\be_{i}^{2}}{\varep_{i}^{2}}; d, \frac{\LN \cb_{i_{0}} - \xb \RN^{2}}{\varep_{i}^{2}} \RP$ is the CDF of the noncentral chi-squared distribution $\chi^{2}$ with $d$ degrees of freedom and non-centrality parameter $\frac{\LN \cb_{i_{0}} - \xb \RN^{2}}{\varep_{i}^{2}}$. Here, Equation~\eqref{eq:expectation_over_hard_treshold_function_2} is obtained by replacing $Q_{\cb_{i_{0}}, \Ib_{i} \varep_{i}^{2}}$ and $K(\cb_{i}, \xb, \be_{i})$ by their definitions. Equation~\eqref{eq:expectation_over_hard_treshold_function_3} is derived from the chosen metric $d$ as $d(\cb_{i}, \xb) \eqdef \LN \cb_{i} - \xb \RN$. Equation~\eqref{eq:expectation_over_hard_treshold_function_4} is a consequence of the change of variable $\ub \eqdef \cb_{i} - \xb$. Equation~\eqref{eq:expectation_over_hard_treshold_function_5} is just an equivalent form of Equation~\eqref{eq:expectation_over_hard_treshold_function_4}. Equation~\eqref{eq:expectation_over_hard_treshold_function_6} follows from the definition of $\LN . \RN$ ($\ell2$ norm). Equation~\eqref{eq:expectation_over_hard_treshold_function_7} results from the definition of the noncentral chi-squared distribution $\chi^{2}$ (see \citealp{patnaik1949non}).

Therefore, plugging~\eqref{eq:expectation_over_hard_treshold_function_8} into~\eqref{eq:perte_Q_exp_generale_clf_gn_search_2} yields
\begin{equation}\label{eq:perte_Q_exp_generale_clf_gn_search_3}
    \begin{aligned}
        \ell\Bigl[ Q, (\bpsi, \xb, y) \Bigr] &= \sum_{i=1}^{n} \LB \Up_{d, \LN \cb_{i_{0}} - \xb \RN, \varep_{i}, k_{i}, \tau_{i}} \times \LC 1 - \Phi\LP \frac{y (\langle \wb_{i}, \xb \rangle + \mu_{i})}{\sqrt{\sg_{i}^{2} + \LN \xb \RN^{2}}} \RP \RC \RB \\ 
        & \quad + \prod_{i=1}^{n} \LB 1 - \Up_{d, \LN \cb_{i_{0}} - \xb \RN, \varep_{i}, k_{i}, \tau_{i}} \RB \times \LC 1 - \Phi\LP \frac{y (\langle \wb_{ext}, \xb \rangle + \mu_{ext})}{\sqrt{\sg_{ext}^{2} + \LN \xb \RN^{2}}} \RP \RC \textrm{.} 
    \end{aligned}
\end{equation}

Note that $P \LP \frac{\be_{i}^{2}}{\varep_{i}^{2}}; d, \frac{\LN \cb_{i_{0}} - \xb \RN^{2}}{\varep_{i}^{2}} \RP$ is equal to $1-Q_{\frac{d}{2}}\LP \frac{\LN \cb_{i_{0}} - \xb \RN}{\varep_{i}}, \frac{\be_{i}}{\varep_{i}} \RP$ according to \citet{nuttall1975some}, where $Q_{M}\LP a, b\RP$ is the Marcum Q-function defined as 
\begin{equation}\label{eq:Marcum_Q_series}
    Q_{M}\LP a, b\RP \eqdef 1 - e^{-a^{2}/2} \sum_{s=0}^{\infty} \frac{1}{s!} \LP \frac{a^{2}}{2} \RP^{s} \sum_{j=0}^{\infty} \frac{\LP \frac{b^{2}}{2} \RP^{M + s + j} e^{-\frac{b^{2}}{2}}}{\Gm(M + s + j + 1)} \textrm{.}
\end{equation} 
Then, from~\eqref{eq:expectation_over_hard_treshold_function_8}, we have 
\begin{align}
    \Up_{d, \LN \cb_{i_{0}} - \xb \RN, \varep_{i}, k_{i}, \tau_{i}} &\eqdef \underset{ \be_{i} \sim Q_{k_{i}, \tau_{i}} }{\EE} \LB P \LP \frac{\be_{i}^{2}}{\varep_{i}^{2}}; d, \frac{\LN \cb_{i_{0}} - \xb \RN^{2}}{\varep_{i}^{2}} \RP \RB \label{eq:upsilon_def} \\ 
    &= \underset{ \be_{i} \sim Q_{k_{i}, \tau_{i}} }{\EE} \LB 1-Q_{\frac{d}{2}}\LP \frac{\LN \cb_{i_{0}} - \xb \RN}{\varep_{i}}, \frac{\be_{i}}{\varep_{i}} \RP \RB \label{eq:Nuttal_equivalent} \\ 
    &= e^{- \frac{\LN \cb_{i_{0}} - \xb \RN^{2}}{2\varep_{i}^{2}}} \sum_{s=0}^{\infty} \frac{1}{s!} \LP \frac{\LN \cb_{i_{0}} - \xb \RN^{2}}{2\varep_{i}^{2}} \RP^{s} \sum_{j=0}^{\infty} \frac{\underset{ \be_{i} \sim Q_{k_{i}, \tau_{i}} }{\EE} \LB \LP \frac{\be_{i}^{2}}{2\varep_{i}^{2}} \RP^{\frac{d}{2} + s + j} e^{-\frac{\be_{i}^{2}}{2\varep_{i}^{2}}} \RB}{\Gm\LP \frac{d}{2} + s + j + 1 \RP} \label{eq:monotone_convergence} \\ 
    &= e^{- \frac{\LN \cb_{i_{0}} - \xb \RN^{2}}{2\varep_{i}^{2}}} \sum_{s=0}^{\infty} \frac{1}{s!} \LP \frac{\LN \cb_{i_{0}} - \xb \RN^{2}}{2\varep_{i}^{2}} \RP^{s} \sum_{j=0}^{\infty} \frac{\underset{ \Tilde{\be_{i}} \sim Q_{k_{i}, \sqrt{2}\tau_{i}\varep_{i}} }{\EE} \LB \LP \Tilde{\be_{i}}^{2} \RP^{\frac{d}{2} + s + j} e^{-\Tilde{\be_{i}}^{2}} \RB}{\Gm\LP \frac{d}{2} + s + j + 1 \RP} \label{eq:variable_changement} \textrm{,} 
\end{align}
where $\Tilde{\be_{i}} \eqdef \frac{\be_{i}}{\sqrt{2}\varep_{i}}$. Here, Equation~\eqref{eq:upsilon_def} just defines the quantity $\Up_{d, \LN \cb_{i_{0}} - \xb \RN, \varep_{i}, k_{i}, \tau_{i}}$ and~\eqref{eq:Nuttal_equivalent} is the equivalent form of~\citet{nuttall1975some}. Equation~\eqref{eq:monotone_convergence} is obtained by applying the monotone convergence theorem (see \citealp{bibby1974axiomatisations}) and~\eqref{eq:variable_changement} follows from the change of variable $\Tilde{\be_{i}} \eqdef \frac{\be_{i}}{\sqrt{2}\varep_{i}}$.

Recall that, from \citet{gradshteyn2014table}, Section 3.462, Eq 1, we have
\[
    \int_{0}^{\infty} x^{\nu - 1} \; e^{-a x^{2} - b x} dx = \LP 2 a\RP^{-\nu/2} \Gm(\nu) \exp\LP \frac{b^{2}}{8a} \RP D_{-\nu}\LP \frac{b}{\sqrt{2a}}\RP \textrm{,} \; \LB Re \; a > 0 \textrm{,} \quad Re \; b > 0 \RB \textrm{,}
\]
where $D_{a}\LP x\RP$ denotes the parabolic cylinder function (see \citealp{whittaker1990parabolic}). Consequently, we get 
\[
    \begin{aligned}
        \underset{ \Tilde{\be_{i}} \sim Q_{k_{i}, \sqrt{2}\tau_{i}\varep_{i}} }{\EE} \LB \LP \Tilde{\be_{i}}^{2} \RP^{\frac{d}{2} + s + j} e^{-\Tilde{\be_{i}}^{2}} \RB &= \frac{\LP \sqrt{2}\tau_{i}\varep_{i} \RP^{k_{i}}}{\Gm\LP k_{i} \RP} \int_{0}^{\infty} \Tilde{\be_{i}}^{d + 2s + 2j + k_{i} - 1} \; e^{-\Tilde{\be_{i}}^{2} - \sqrt{2}\tau_{i}\varep_{i} \Tilde{\be_{i}}} d\Tilde{\be_{i}} \\
        &= \LB \frac{\LP \sqrt{2}\tau_{i}\varep_{i} \RP^{k_{i}}}{\Gm\LP k_{i} \RP} \RB 2^{-\frac{\LP d + 2s + 2j + k_{i} \RP}{2}} e^{\frac{\tau_{i}^{2} \varep_{i}^{2}}{4}} \\ 
        & \quad \quad \quad \times \Gm\LP d + 2s + 2j + k_{i} \RP D_{-\LP d + 2s + 2j + k_{i} \RP}\LP \tau_{i} \varep_{i} \RP \textrm{.}
    \end{aligned}
\]
Therefore, plugging this result into~\eqref{eq:variable_changement} finally yields
\begin{equation}\label{eq:upsilon_final}
    \begin{aligned}
        \Up_{d, \LN \cb_{i_{0}} - \xb \RN, \varep_{i}, k_{i}, \tau_{i}} &= \LB \frac{\LP \tau_{i}\varep_{i} \RP^{k_{i}}}{\Gm\LP k_{i} \RP} \RB e^{ \frac{\tau_{i}^{2} \varep_{i}^{2}}{4} - \frac{\LN \cb_{i_{0}} - \xb \RN^{2}}{2\varep_{i}^{2}} }  \\ 
        & \quad \times \sum_{s=0}^{\infty} \frac{1}{s!} \LP \frac{\LN \cb_{i_{0}} - \xb \RN^{2}}{2\varep_{i}^{2}} \RP^{s} \sum_{j=0}^{\infty} \frac{\Gm\LP d + 2s + 2j + k_{i} \RP D_{-\LP d + 2s + 2j + k_{i} \RP}\LP \tau_{i} \varep_{i} \RP}{2^{\frac{d}{2} + s + j} \Gm\LP \frac{d}{2} + s + j + 1 \RP} \textrm{.}
    \end{aligned}
\end{equation}
Notice that $\Up_{d, \LN \cb_{i_{0}} - \xb \RN, \varep_{i}, k_{i}, \tau_{i}}$ represents the probability that the instance $\xb$ will be situated within a ball centered at the point $\cb_{i_{0}}$ with some variance $\varep_{i}^{2}$ and the unknown ball radius is expressed in terms of $\varep_{i}$, $k_{i}$ and $\tau_{i}$. However, due to the limitations in computational precision of~\eqref{eq:upsilon_final}, we use the Quasi-Monte Carlo method to compute an approximation to $\Up_{d, \LN \cb_{i_{0}} - \xb \RN, \varep_{i}, k_{i}, \tau_{i}}$ directly from~\eqref{eq:upsilon_def}. As explained in Appendix~\ref{app:Binary_Linear_clf_gn_search}, with only $60$ $\beta_{i}$ drawn from the $\Gamma\LP k_{i}, \tau_{i} \RP$ distribution, we are able to achieve an accurate approximation.

Then, the loss $\ell\Bigl[ Q, (\bpsi, \xb, y) \Bigr]$ of~\eqref{eq:perte_Q_exp_generale_clf_gn_search_3} finally becomes
\begin{equation}\label{eq:perte_Q_exp_generale_clf_gn_search_4}
    \begin{aligned}
        \ell\Bigl[ Q, (\bpsi, \xb, y) \Bigr] &= \sum_{i=1}^{n} \LB \Up_{d, \LN \cb_{i_{0}} - \xb \RN, \varep_{i}, k_{i}, \tau_{i}} \times \LC 1 - \Phi\LP \frac{y (\langle \wb_{i}, \xb \rangle + \mu_{i})}{\sqrt{\sg_{i}^{2} + \LN \xb \RN^{2}}} \RP \RC \RB \\ 
        & \quad + \prod_{i=1}^{n} \LB 1 - \Up_{d, \LN \cb_{i_{0}} - \xb \RN, \varep_{i}, k_{i}, \tau_{i}} \RB \times \LC 1 - \Phi\LP \frac{y (\langle \wb_{ext}, \xb \rangle + \mu_{ext})}{\sqrt{\sg_{ext}^{2} + \LN \xb \RN^{2}}} \RP \RC \textrm{,} 
    \end{aligned}
\end{equation}
where $\Up_{d, \LN \cb_{i_{0}} - \xb \RN, \varep_{i}, k_{i}, \tau_{i}}$ is given by~\eqref{eq:upsilon_final}.

This expression of $\ell\Bigl[ Q, (\bpsi, \xb, y) \Bigr]$ given by~\eqref{eq:perte_Q_exp_generale_clf_gn_search_4} allows us to deduce the expression of the true risk $L_{\mathcal{D}}(Q)$ and the empirical risk $L_{S}(Q)$ given by~\eqref{eq:risque_généralisation_forme_espérance} and~\eqref{eq:nouveau_risque_généralisation_forme_empirique}. Combining the empirical risk $L_{S}(Q)$ with the obtained closed-form expression of the $KL$ divergence, we can then perform the PAC-Bayesian learning algorithm, which now becomes
\begin{equation}\label{eq:clf_gn_bound_minimize_form_search}
    \mathbf{PB^{\lambda}}(S) = \underset{\LC \cb_{i_{0}}, \varep_{i}, \wb_{i}, \mu_{i}, \sg_{i}, k_{i}, \tau_{i} \RC_{i=1}^{n}, \wb_{ext}, \mu_{ext}, \sg_{ext}}{\mathrm{argmin}} L_{S}(Q) + \frac{1}{\lambda} KL_{cl}^{'}(Q||P) \; \textrm{.}
\end{equation}
As in the previous subsections, the minimization objective is non-convex in the parameters and, consequently, requires several random restarts to obtain a good solution.

\subsection{Linear Regression} \label{Linear_Regression_reg_search_gn}
Similarly to the Subsection \ref{Binary_Linear_clf_gn_search}, for the linear regression problem we obtain the following global $KL$-divergence 

\begin{equation}\label{eq:kl_reg_search_gn_final}
    KL_{reg}^{'}\left(Q || P\right) = KL_{reg}\left(Q || P\right) - \sum_{i=1}^{n} \ln{\Bigl( \varep_{i}^{d} \Bigr)} + \frac{1}{2} \sum_{i=1}^{n} \Bigl( d \varep_{i}^{2} - d \Bigr) + \frac{1}{2} \sum_{i=1}^{n} \left \| \cb_{i_{0}} \right \|^{2} \textrm{,}
\end{equation}
where $KL_{reg}\left(Q || P\right)$ is given by~\eqref{eq:kl_reg_gn_final}.

As discussed in Appendix \ref{app:Linear_Regression_reg_gn}, the loss $\ell\Bigl[ Q, (\xb, y) \Bigr]$ of posterior $Q$ on example $(\xb,y)$ is now given by
\begin{equation}\label{eq:perte_Q_exp_generale_reg_gn_search}
    \begin{aligned}
        \ell\Bigl[ Q, (\xb, y) \Bigr] &= \sum_{i=1}^{n} \LB \Up_{d, \LN \cb_{i_{0}} - \xb \RN, \varep_{i}, k_{i}, \tau_{i}} \times \LC \LN \xb \RN^{2} \rho_{i}^{2} + \sg_{i}^{2} + \Bigl( \langle \wb_{i}, \xb \rangle + \mu_{i} - y \Bigr)^{2} \RC \RB \\ 
        & \quad + \prod_{i=1}^{n} \LB 1 - \Up_{d, \LN \cb_{i_{0}} - \xb \RN, \varep_{i}, k_{i}, \tau_{i}} \RB \\ 
        & \quad \quad \quad \quad \quad \quad \quad \times \LC \LN \xb \RN^{2} \rho_{ext}^{2} + \sg_{ext}^{2} + \Bigl( \langle \wb_{ext}, \xb \rangle + \mu_{ext} - y \Bigr)^{2} \RC \textrm{.} 
    \end{aligned}
\end{equation}

The expression of $\ell\Bigl[ Q, (\mathbf{x}, y) \Bigr]$ of~\eqref{eq:perte_Q_exp_generale_reg_gn_search} yields an expression of the true risk $L_{\mathcal{D}}(Q)$ and of the empirical risk $L_{S}(Q)$ given by~\eqref{eq:risque_généralisation_forme_espérance} and~\eqref{eq:risque_généralisation_forme_empirique}. Combining that empirical risk $L_{S}(Q)$ with the obtained closed-form expression of the $KL$ divergence, we can then perform the PAC-Bayesian learning algorithm, which now becomes
\begin{equation}\label{eq:reg_gn_bound_minimize_form_search}
    \mathbf{PB^{\lambda}}(S) = \underset{\LC \cb_{i_{0}}, \varep_{i}, \wb_{i}, \rho_{i}, \mu_{i}, \sg_{i}, k_{i}, \tau_{i} \RC_{i=1}^{n}, \wb_{ext}, \rho_{ext}, \mu_{ext}, \sg_{ext}}{\mathrm{argmin}}  L_{\Scal}(Q) + \frac{1}{\lambda} KL_{reg}^{'}(Q||P) \; \textrm{.}
\end{equation}
As in the Subsection~\ref{Binary_Linear_clf_gn_search}, the minimization objective is non convex in the parameters. Hence, several restarts at randomly selected positions are needed to explore the parameter space with some success.

\subsection{Model Training}
\subsubsection{Non-Convex Problems and Softwares}
Unlike the convex optimization problems, non-convex problems do not guarantee a convergence to a global minimum due to the presence of multiple local minima, saddle points, and regions of flat curvature during their optimization process. This has led in literature, to the development of specialized optimization algorithms aimed at improving convergence to useful minima in non-convex settings. Optimization techniques such as stochastic gradient descent (SGD) and its adaptive variants like Adam \citep{kingma2014adam}, AdaGrad \citep{JMLR:v12:duchi11a}, NAdam \citep{dozat2016incorporating} and RMSProp \citep{tieleman2012lecture} are some examples. In this paper, we will simply use the NAdam optimization technique with a number of random restarts to solve optimization problems \eqref{eq:clf_gn_bound_minimize_form}, \eqref{eq:reg_gn_bound_minimize_form}, \eqref{eq:clf_gn_bound_minimize_form_search}, and \eqref{eq:reg_gn_bound_minimize_form_search}.

\subsubsection{Randomly Splitting of Data} For learning purpose, each data set is randomly partitioned into a training set, a validation set and a test set; with ratios of $70\%$, $15\%$ and $15\%$, respectively. The training set is used for learning the model, while the validation set is for tuning the regularization hyperparameter $\lambda$. The test set serves to evaluate the performance of the model. As many practitioners have observed that the best $\lambda$ would generally be $\lambda^{'} \times m\_train$, where $m\_train$ is the training set sample size and $\lambda^{'}$ some unknown constant; by following this concept, tuning the regularization hyperparameter $\lambda$ is equal of seeking the best value of the constant $\lambda^{'}$. Our best value of the constant $\lambda^{'}$ will be chosen from the set of values $\{1, 1.25, 1.5, 1.75, 2, 2.25, 2.5, 2.75, 3, 3.25, 3.5, 3.75, 4, 4.25, 4.5, 4.75, 5\}$. 

\subsubsection{Choice of the Distance Metric}
The correct choice of the distance metric $d(\cb, \xb)$ used in the vicinity function $K(\cb,$ $\xb,$ $ \be)$ is crucial, because it will determine the shape of the locality that will be used for a predictor. More specifically, the metric $d(\cb, \xb)$ affects only the loss function of the \textit{posterior} $Q$ through the vicinity function $K(\cb, \xb, \be)$, and does not affect the KL divergence. Moreover, any metric $d(\cb, \xb)$ could be used for the computation of $\ell\Bigl[ Q, (\bpsi, \xb, y) \Bigr]$ and $\ell\Bigl[ Q, (\xb, y) \Bigr]$ when the points of interests are given (Section \ref{section_given_gn}). However, this is not the case when the points of interest must be found (Section \ref{section_search_gn}) where the provided functions $\ell\Bigl[ Q, (\bpsi, \xb, y) \Bigr]$ and $\ell\Bigl[ Q, (\xb, y) \Bigr]$ are valid only for the Euclidean metric. In all our experiments below, the Euclidean norm was used.

\subsubsection{Total Number of Executions of the PAC-Bayesian Learning Algorithm}\label{Total_Number_of_Executions}
Note that as the number $n$ of points of interest increases, the number of local minima increases. Therefore, for any random split of data into train-valid-test sets, a given value of $n$ and a given value of $\lambda$ (meaning here $\lambda^{'}$), we do $10 \times n$ random restarts of the algorithm and select the best solution based on the numeric values of Equations~\eqref{eq:clf_gn_bound_minimize_form}, \eqref{eq:reg_gn_bound_minimize_form}, \eqref{eq:clf_gn_bound_minimize_form_search} and \eqref{eq:reg_gn_bound_minimize_form_search}. So, with a set containing $17$ values of $\lambda$, we need $17 \times 10 \times n = 170 \times n$ executions of the algorithm to complete the learning task. The best value of $\lambda$ among the $17$ values is chosen based on the performance of the deterministic predictors on the validation set.

\subsubsection{Comparison}
For comparison purposes, we will use the well-known support vector machines (SVMs) and support vector regression (SVR), both with linear and Gaussian kernels. Note that all implementations were developed in Jupyter notebook\footnote{Jupyter notebook project page, https://jupyter.org, accessed on September 10, 2024, version 7.2.2} with Python\footnote{Python project page, https://www.python.org, accessed on September 10, 2024, version 3.10.15} and Pytorch\footnote{Pytorch project page, https://pytorch.org, accessed on May 15, 2024, version 2.2.2}. The project of mixtures of transparent local models (MoTLM) and its requirements package files can be found on GitHub\footnote{MoTLM project page, https://github.com/ncod140/MoTLM}.

\section{Experiments}\label{Experiments}
In this section, we will first conduct experiments on synthetic data sets to validate whether the algorithms produce what they are supposed to give in idealized situations. Then, we move on to experiments on real data sets from public databases. These last experiments aim to validate the usefulness of the proposed algorithms in comparison with state-of-the-art learning algorithms that produce opaque predictors. Note that the results reported in this document are the average of $10$ reproductions of Section \ref{Total_Number_of_Executions}.

\subsection{Experiments with Known Points of Interest}
This section deals with the scenario in which the points of interest $\LC \cb_{i} \RC_{i=1}^{n}$ are known. We start with the case of binary linear classification followed by that of linear regression.

\subsubsection{Binary Linear Classification}
\begin{figure}[H]
    \begin{subfigure}[b]{0.48\textwidth}
        \includegraphics[scale=0.0899]{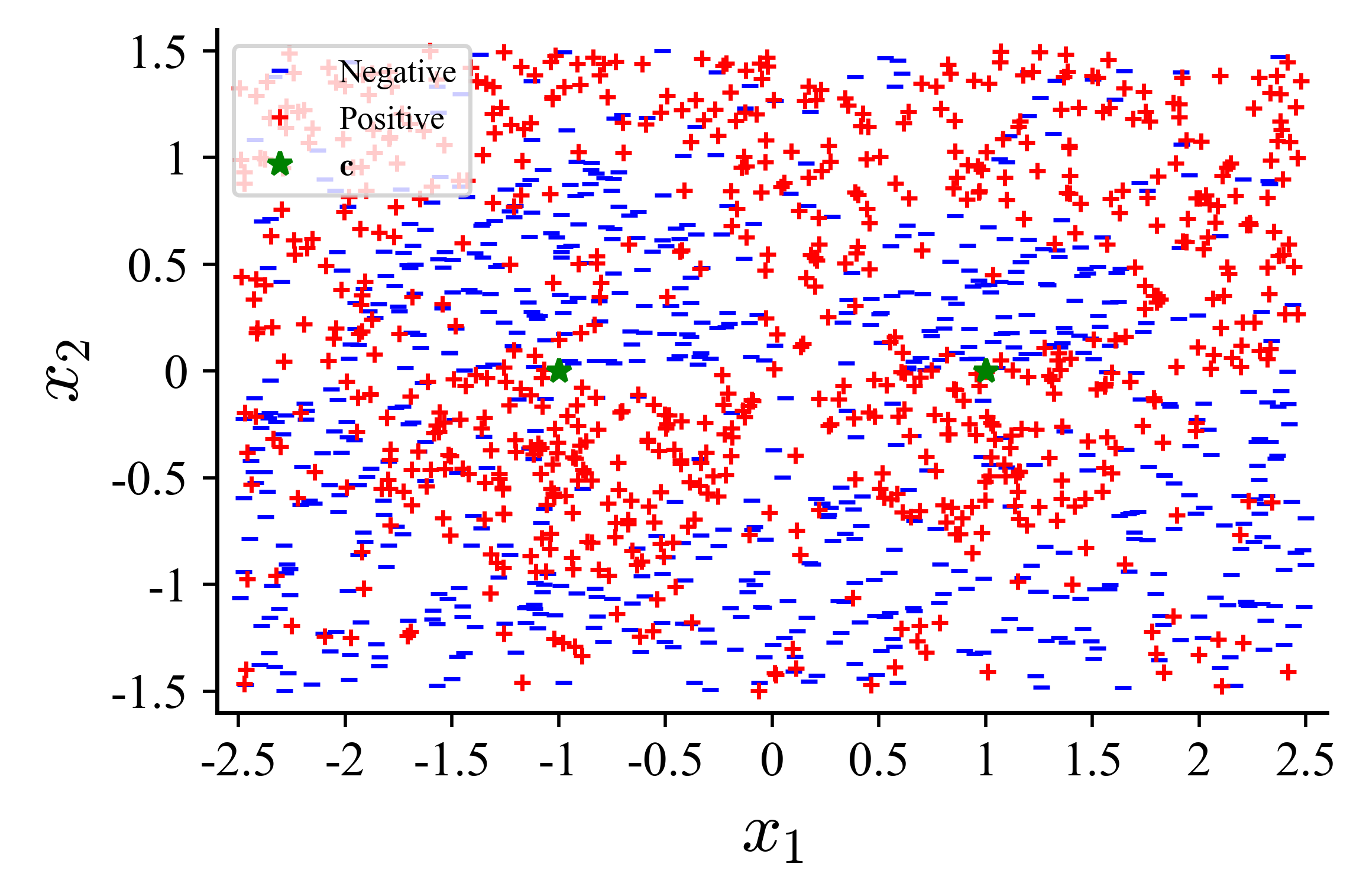}
        \caption{Synthetic data for Case $1$.}
        \label{fig:img_synthetic_data_Case 1_clf_gn_given}
    \end{subfigure}
    \hfill
    \begin{subfigure}[b]{0.48\textwidth}
        \includegraphics[scale=0.0899]{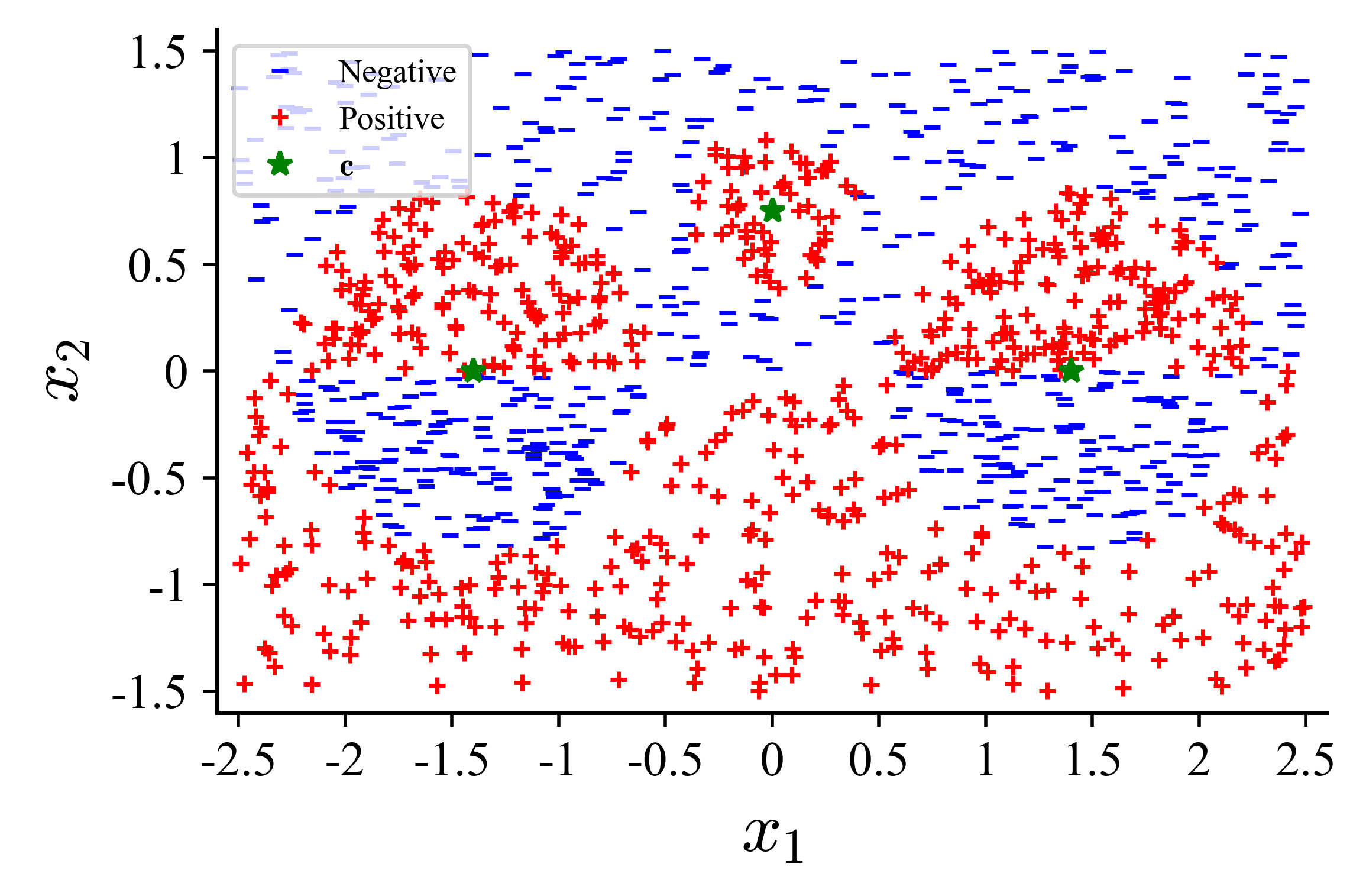}
        \caption{Synthetic data for Case $2$.}
        \label{fig:img_synthetic_data_Case 2_clf_gn_given}
    \end{subfigure}
    \caption{Synthetic data sets for binary linear classification with known points of interest.}
    \label{fig:Linear_Binary_clf_gn_given_synthetic_data}
\end{figure}

Figure \ref{fig:Linear_Binary_clf_gn_given_synthetic_data} shows the scatter plots of two synthetic data sets composed of $1062$ and $1277$ labeled examples, respectively. The points of interest are marked with green stars. 
Here, a classical linear classifier would not be suitable to classify these observations. Nevertheless, we can observe that it would be more interesting to learn a \textit{mixture of transparent local linear classifiers} in the neighborhoods of the known points of interest $\mathbf{c}_{i}$ ($i \in [2]$ for Figure \ref{fig:img_synthetic_data_Case 1_clf_gn_given} and $i \in [3]$ for Figure \ref{fig:img_synthetic_data_Case 2_clf_gn_given}).

\begin{figure}[H]
   \begin{subfigure}[b]{0.48\textwidth}
        \includegraphics[scale=0.0899]{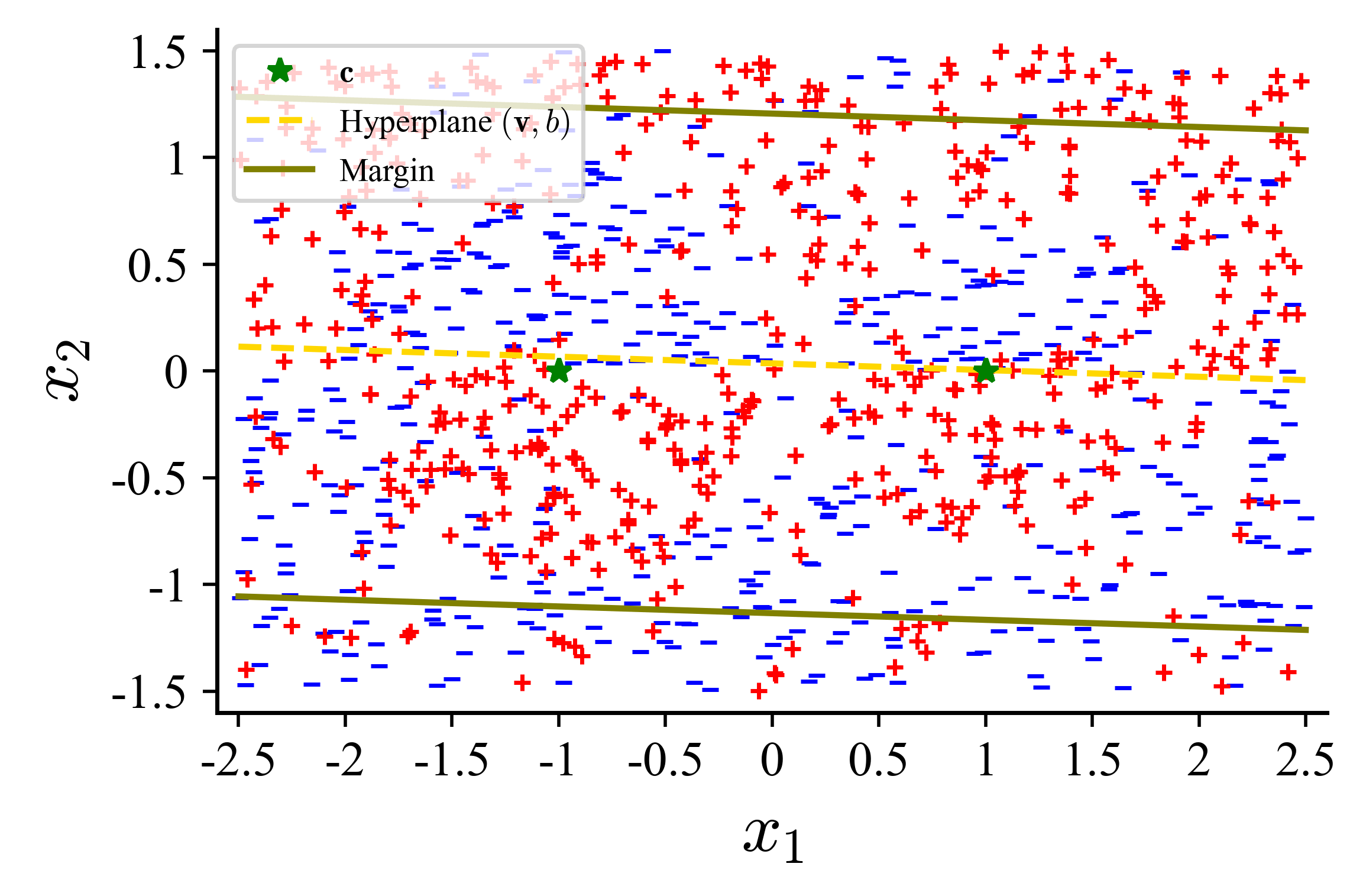}
        \caption{Separating hyperplane in Case $1$.}
    \end{subfigure}
    \hfill
    \begin{subfigure}[b]{0.48\textwidth}
        \includegraphics[scale=0.0899]{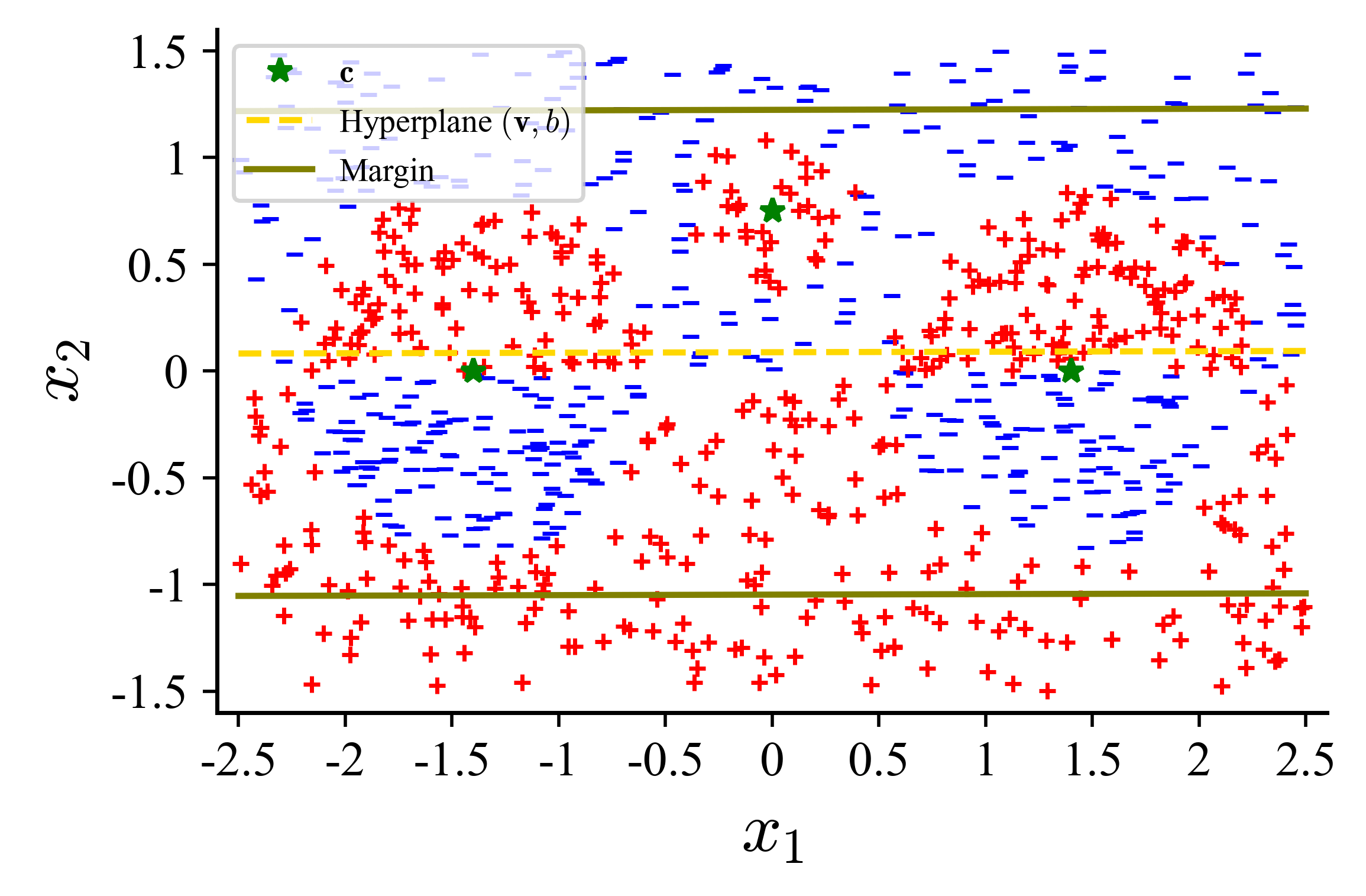}
        \caption{Separating hyperplane in Case $2$.}
    \end{subfigure}
    \caption{Separating hyperplanes obtained using the linear SVM on the training set.}
    \label{fig: img_visualize_hyperplane_global_model_Case 1_2_Jeu d'entraînement_clf_gn}
\end{figure}

\begin{figure}[H]
    \centering
    \begin{subfigure}[b]{0.48\textwidth}
        \centering
        \includegraphics[scale=0.044]{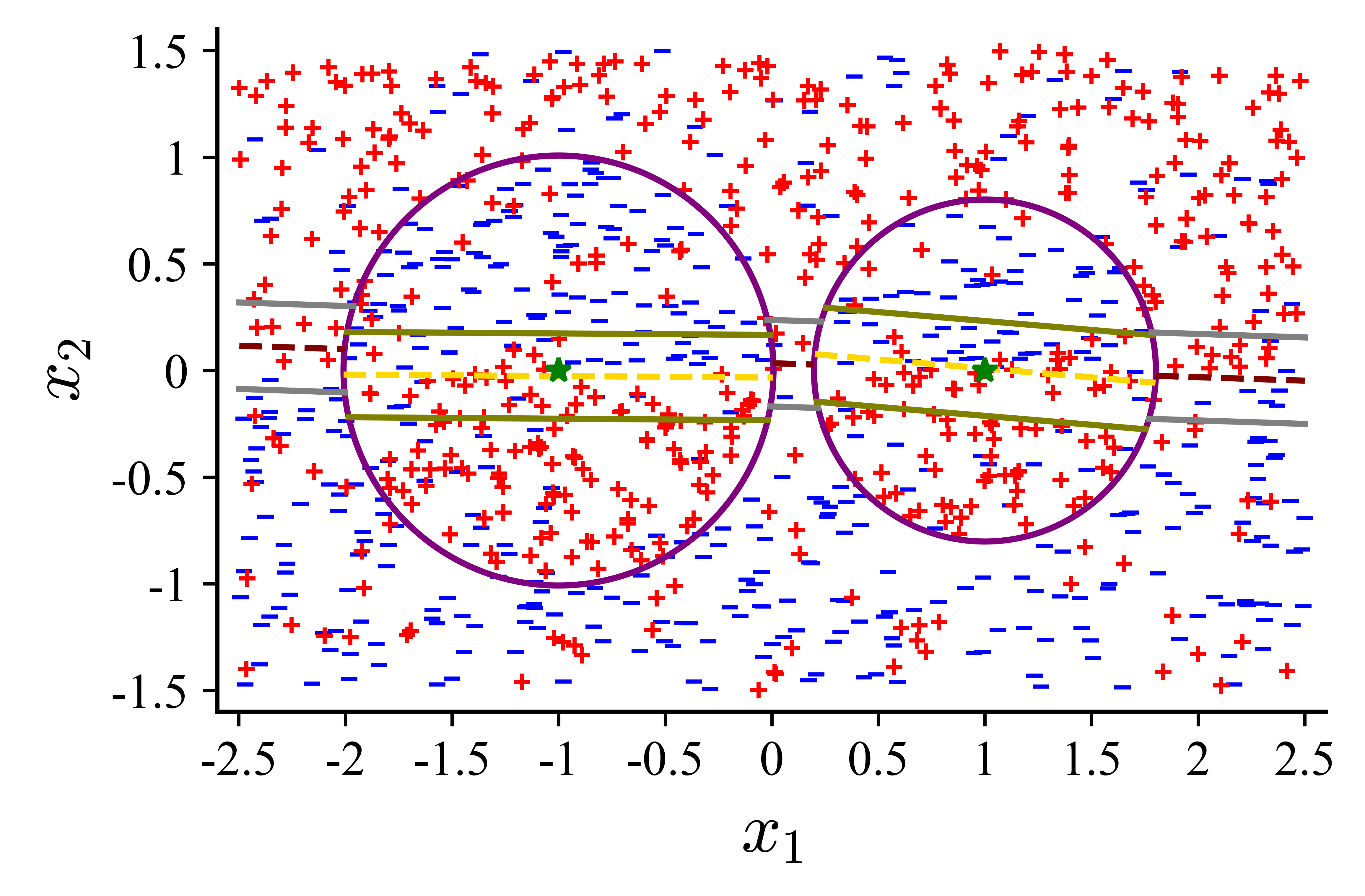}
        \caption{Results obtained via $\mathbf{PB^{\lambda}}(S)$ in Case $1$.}
        \label{fig:img_visualize_hyperplane_hybrid_model_Case 1_Jeu d'entraînement_clf_gn_given}
    \end{subfigure}
    \hfill
    \begin{subfigure}[b]{0.48\textwidth}
        \centering
        \includegraphics[scale=0.044]{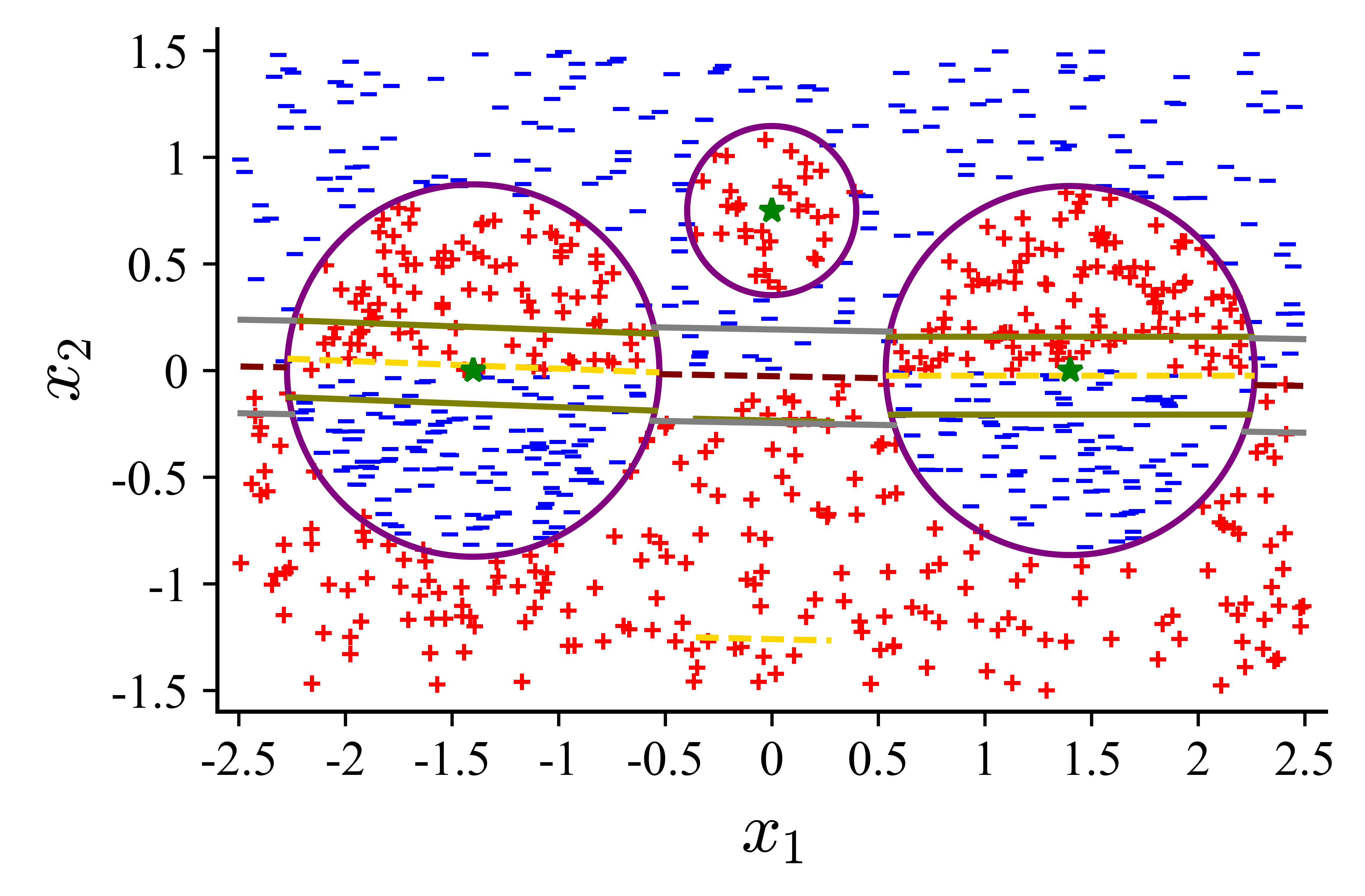}
        \caption{Results obtained via $\mathbf{PB^{\lambda}}(S)$ in Case $2$.}
        \label{fig:img_visualize_hyperplane_hybrid_model_Case 2_Jeu d'entraînement_clf_gn_given}
    \end{subfigure}
    \hfill
    \begin{subfigure}[b]{\textwidth}
    \begin{subfigure}[b]{\textwidth}
        \centering
        \vspace{-51mm}
        \includegraphics[scale=0.0705]{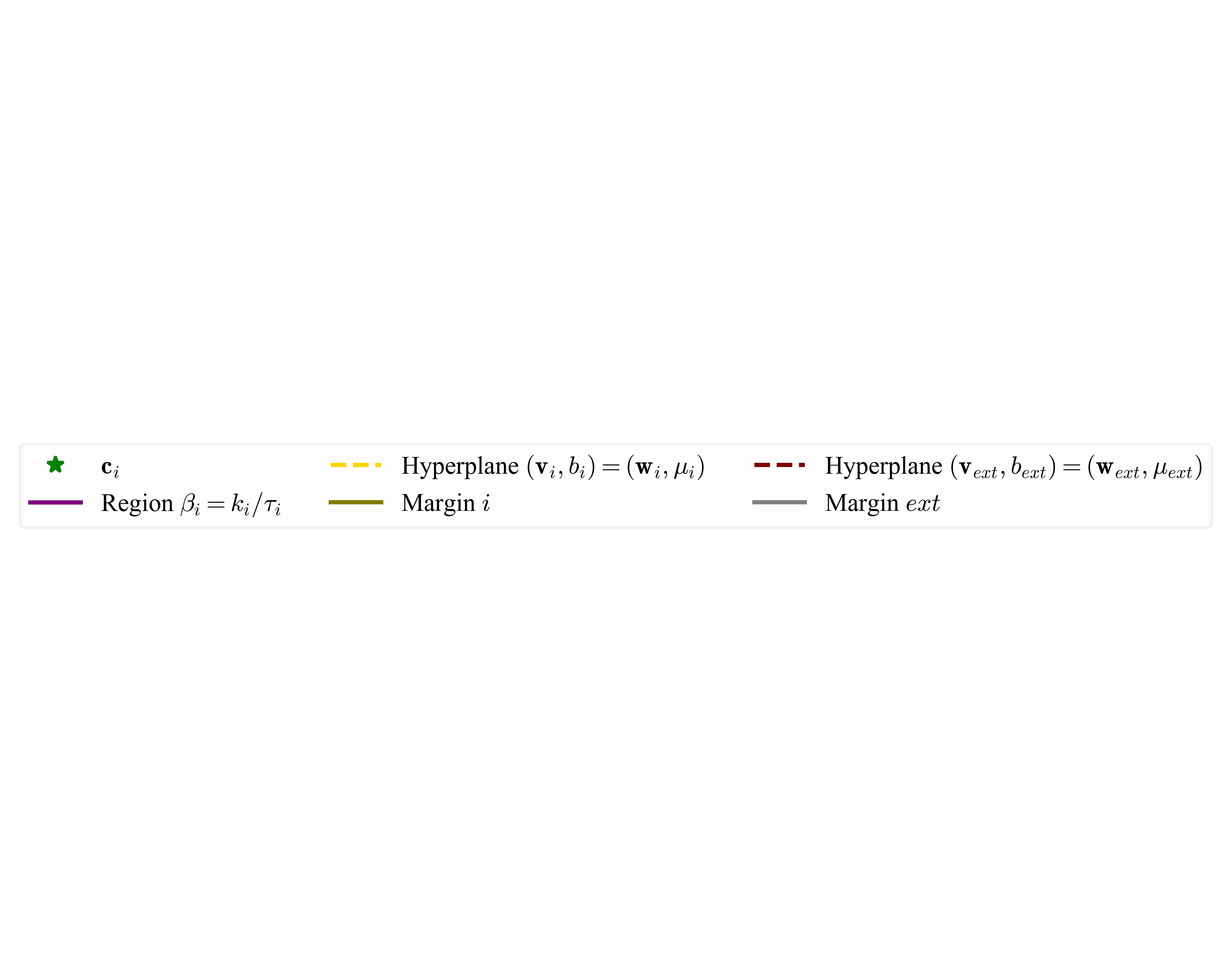}
    \end{subfigure}
    \end{subfigure}
    \vspace{-58mm}
    \caption{Results of mixtures of transparent local linear classifiers on the training set, with known points of interest.}
    \label{fig:img_visualize_hyperplane_hybrid_model_Case 1_2_Jeu d'entraînement_clf_gn_given}
\end{figure}

After learning the linear SVM, the separating hyperplanes obtained on the training set
are illustrated in Figure \ref{fig: img_visualize_hyperplane_global_model_Case 1_2_Jeu d'entraînement_clf_gn}.
As expected, we clearly see that a linear SVM is not the best choice for these tasks. On the other hand, after learning the \textit{mixtures of transparent local linear classifiers}, the predictions made from the averaged deterministic predictors, are more attractive, as shown in Figure ~\ref{fig:img_visualize_hyperplane_hybrid_model_Case 1_2_Jeu d'entraînement_clf_gn_given}.

According to Figure \ref{fig:img_visualize_hyperplane_hybrid_model_Case 1_2_Jeu d'entraînement_clf_gn_given}, on these synthetic data sets in which contamination in areas of interest is limited or absent, the algorithm $\mathbf{PB^{\lambda}}(S)$ not only identifies interesting regions (which are circles) but also produces local separating hyperplanes that are better adapted to them. So we can confirm that the algorithm $\mathbf{PB^{\lambda}}(S)$ from~\eqref{eq:clf_gn_bound_minimize_form} works when the points of interest are known and the localities are circles centered on these points.

Table \ref{tab:accuracy_on_synth_data_clf_gn_given} summarizes for each classifier the average accuracy (Avg acc) (\%) accompanied by their standard deviations (Std dev). The highest scores for the averaged accuracy are shown in bold.
\begin{table}[H]
\centering
\begin{tabular}{ll|cc|cc|cc}
\toprule
\multicolumn{2}{c|}{\multirow{2}{*}{Data set}} & \multicolumn{2}{c}{Linear SVM} & \multicolumn{2}{c}{Gaussian SVM} & \multicolumn{2}{c}{Our classifier} \\
\cline{3-8}
\multicolumn{2}{c|}{} & Avg acc & Std dev & Avg acc & Std dev & Avg acc & Std dev \\
\midrule
\multirow{3}{*}{Case 1} & Training set   & 53.22 & 0.92 & 76.30 & 1.07 & \textbf{78.66} & 0.69  \\
                        & Validation set & 54.93 & 3.29 & 75.69 & 1.99 & \textbf{80.53} & 1.50  \\
                        & Test set       & 54.31 & 2.10 & 73.97 & 1.86 & \textbf{78.23} & 2.05  \\ 
\midrule
\multirow{3}{*}{Case 2} & Training set   & 49.35 & 1.33 & 95.83 & 0.51 & \textbf{97.66} & 0.38  \\
                        & Validation set & 50.05 & 3.76 & 94.97 & 1.73 & \textbf{98.02} & 0.96  \\
                        & Test set       & 50.64 & 2.52 & 93.80 & 1.63 & \textbf{96.69} & 1.21  \\
\bottomrule
\end{tabular}
\caption{Averaged accuracy and standard deviations of classifiers on synthetic data sets with known points of interest.}
\label{tab:accuracy_on_synth_data_clf_gn_given}
\end{table}
Although both our classifier and the linear SVM are interpretable, we have a very significant performance gain from our classifier. Moreover, we can see that our classifier does not sacrifice performance to gain interpretability since it performs relatively well compared to the Gaussian SVM (which is an opaque model).

In Table \ref{tab:risk_bounds_on_synth_data_clf_gn_given}, we present the empirical risk $L_{S}(Q)$ and the bound ``core component'' $L_{S}(Q) + \frac{1}{\lambda} KL_{cl}(Q||P)$ of Corollary~\ref{absolute_bound}, of the training set. In both synthetic data sets, we notice that the bound core component is tight, as it is less than $1.5$ times the $L_{S}(Q)$.

\begin{table}[H]
\centering
\begin{tabular}{lcc}
\toprule
Data set & $L_{S}(Q)$ & Bound core component \\
\midrule
\multirow{1}{*}{Case 1} & 0.2918 & 0.3236  \\
\midrule
\multirow{1}{*}{Case 2} & 0.1592 & 0.2320  \\
\bottomrule
\end{tabular}
\caption{$L_{S}(Q)$ and the bound core component from~\eqref{eq:nouveau_risque_généralisation_forme_empirique} and \eqref{eq:clf_gn_bound_minimize_form}, of our classifier on the training set, with known points of interest.}
\label{tab:risk_bounds_on_synth_data_clf_gn_given}
\end{table}

\begin{figure}[H]
   \begin{subfigure}[b]{0.48\textwidth}
        \includegraphics[scale=0.92] {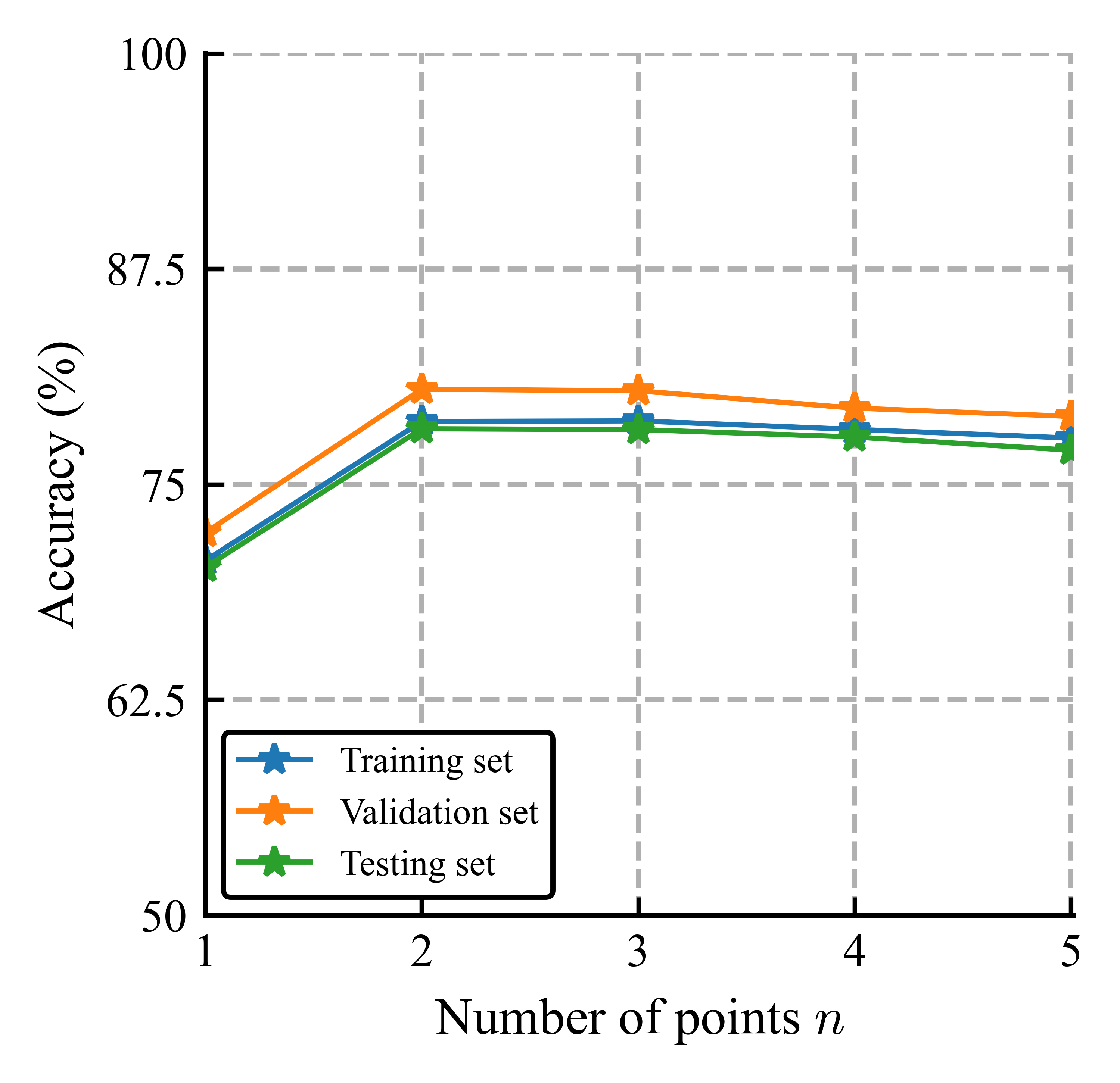}
        \caption{Influence of $n$ on the accuracy.}
        \label{fig:img_influence_of_n_Case 1_Accuracy_clf_gn_given}
    \end{subfigure}
    \hfill
    \begin{subfigure}[b]{0.48\textwidth}
        \includegraphics[scale=0.92]{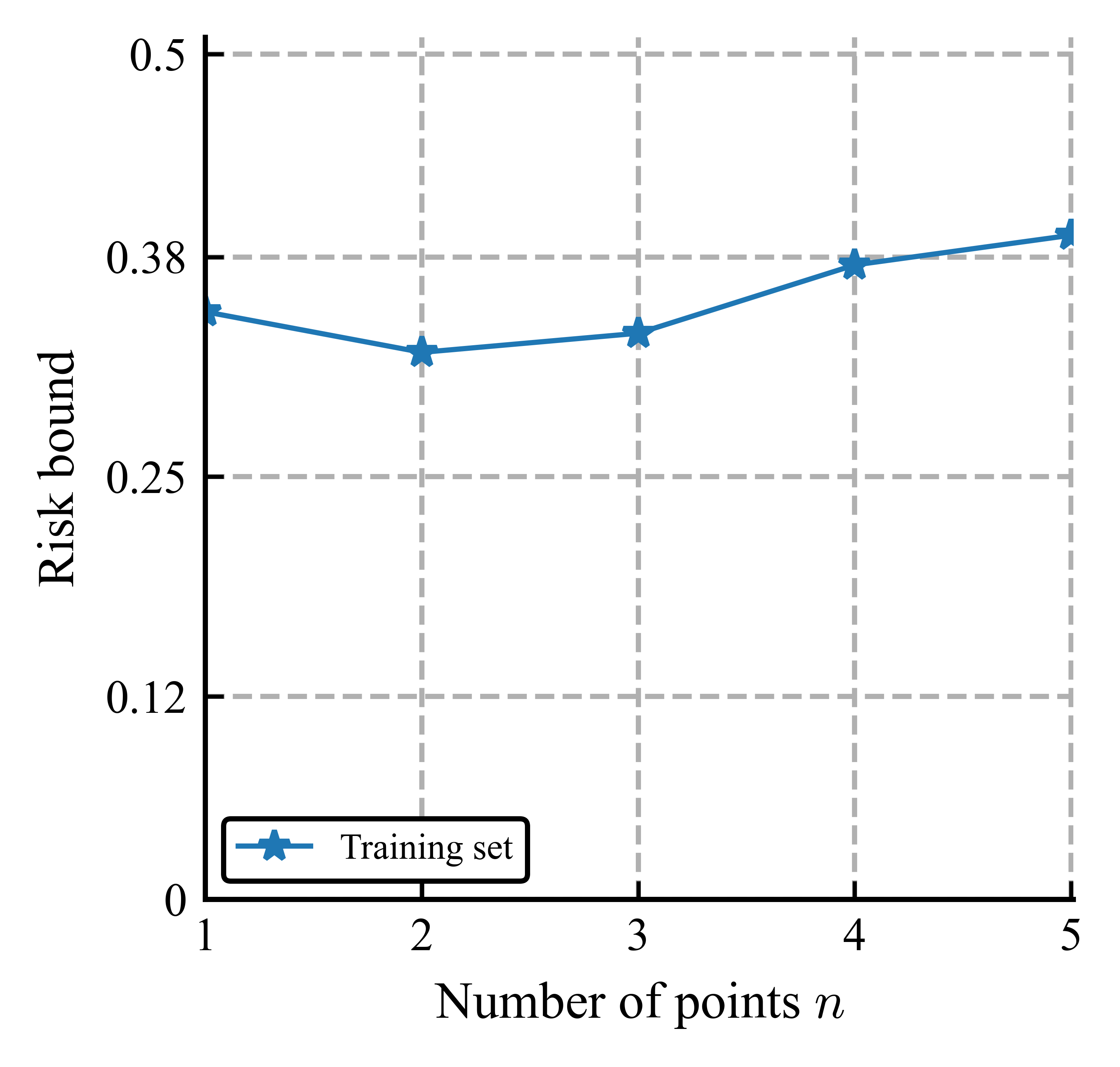}
        \caption{Influence of $n$ on the core component of the risk bound.}
        \label{fig:img_influence_of_n_Case 1_Bound_clf_gn_given}
    \end{subfigure}
    \caption{Influence of the number of points of interest $n$ on our classifier performance in Case $1$, with known points of interest.}
    \label{fig:img_influence_of_n_Case 1_2_Bound_clf_gn_given}
\end{figure}

Now, we will examine the influence of the number of points of interest $n$ on our model performance. To do this, let us focus only on the Case $1$ and vary $n$ from $1$ to $5$ while choosing the set $\bm{c} = \{\mathbf{c}_{i}\}_{i=1}^{5} = \{(-1, 0), (1, 0), (-2, -1), (2, 1), (-1, 0.5)\}$. Then, the accuracy and the risk bound core component (Equation~\ref{eq:clf_gn_bound_minimize_form}) obtained for each value of $n$ are presented in Figure \ref{fig:img_influence_of_n_Case 1_2_Bound_clf_gn_given}. 

The Figure \ref{fig:img_influence_of_n_Case 1_2_Bound_clf_gn_given} reveals that choosing $n=1$ leads to the worst performance in terms of accuracy and the risk bound core component. As for the accuracy obtained for values of $n>1$, they seem to be comparable. However, the Figure \ref{fig:img_influence_of_n_Case 1_Bound_clf_gn_given} reveals that the value of $n=2$ is the one with the best guarantee on the true risk (ie.., having the lowest risk). So, Figure \ref{fig:img_influence_of_n_Case 1_2_Bound_clf_gn_given} suggests that the addition of unnecessary points of interest has negligible impact on our classifier's performance and actually increases the value of its risk bound core component.

\subsubsection{Linear Regression}
\begin{figure}[H]
    \begin{subfigure}[b]{0.48\textwidth}
        \includegraphics[scale=0.62]{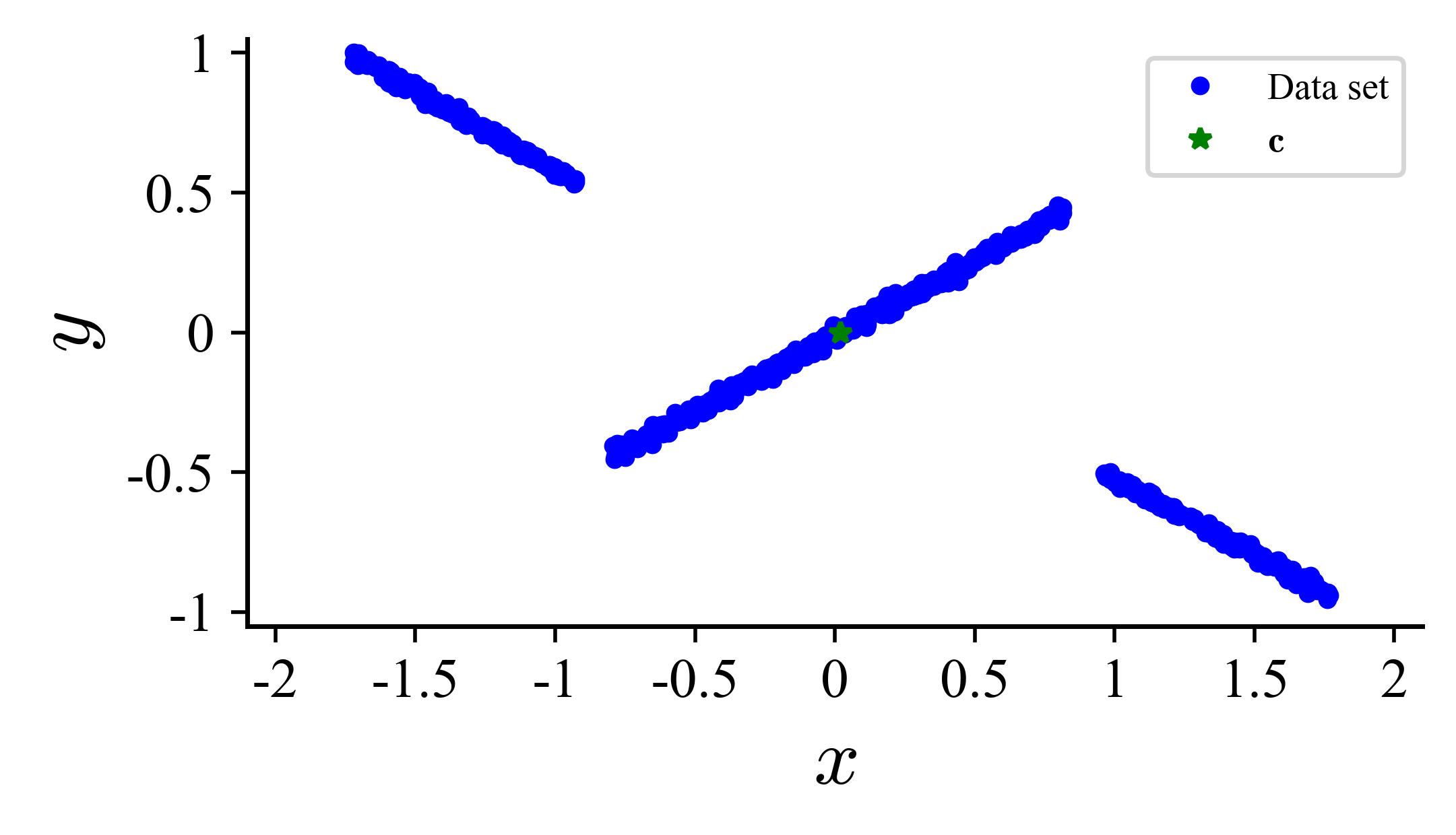}
        \caption{Synthetic data for Case $1$.}
        \label{fig:img_synthetic_data_Case 1_reg_gn_given}
    \end{subfigure}
    \hfill
    \begin{subfigure}[b]{0.48\textwidth}
        \includegraphics[scale=0.62]{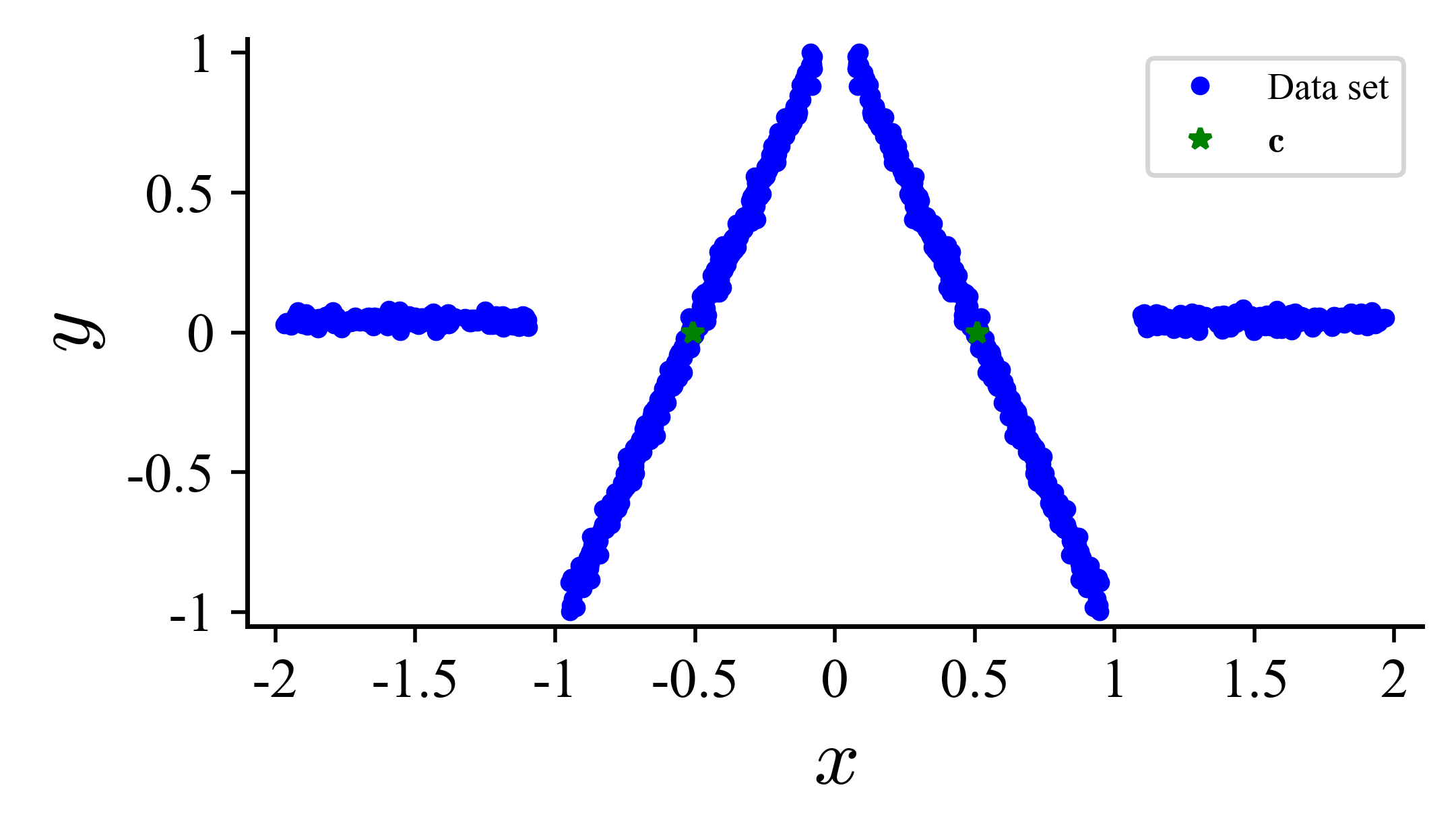}
        \caption{Synthetic data for Case $2$.}
        \label{fig:img_synthetic_data_Case 2_reg_gn_given}
    \end{subfigure}
    \caption{Synthetic data sets for linear regression with known points of interest.}
    \label{fig:Linear_Regression_reg_gn_given_synthetic_data}
\end{figure}

Here, our two synthetic data sets are composed of $1058$ and $1411$ labeled examples, respectively, and shown on Figure \ref{fig:Linear_Regression_reg_gn_given_synthetic_data}. The points of interest are still depicted by green stars.

\begin{figure}[H]
   \begin{subfigure}[b]{0.48\textwidth}
        \includegraphics[scale=0.62]{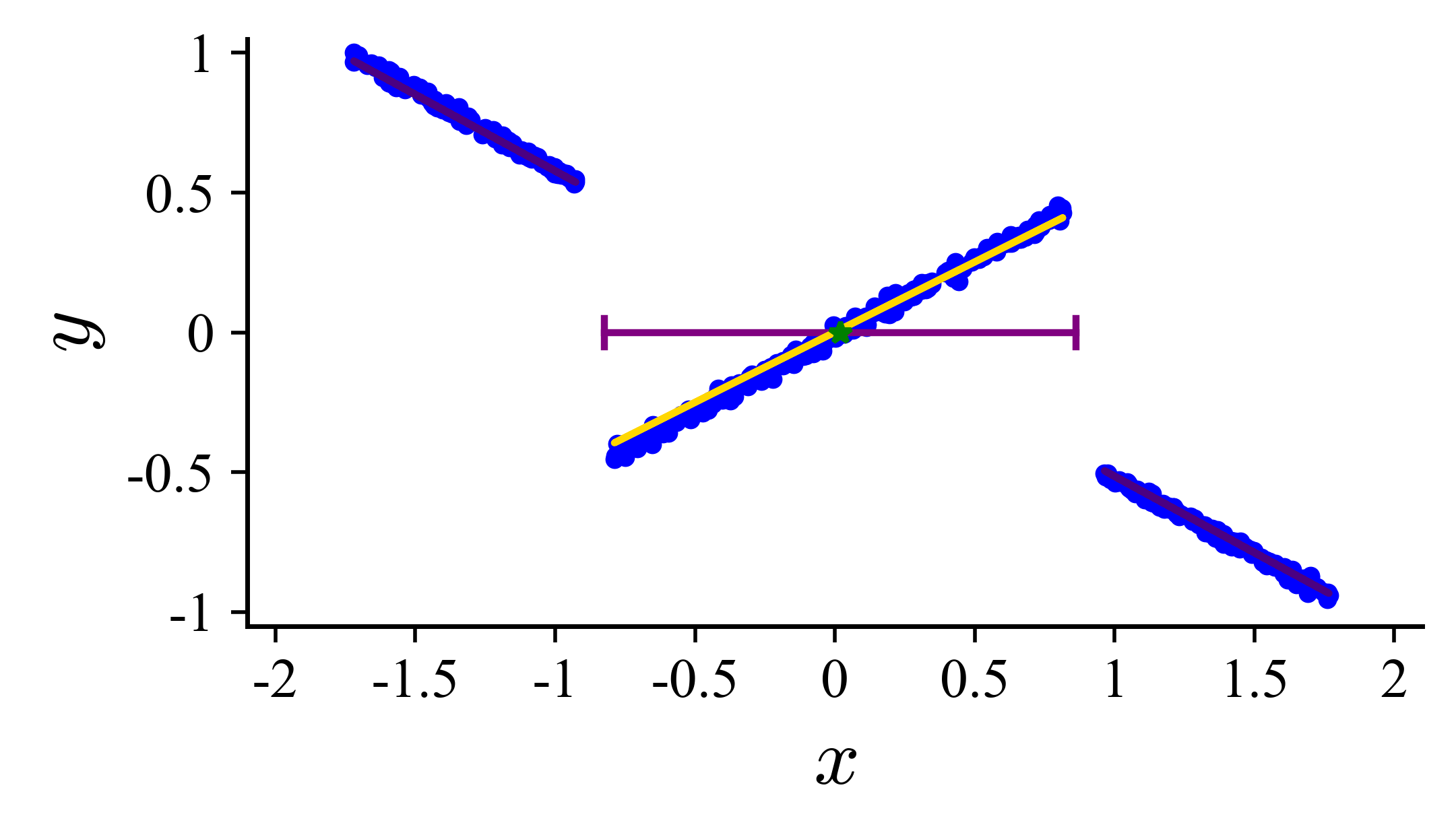}
        \caption{Results obtained via $\mathbf{PB^{\lambda}}(S)$ in Case $1$.}
        \label{fig:img_visualize_hyperplane_hybrid_model_Case 1_Apprentissage_reg_gn_given}
    \end{subfigure}
    \hfill
    \begin{subfigure}[b]{0.48\textwidth}
        \includegraphics[scale=0.62]{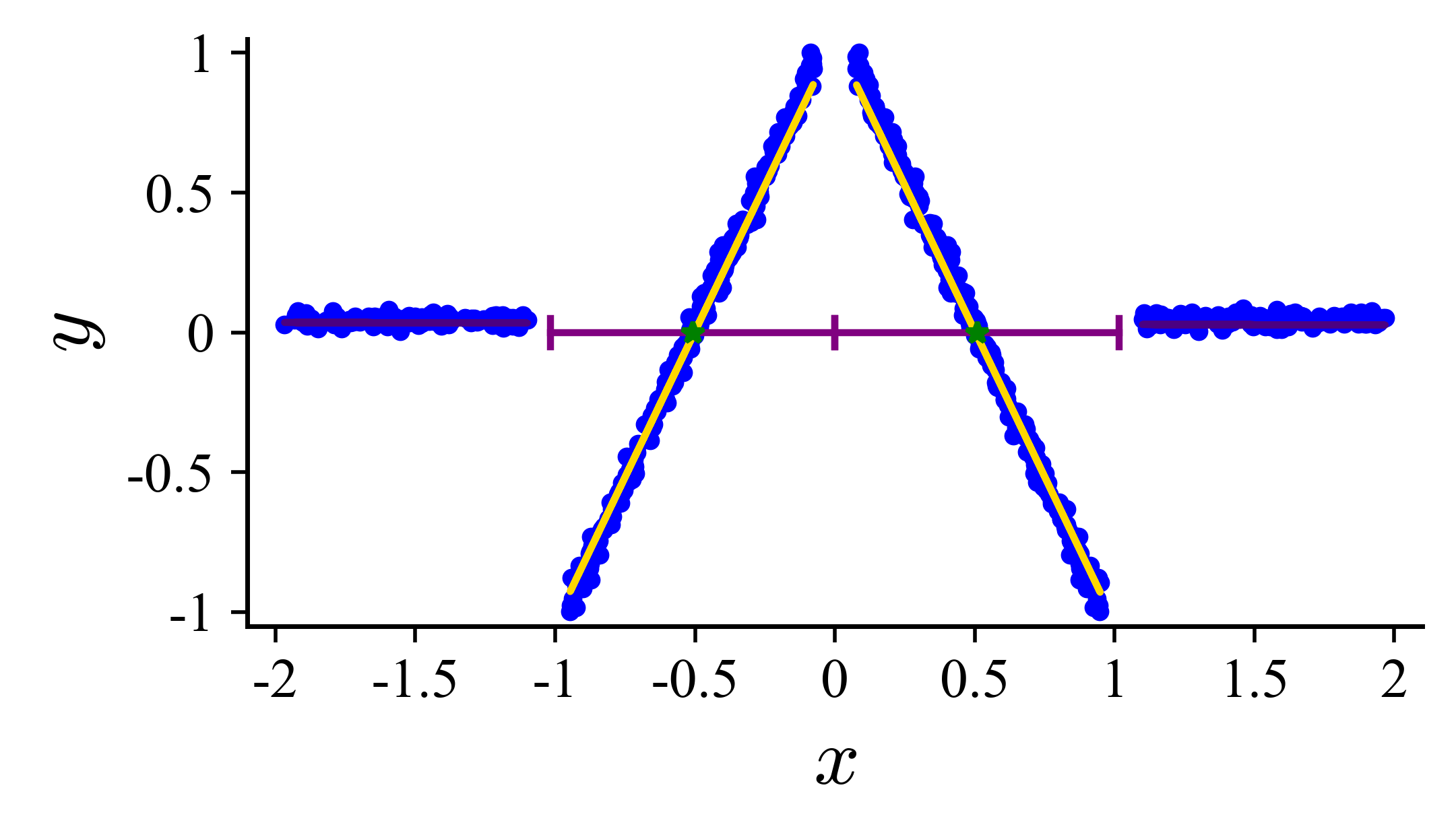}
        \caption{Results obtained via $\mathbf{PB^{\lambda}}(S)$ in Case $2$.}
        \label{fig:img_visualize_hyperplane_hybrid_model_Case 2_Apprentissage_reg_gn_given}
    \end{subfigure}
    \hfill
    \begin{subfigure}[b]{\textwidth}
    \begin{subfigure}[b]{\textwidth}
        \centering
        \vspace{-49mm}
        \includegraphics[scale=0.95]{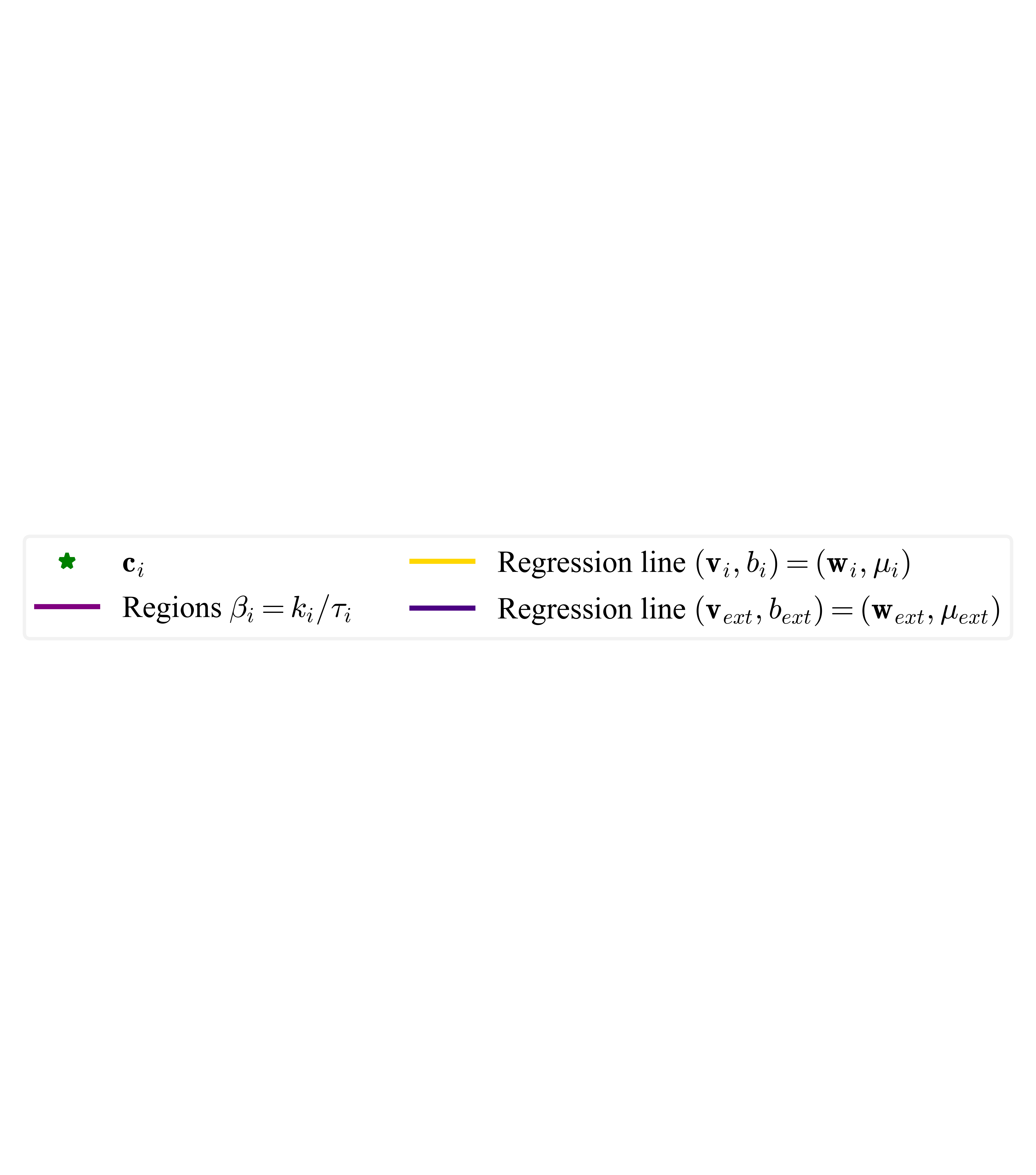}
    \end{subfigure}
    \end{subfigure}
    \vspace{-57mm}
    \caption{Results of mixtures of transparent local linear regressors on the training set, with known points of interest.}
    \label{fig:img_visualize_hyperplane_hybrid_model_Case 1_2_Apprentissage_reg_gn_given}
\end{figure}

It is evident that a classical linear regression model would not be appropriate for these tasks. However, after learning our \textit{mixtures of transparent local linear regressors}, the predictions made from the averaged deterministic predictors give us the Figure \ref{fig:img_visualize_hyperplane_hybrid_model_Case 1_2_Apprentissage_reg_gn_given}.

In this idealized situation with no noisy synthetic data sets, Figure \ref{fig:img_visualize_hyperplane_hybrid_model_Case 1_2_Apprentissage_reg_gn_given} shows that the \textit{posterior} distributions returned by the algorithm $\mathbf{PB^{\lambda}}(S)$ are quite accurate. So we can confirm that the algorithm $\mathbf{PB^{\lambda}}(S)$ from~\eqref{eq:reg_gn_bound_minimize_form} works as well as it should, when the points of interest are known and the localities are bounded intervals centered on these points. 

Table \ref{tab:R2_score_on_synth_data_reg_gn_given} below summarizes, for different regression models, the averaged coefficients of determination $R^{2}$ ($R2$ score).

\begin{table}[H]
\centering
\begin{tabular}{llccc}
\toprule
\multicolumn{2}{c}{Data set} & Linear SVR & Gaussian SVR & Our regression model \\
\midrule
\multirow{3}{*}{Case 1} & Training set   & 0.6514 & 0.9737 & \textbf{0.9984}  \\
                        & Validation set & 0.6460 & 0.9720 & \textbf{0.9986}  \\
                        & Test set       & 0.5958 & 0.9715 & \textbf{0.9984}  \\ 
\midrule
\multirow{3}{*}{Case 2} & Training set   & 0.0002  & 0.9433 & \textbf{0.9933}  \\
                        & Validation set & -0.0052 & 0.9396 & \textbf{0.9935}  \\
                        & Test set       & -0.0083 & 0.9385 & \textbf{0.9933}  \\
\bottomrule
\end{tabular}
\caption{Averaged R2 score of the regression models on synthetic data sets with known points of interest.}
\label{tab:R2_score_on_synth_data_reg_gn_given}
\end{table}
On these types of complex tasks, this experiment suggests that our \textit{mixture of transparent local linear regressors} performs better than its interpretable counterpart SVR (with linear kernel). It doesn't also lose too much performance to gain interpretability since it performs relatively well compared with Gaussian kernel SVR.

In Table \ref{tab:risk_bounds_on_synth_data_reg_gn_given}, we present the empirical risk $L_{S}(Q)$ obtained and the core component of the risk bound $L_{S}(Q) + \frac{1}{\lambda} KL_{reg}(Q||P)$ of our regressors given by~\eqref{eq:risque_généralisation_forme_empirique} and \eqref{eq:reg_gn_bound_minimize_form}. We observe that the bound core component is tight, since it is less than $4.5$ times the $L_{S}(Q)$ in both cases.

\begin{table}[H]
\centering
\begin{tabular}{lcc}
\toprule
Data set & $L_{S}(Q)$ & Bound core component \\
\midrule
\multirow{1}{*}{Case 1} & 0.0117 & 0.0265  \\
\midrule
\multirow{1}{*}{Case 2} & 0.0103 & 0.0463  \\
\bottomrule
\end{tabular}
\caption{$L_{S}(Q)$ and the bound core component from~\eqref{eq:risque_généralisation_forme_empirique} and \eqref{eq:reg_gn_bound_minimize_form}, of our regressors on the training set, with known points of interest.}
\label{tab:risk_bounds_on_synth_data_reg_gn_given}
\end{table}

\subsection{Experiments with Unknown Points of Interest}
As in the previous subsection, we still work with the same synthetic data sets. However, in this subsection, the points of interest are unknown; only their number $n$ is known.

\subsubsection{Binary Linear Classification}
\begin{figure}[H]
   \centering
   \begin{subfigure}[b]{0.48\textwidth}
        \includegraphics[scale=0.044]{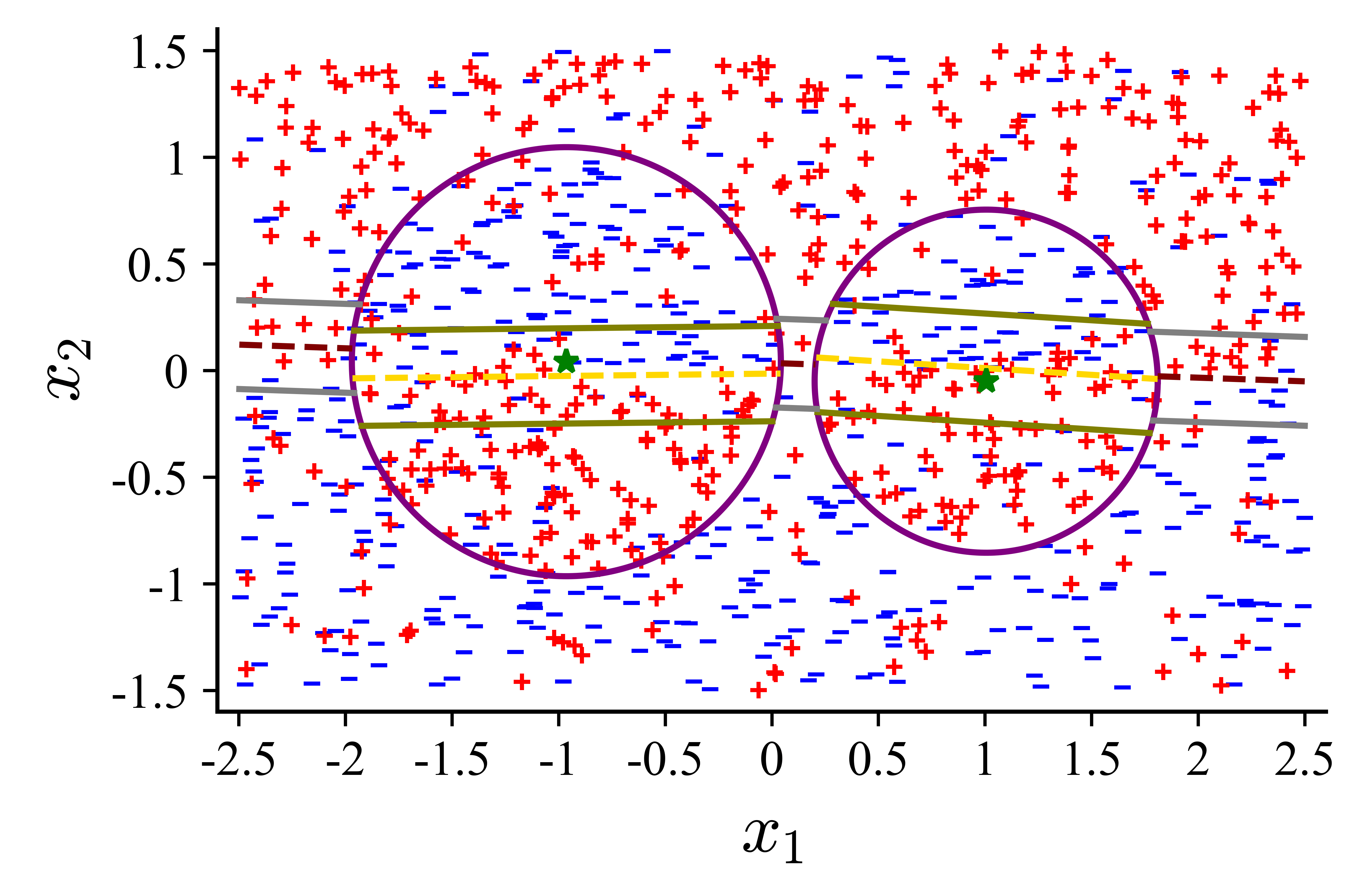}
        \caption{Results obtained via $\mathbf{PB^{\lambda}}(S)$ in Case $1$.}
        \label{fig:img_visualize_hyperplane_hybrid_model_Case 1_Jeu d'entraînement_clf_gn_search}
    \end{subfigure}
    \hfill
    \begin{subfigure}[b]{0.48\textwidth}
        \includegraphics[scale=0.044]{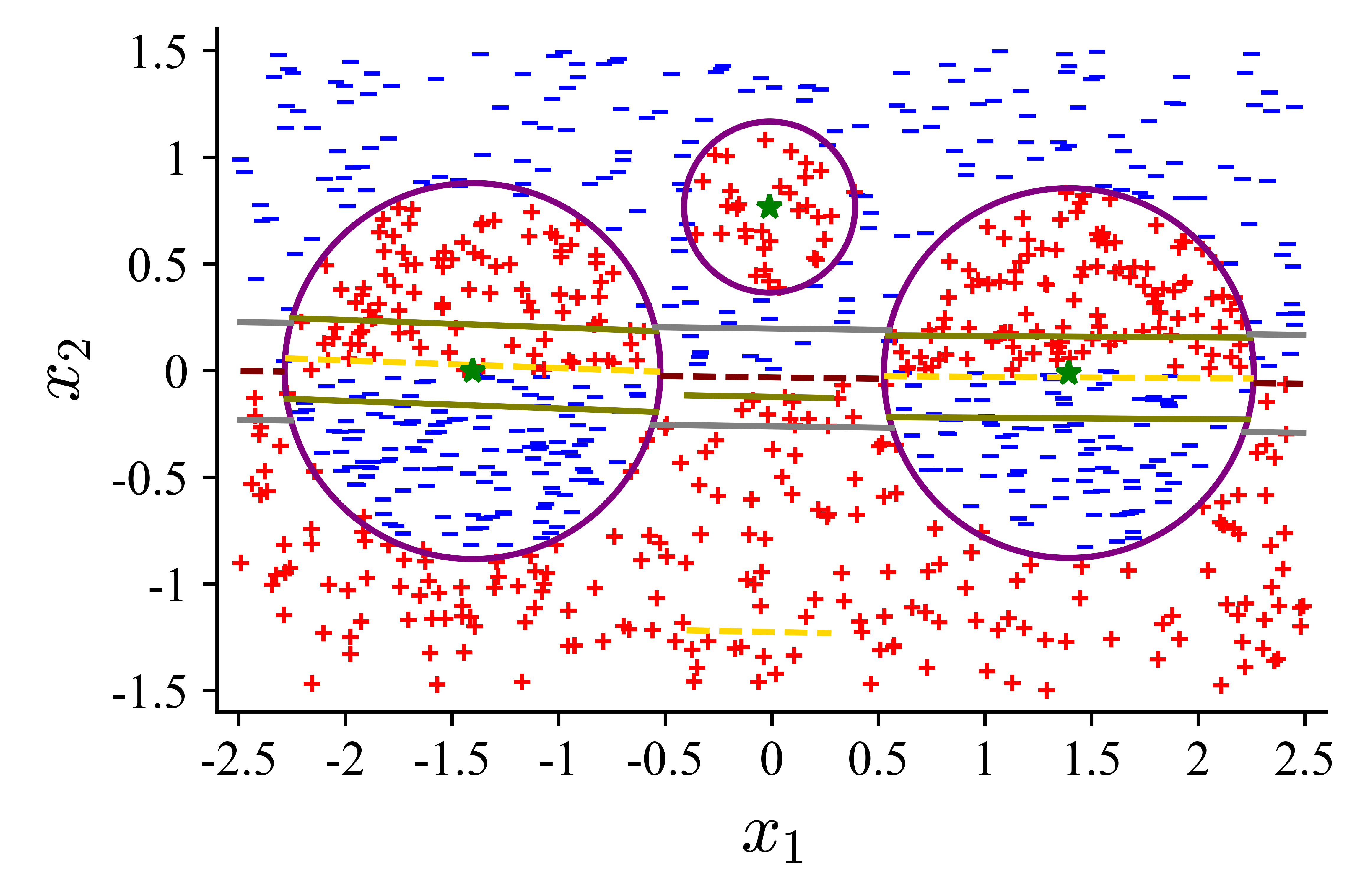}
        \caption{Results obtained via $\mathbf{PB^{\lambda}}(S)$ in Case $2$.}
        \label{fig:img_visualize_hyperplane_hybrid_model_Case 2_Jeu d'entraînement_clf_gn_search}
    \end{subfigure}
    \hfill
    \begin{subfigure}[b]{\textwidth}
    \begin{subfigure}[b]{\textwidth}
        \centering
        \vspace{-51mm}
        \includegraphics[scale=0.0705]{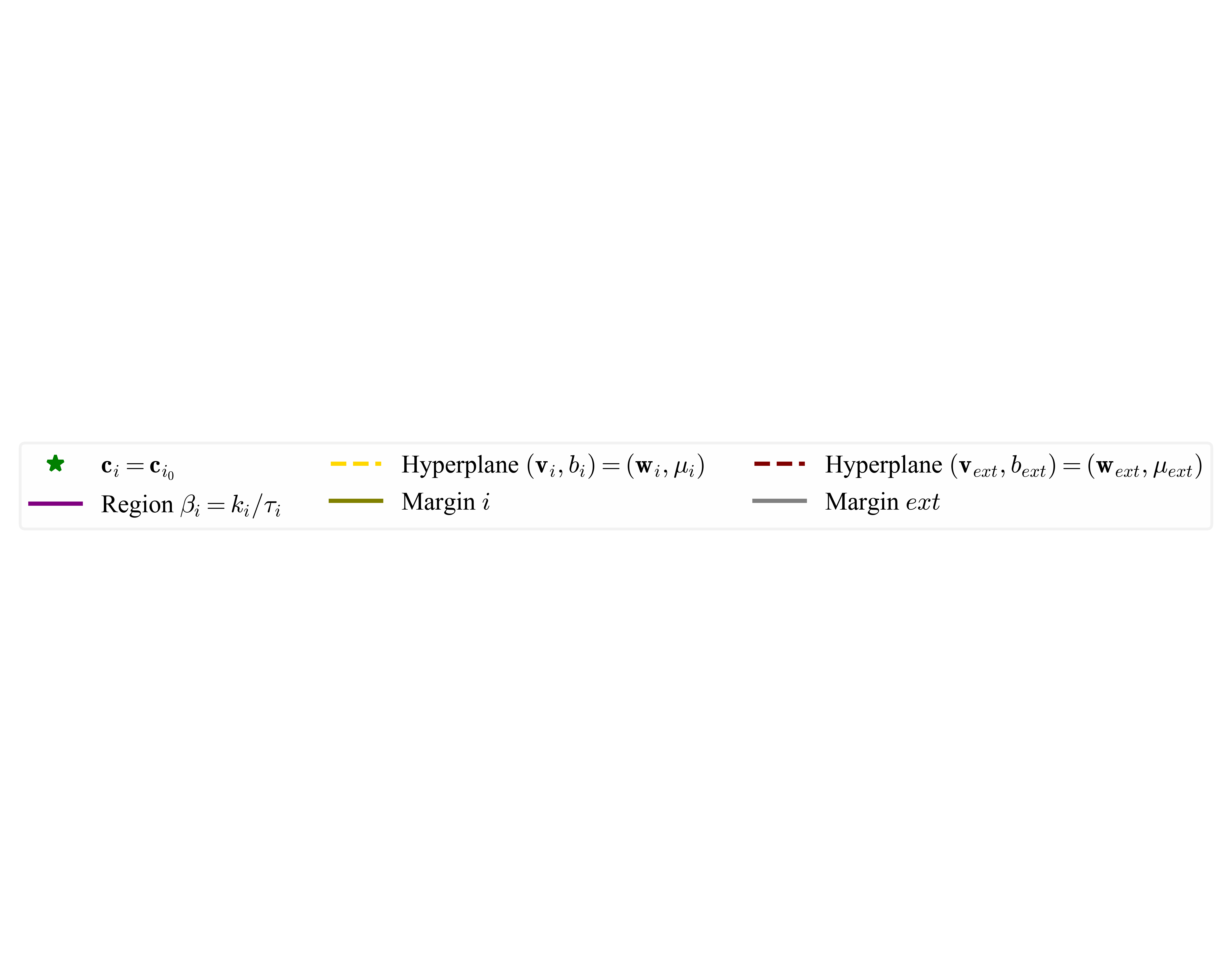}
    \end{subfigure}
    \end{subfigure}
    \vspace{-58mm}
    \caption{Results of mixtures of transparent local linear classifiers on the training set, with unknown points of interest.}
    \label{fig:img_visualize_hyperplane_hybrid_model_Case 1_2_Jeu d'entraînement_clf_gn_search}
\end{figure}

In this setup, we consider $n = 2$ for Figure \ref{fig:img_synthetic_data_Case 1_clf_gn_given} and $n = 3$ for Figure \ref{fig:img_synthetic_data_Case 2_clf_gn_given}. After learning the \textit{mixtures of transparent local linear classifiers}, the predictions made from the averaged deterministic predictors, are illustrated in Figure \ref{fig:img_visualize_hyperplane_hybrid_model_Case 1_2_Jeu d'entraînement_clf_gn_search}. 

Here, one can see that the proposed points of interest along with the locality parameters and their corresponding linear separating hyperplanes returned by the algorithm $\mathbf{PB^{\lambda}}(S)$ are pretty good when the number of points of interest $n$ is correctly specified. So we can confirm that the algorithm $\mathbf{PB^{\lambda}}(S)$ from~\eqref{eq:clf_gn_bound_minimize_form_search} works as well as it should in this setup.

Table \ref{tab:risk_bounds_on_synth_data_clf_gn_search} presents the averaged accuracy, the empirical risk $L_{S}(Q)$ of~\eqref{eq:nouveau_risque_généralisation_forme_empirique} and the bound core component of our classifier from~\eqref{eq:clf_gn_bound_minimize_form_search}. In both synthetic data sets, we got a tight bound core component as it is less than $1.65$ times the $L_{S} (Q)$, while the accuracy is still similar to that of the case where the points of interest were known (see Table \ref{tab:accuracy_on_synth_data_clf_gn_given}).

\begin{table}[H]
\centering
\begin{tabular}{llccc}
\toprule
\multicolumn{2}{c}{Data set} & Accuracy (\%) & $L_{S}(Q)$ & Bound core component \\
\midrule
\multirow{3}{*}{Case 1} & Training set   & 77.9570 & 0.2948 & 0.3421  \\
                        & Validation set & 79.4730 & - & -  \\
                        & Test set       & 77.0820 & - & -  \\ 
\midrule
\multirow{3}{*}{Case 2} & Training set   & 96.7860 & 0.1588 & 0.2563  \\
                        & Validation set & 96.8460 & - & -  \\
                        & Test set       & 95.2410 & - & -  \\
\bottomrule
\end{tabular}
\caption{Averaged accuracy, $L_{S}(Q)$ and bound core component of our classifier on synthetic data sets with unknown points of interest.}
\label{tab:risk_bounds_on_synth_data_clf_gn_search}
\end{table}

\subsubsection{Linear Regression}
\begin{figure}[H]
   \begin{subfigure}[b]{0.48\textwidth}
        \includegraphics[scale=0.62]{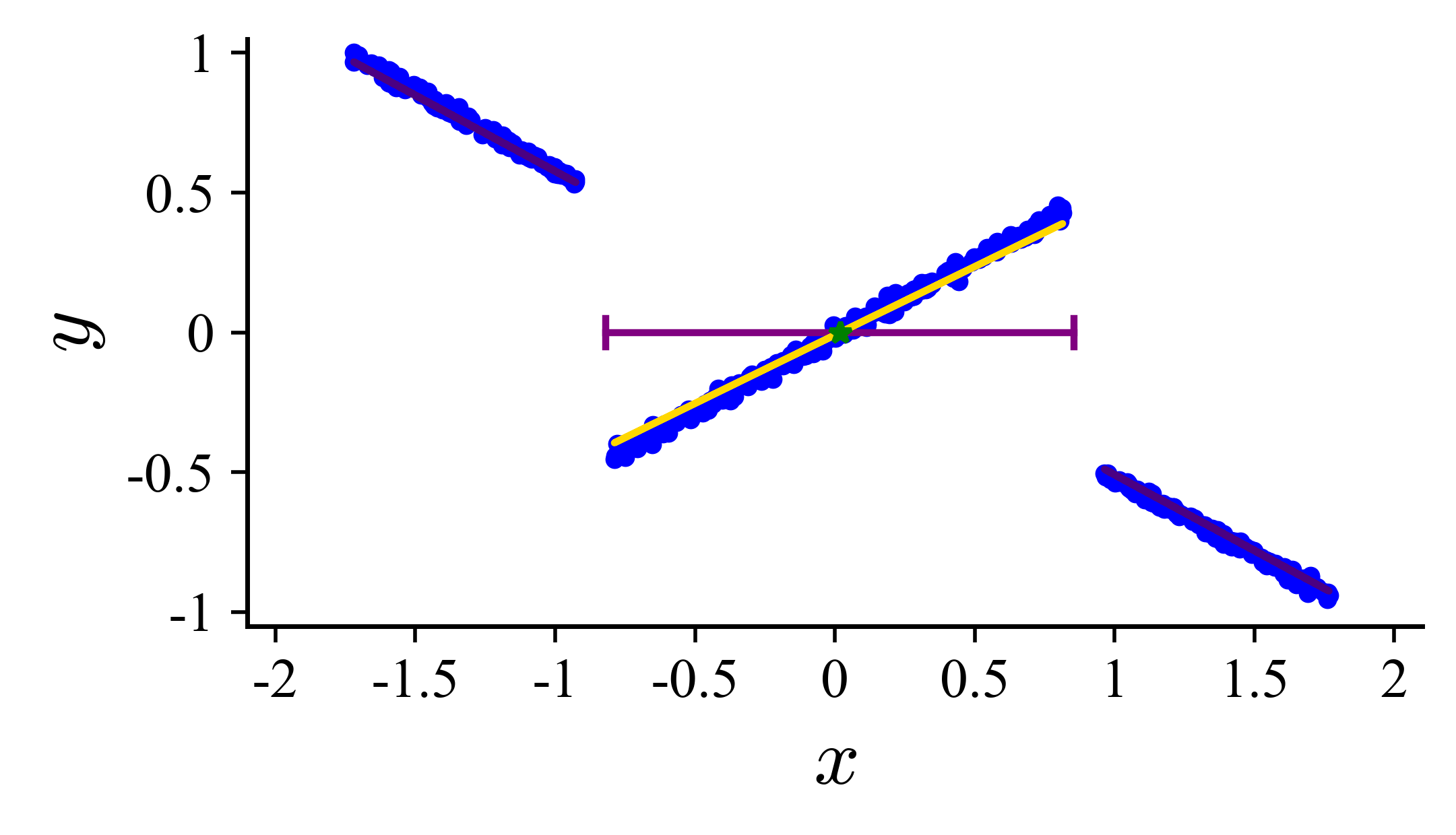}
        \caption{Results obtained via $\mathbf{PB^{\lambda}}(S)$ in Case $1$.}
        \label{fig:img_visualize_hyperplane_hybrid_model_Case 1_Apprentissage_reg_gn_search}
    \end{subfigure}
    \hfill
    \begin{subfigure}[b]{0.48\textwidth}
        \includegraphics[scale=0.62]{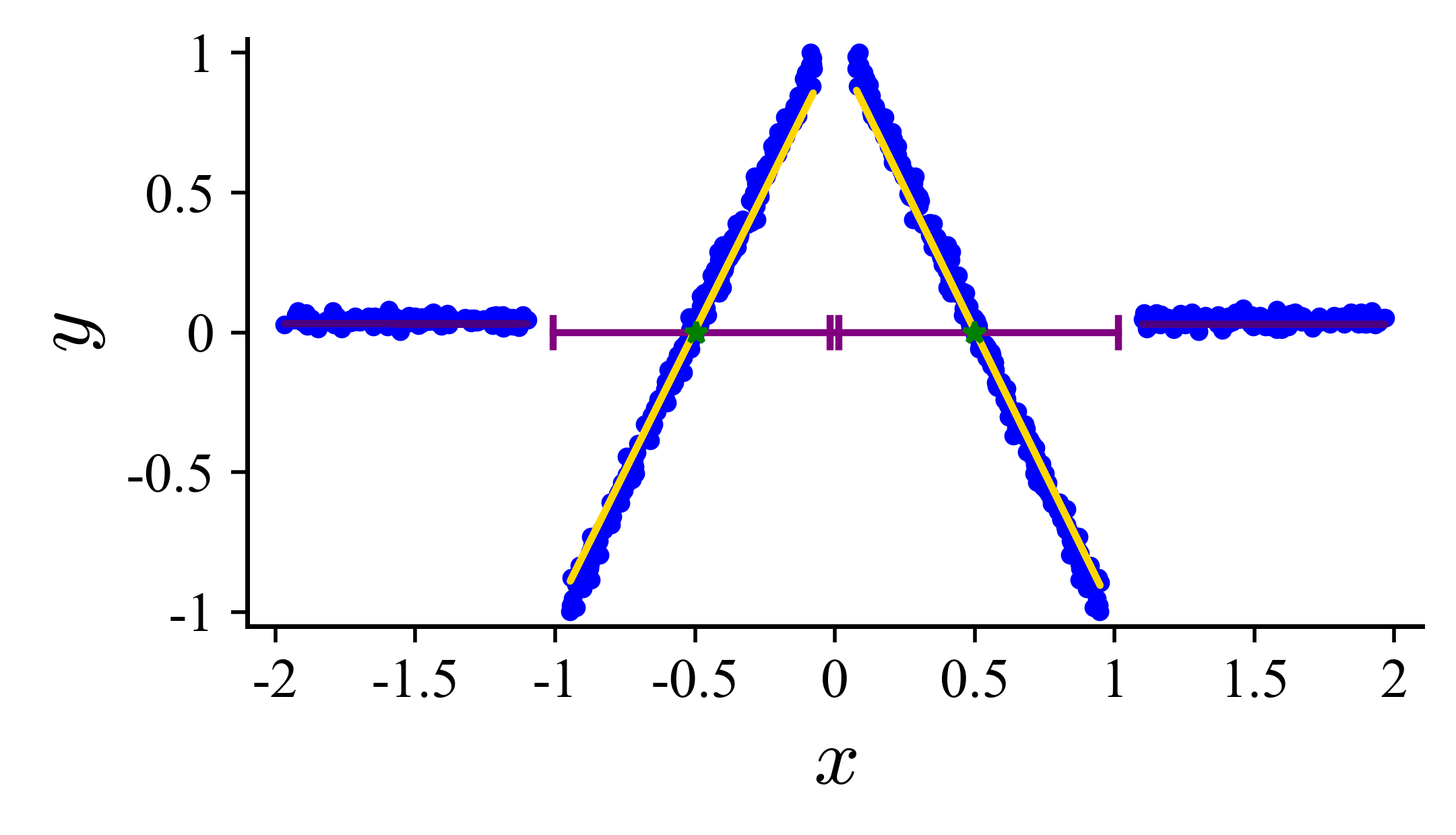}
        \caption{Results obtained via $\mathbf{PB^{\lambda}}(S)$ in Case $2$.}
        \label{fig:img_visualize_hyperplane_hybrid_model_Case 2_Apprentissage_reg_gn_search}
    \end{subfigure}
    \hfill
    \begin{subfigure}[b]{\textwidth}
    \begin{subfigure}[b]{\textwidth}
        \centering
        \vspace{-49mm}
        \includegraphics[scale=0.95]{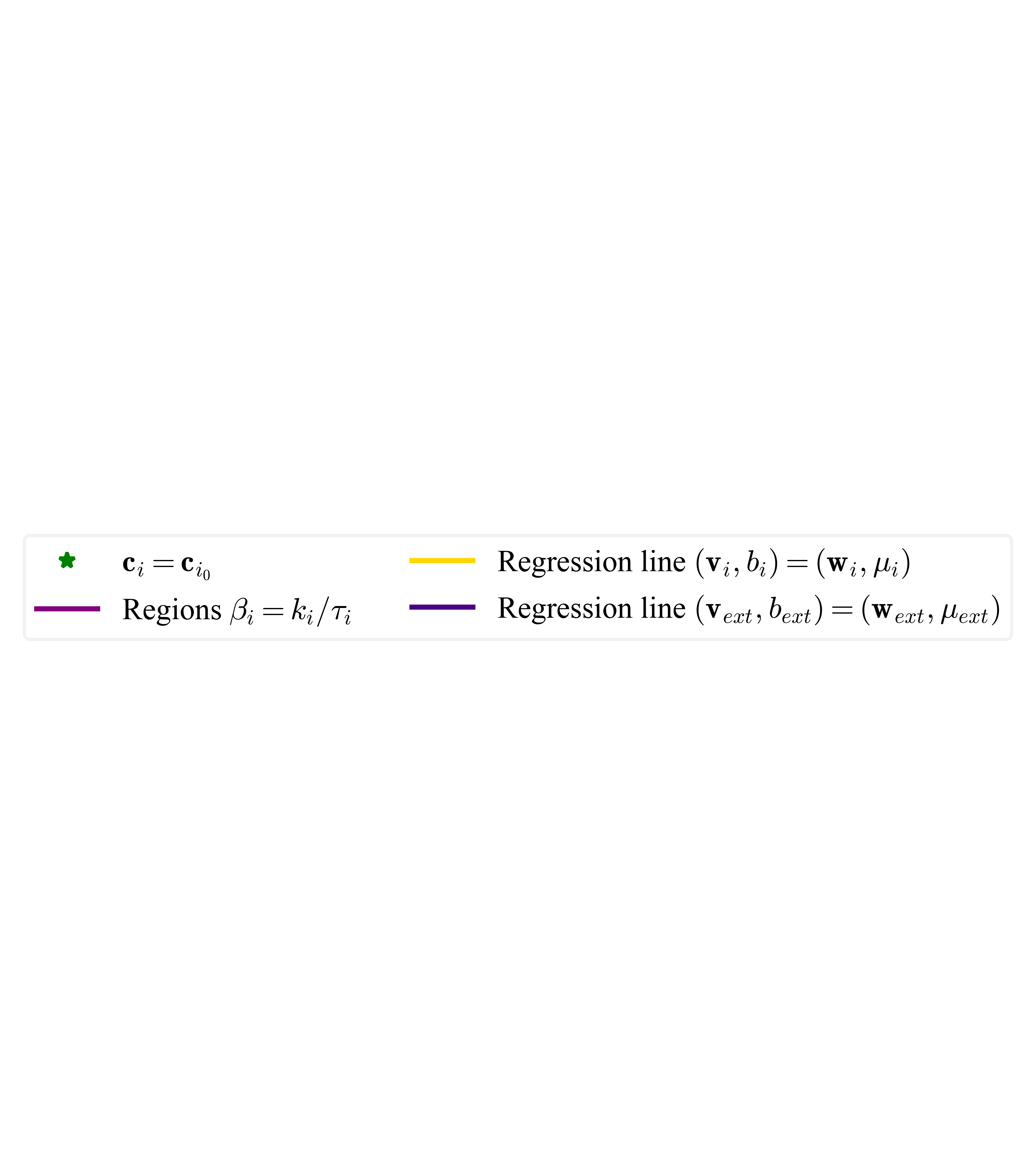}
    \end{subfigure}
    \end{subfigure}
    \vspace{-57mm}
    \caption{Results of mixtures of transparent local linear regressors on the training set, with unknown points of interest.}
    \label{fig:img_visualize_hyperplane_hybrid_model_Case 1_2_Apprentissage_reg_gn_search}
\end{figure}

In linear regression problem, let us choose $n = 1$ for Figure \ref{fig:img_synthetic_data_Case 1_reg_gn_given} and $n = 2$ for Figure \ref{fig:img_synthetic_data_Case 2_reg_gn_given}. After learning the \textit{mixtures of transparent local linear regressors}, the predictions made from the averaged deterministic predictors, are shown in Figure \ref{fig:img_visualize_hyperplane_hybrid_model_Case 1_2_Apprentissage_reg_gn_search}.

Figure \ref{fig:img_visualize_hyperplane_hybrid_model_Case 1_2_Apprentissage_reg_gn_search} reveals that our last algorithm $\mathbf{PB^{\lambda}}(S)$ from~\eqref{eq:reg_gn_bound_minimize_form_search} works as expected in this idealized situation. As shown in Table \ref{tab:risk_bounds_on_synth_data_reg_gn_search}, we also have a tight bound core component for both synthetic data sets.

\begin{table}[H]
\centering
\begin{tabular}{llccc}
\toprule
\multicolumn{2}{c}{Data set} & R2 score & $L_{S}(Q)$ & Bound core component \\
\midrule
\multirow{3}{*}{Case 1} & Training set   & 0.9955 & 0.0160 & 0.0370  \\
                        & Validation set & 0.9983 & - & -  \\
                        & Test set       & 0.9956 & - & -  \\ 
\midrule
\multirow{3}{*}{Case 2} & Training set   & 0.9884 & 0.0264 & 0.0865  \\
                        & Validation set & 0.9883 & - & -  \\
                        & Test set       & 0.9884 & - & -  \\
\bottomrule
\end{tabular}
\caption{Averaged R2 score, $L_{S}(Q)$ and bound core component of our regressor on synthetic data sets with unknown points of interest.}
\label{tab:risk_bounds_on_synth_data_reg_gn_search}
\end{table}

\subsection{Real Data Experiments with Unknown Points of Interest}
Now, let's exemplify our works on some real data sets. We have downloaded $16$ data sets from the KEEL dataset repository (Knowledge Extraction based on Evolutionary Learning)~\citep{alcala2011keel} . There are $8$ data sets for binary linear classification task and the $8$ other for linear regression. Our objective in this subsection is twofold. As in the previous subsection, we want to show that our approch doesn't sacrifice too much performance to bring interpretability. The second objective is, in the binary linear classification problem, to see how much our classifier can be a good candidate (in terms of performance) when compared with \citet{vogel2024meta}'s League of Experts (with linear experts).

In order to simplify our analysis, in the binary linear classification problem we will apply our method with the number of unknown points of interest $n \in \{1, 2\}$. As for \citet{vogel2024meta}'s method, we take the number of linear experts to be $n\_experts \in \{2, 3\}$. 

Note that for each of the $16$ data sets, the rows containing missing values were dropped, as well as duplicated rows. Then, the verification of multicollinearity is done. Multicollinearity occurs when a variable is highly correlated with one or more other variables. This aspect generally makes the prediction much less reliable in the multiple linear regression and binary linear classification cases. To solve this problem, we calculate the Variance Inflation Factor (VIF) and remove any variable with a VIF greater than $10$ ~\citep{alauddin2010instructional, salmeron2018variance}.

\subsubsection{Binary Linear Classification}
The performance comparison between our \textit{mixtures of transparent local classifiers} and the LoE (with linear experts) is given in Table \ref{tab:accuracy_on_real_data_clf_gn_search}. More specifically, we present the averaged accuracy $(\%)$ over $10$ reproductions of Section \ref{Total_Number_of_Executions}. The data sets with a star symbol are those containing both numerical and categorical features, while the remaining contain only numerical features. The values in bracket indicate the data points and the number of features, respectively. The one-hot encoding method was applied to convert categorical features into numeric ones, before the standardization of all the features.
\begin{table}[H]
\centering
\begin{tabular}{l|cc|cc}
\toprule
\multirow{2}{*}{Data set} & \multicolumn{2}{c}{Our classifier (\%)} & \multicolumn{2}{c}{LoE (\%)}  \\
\cline{2-5}
& {$n = 1$} & {$n = 2$} & {$n\_experts = 2$} & {$n\_experts = 3$} \\
\midrule
Banana (5292, 3) & 83.48 & \textbf{87.33} & 72.63 & 77.46  \\

Breast Cancer Wisconsin (569, 31) & \textbf{93.95} & 93.37 & 93.72 & 92.56  \\

German* (1000, 21) & \textbf{73.67} & 73.67 & 72.00 & 70.80  \\

Phoneme (5349, 6) & 79.50 & \textbf{80.18} & 77.05 & 79.00  \\

Ringnorm (7400, 21) & \textbf{94.60} & 88.52 & 77.52 & 78.84  \\
                        
Saheart* (462, 10) & 68.43 & 68.00 & 68.43 & \textbf{69.86}  \\

Spectfheart (267, 45) & \textbf{78.29} & 77.56 & 77.07 & 77.81  \\

Twonorm (7400, 21) & \textbf{97.98} & 97.94 & 97.78 & 97.50  \\
\bottomrule
\end{tabular}
\caption{Averaged accuracy of our classifier and that of the LoE (with linear experts) on the testing set.}
\label{tab:accuracy_on_real_data_clf_gn_search}
\end{table}
We distinguish mainly three cases here. First, on the Banana and Ringnorm data sets, we notice that the performance of our classifier is very high compared to that obtained by the LoE, with a similar number of linear experts. Second, our classifier brings a slight improvement in performance on the German and Phoneme data sets. Finally, for the remaining data sets the two methods appear to have similar performance.

According to these results, our method provides a slight advantage compared with the LoE (with linear experts). This advantage potentially comes from the non-partitioning of the instance space, thus allowing overlap between regions, unlike the disjoint partition method performed in the LoE.

\subsubsection{Linear Regression}
In addition to the verification of multicollinearity, we also checked the presence of outliers by using the well-known Box method~\citep{walfish2006review, dawson2011significant}. After pre-processing the data, we move on to the design of our \textit{mixtures of transparent local linear regressors} with the number of unknown points of interest $n \in \{0, 1, 2, 3, 4\}$. Notice that the case $n = 0$ represents a global model without any notion of points of interest and localities. We also learn the SVR with linear and Gaussian kernels. Table \ref{tab:r2_score_on_real_data_reg_gn_search} displays the averaged R2 score over $10$ reproductions of Section \ref{Total_Number_of_Executions}. The bold italic characters show the second highest scores.

\begin{table}[H]
\centering
\begin{tabular}{l|ccccc|cc}
\toprule
\multirow{2}{*}{Data set} & \multicolumn{5}{c}{Our regressor} & \multicolumn{1}{c}{L SVR} & \multicolumn{1}{c}{G SVR} \\
\cline{2-8}
& {$n = 0$} & {$n = 1$} & {$n = 2$} & {$n = 3$} & {$n = 4$} & {-} & {-} \\
\midrule
Auto MPG8 (372, 8) & 0.64 & 0.65 & \textbf{0.66} & 0.65 & 0.65 & 0.64 & \textbf{\textit{0.66}} \\

Baseball (190, 14) & \textbf{\textit{0.63}} & 0.63 & 0.58 & 0.61 & 0.62 & \textbf{0.65} & 0.61 \\

Concrete (911, 9) & \textbf{\textit{0.74}} & 0.74 & 0.70 & 0.73 & 0.73 & 0.74 & \textbf{0.81} \\

Daily Electricity (339, 7) & \textbf{\textit{0.75}} & 0.75 & 0.75 & 0.74 & 0.72 & 0.75 & \textbf{0.77} \\

Laser Generated (881, 5) & 0.72 & 0.73 & 0.74 & 0.75 & \textbf{\textit{0.77}} & 0.73 & \textbf{0.91} \\

Stock Prices (926, 10) & 0.62 & 0.68 & 0.75 & 0.52 & \textbf{\textit{0.75}} & 0.63 & \textbf{0.85} \\

Treasury (933, 16) & \textbf{\textit{0.99}} & 0.99 & 0.99 & 0.98 & 0.99 & 0.99 & \textbf{0.99} \\

Weather Izmir (911, 8) & 0.88 & \textbf{0.90} & 0.88 & 0.88 & 0.88 & \textbf{\textit{0.89}} & 0.88 \\

\bottomrule
\end{tabular}
\caption{Mean R2 score of the regression models on the testing set.}
\label{tab:r2_score_on_real_data_reg_gn_search}
\end{table}
Here, we can see that the Gaussian SVR (G SVR) generally dominates the other methods, but sacrifices interpretability for the benefit of performance gain. As for our method, its performance is between that of linear SVR (L SVR) and Gaussian SVR. For example, on Auto, Laser, Stock, and Weather data, our method's performances are not far from those of Gaussian SVR, and strictly higher than those obtained by linear SVR. The results of these four data sets suggest that there is an interest in using our approach and also that the Euclidean norm used as the default metric of similarity distance $d(\cb, \xb)$ is suitable. In contrast, the performance observed on the remaining data sets reveals instances where our regressor's performance remains minimal but usually improves on linear SVR (like in Concrete, Daily Electricity and Treasury). The choice of the metric $d(\cb, \xb)$ could be a factor here.

Overall, the experiments of Section \ref{Experiments} show that, depending on the data configurations, our approach can provide a significant performance gain over the classic linear model, sometimes approaching non-interpretable models such as the Gaussian SVM or Gaussian SVR.

\section{Conclusion}
We presented an alternative interpretable model design that contributes to the advancement of research focused on interpretability. Our proposal of mixtures of transparent local models with the linear model as the transparent model seems to be very relevant. We derived PAC-Bayesian guarantees in the scenario where the points of interest are given and also when they aren't but only their number is given. In the first scenario, the provided guarantee is valid for any metric $d(\cb, \xb)$. However, this is not the case in the second scenario, where the provided guarantee is valid for the Euclidean metric only.
Therefore, it would be interesting to examine in future work the possibility of introducing flexibility in the choice of the metric $d(\cb, \xb)$ in the case of unknown points of interest. Another extension of our work would be to use other types of transparent models instead of linear models.


\acks{This work is supported by the Collaborative Research and Development Grant from Beneva and NSERC, CRDPJ 529584.}



\newpage

\appendix








\section{Proof of the Expression of Loss Function in Section~\ref{Binary_Linear_clf_gn}}
\label{app:Binary_Linear_clf_gn}

This appendix aims to prove the expression of the loss $\ell\Bigl[ Q, (\bpsi, \xb, y) \Bigr]$ given by~\eqref{eq:perte_Q_exp_generale_clf_gn} in Section~\ref{Binary_Linear_clf_gn}. Starting from the definition of the loss $\ell\Bigl[ Q, (\bpsi, \xb, y) \Bigr]$ given by~\eqref{eq:perte_Q_exp_generale_clf_gn_def}, we have 
\begin{equation}\label{eq:perte_Q_exp_generale_clf_gn_proof_1}
    \begin{aligned}
        \ell\Bigl[ Q, (\bpsi, \xb, y) \Bigr] &\eqdef \underset{Q}{\EE} \LC \sum_{i=1}^{n} \Biggl[ \mathds{1}\LB \Bigl( \langle \vb_{i}, \bpsi \rangle + y b_{i} \Bigr) \leq 0 \RB K(\cb_{i}, \xb, \be_{i}) \Biggr] \RC \\
        & \quad \quad \quad \quad \quad + \underset{Q}{\EE} \LC \mathds{1}\LB \Bigl( \langle \vb_{ext}, \bpsi \rangle + y b_{ext} \Bigr) \leq 0 \RB \prod_{i=1}^{n} \bar{K}(\cb_{i}, \xb, \be_{i}) \RC \textrm{,}
    \end{aligned}
\end{equation}
where $\bpsi \eqdef y \xb$.

Knowing that the predictors's parameters $\LC \vb_{i}, b_{i} \RC_{i=1}^{n}$ and their corresponding locality parameters $\LC \be_{i} \RC_{i=1}^{n}$ are all independent, Equation~\eqref{eq:perte_Q_exp_generale_clf_gn_proof_1} becomes
\begin{equation}\label{eq:perte_Q_exp_generale_clf_gn_proof_2}
    \begin{aligned}
        \ell\Bigl[ Q, (\bpsi, \xb, y) \Bigr] &= \sum_{i=1}^{n} \Biggl[ \underset{Q}{\EE} \LC \mathds{1}\LB \Bigl( \langle \vb_{i}, \bpsi \rangle + y b_{i} \Bigr) \leq 0 \RB K(\cb_{i}, \xb, \be_{i}) \RC \Biggr] \\
        & \quad \quad \quad \quad \quad + \underset{Q}{\EE} \LC \mathds{1}\LB \Bigl( \langle \vb_{ext}, \bpsi \rangle + y b_{ext} \Bigr) \leq 0 \RB \prod_{i=1}^{n} \bar{K}(\cb_{i}, \xb, \be_{i}) \RC \\ 
        &= \sum_{i=1}^{n} \LB \underset{(\vb_{i}, b_{i}) \sim \LC Q_{\wb_{i}, \Ib_{i}} \times Q_{\mu_{i}, \sg_{i}^{2}} \RC}{\EE} \mathds{1}\LB \Bigl( \langle \vb_{i}, \bpsi \rangle + y b_{i} \Bigr) \leq 0 \RB \times \underset{ \be_{i} \sim Q_{k_{i}, \tau_{i}} }{\EE} K(\cb_{i}, \xb, \be_{i}) \RB \\
        & \quad + \underset{(\vb_{ext}, b_{ext}) \sim \LC Q_{\wb_{ext}, \Ib_{ext}} \times Q_{\mu_{ext}, \sg_{ext}^{2}} \RC}{\EE} \mathds{1}\LB \Bigl( \langle \vb_{ext}, \bpsi \rangle + y b_{ext} \Bigr) \leq 0 \RB \\ 
        & \quad \quad \quad \quad \quad \quad \quad \quad \quad \quad \quad \quad \quad \quad \quad \quad \quad \quad \quad \quad \times \prod_{i=1}^{n} \LB \underset{ \be_{i} \sim Q_{k_{i}, \tau_{i}} }{\EE} \bar{K}(\cb_{i}, \xb, \be_{i}) \RB \textrm{.} 
    \end{aligned}
\end{equation}

As defined in Subsection~\ref{Binary_Linear_clf_gn}, Equation~\eqref{eq:theta}, we know that
\[
    \underset{ \be_{i} \sim Q_{k_{i}, \tau_{i}} }{\EE} K(\cb_{i}, \xb, \be_{i}) = \underset{ \be_{i} \sim \Gm\LP k_{i}, \tau_{i} \RP }{\EE} \mathds{1}\Bigl[ d(\cb_{i}, \xb) \leq \be_{i} \Bigr] \eqdef \Theta_{k_{i}, \tau_{i}, d(\cb_{i}, \xb)} \textrm{.}
\]
Consequently, Equation~\eqref{eq:perte_Q_exp_generale_clf_gn_proof_2} can be rewritten as
\[
    \begin{aligned}
        \ell\Bigl[ Q, (\bpsi, \xb, y) \Bigr] &= \sum_{i=1}^{n} \LB \Theta_{k_{i}, \tau_{i}, d(\cb_{i}, \xb)} \times \underset{(\vb_{i}, b_{i}) \sim \LC Q_{\wb_{i}, \Ib_{i}} \times Q_{\mu_{i}, \sg_{i}^{2}} \RC}{\EE} \mathds{1}\LB \Bigl( \langle \vb_{i}, \bpsi \rangle + y b_{i} \Bigr) \leq 0 \RB \RB \\
        & \quad + \prod_{i=1}^{n} \LB 1 - \Theta_{k_{i}, \tau_{i}, d(\cb_{i}, \xb)} \RB \\ 
        & \quad \quad \quad \times \underset{(\vb_{ext}, b_{ext}) \sim \LC Q_{\wb_{ext}, \Ib_{ext}} \times Q_{\mu_{ext}, \sg_{ext}^{2}} \RC}{\EE} \mathds{1}\LB \Bigl( \langle \vb_{ext}, \bpsi \rangle + y b_{ext} \Bigr) \leq 0 \RB \textrm{.}
    \end{aligned}
\]

Let's now consider two quantities $\Acal \LP \wb_{i}, \mu_{i}, \sg_{i} \RP$ and $\Acal\LP \wb_{ext}, \mu_{ext}, \sg_{ext} \RP$ such that
\begin{align*}
    & \Acal\LP \wb_{i}, \mu_{i}, \sg_{i} \RP \eqdef \underset{(\vb_{i}, b_{i}) \sim \LC Q_{\wb_{i}, \Ib_{i}} \times Q_{\mu_{i}, \sg_{i}^{2}} \RC}{\EE} \mathds{1}\LB \Bigl( \langle \vb_{i}, \bpsi \rangle + y b_{i} \Bigr) \leq 0 \RB \quad \textrm{and} \\ 
    & \Acal\LP \wb_{ext}, \mu_{ext}, \sg_{ext} \RP \eqdef \underset{(\vb_{ext}, b_{ext}) \sim \LC Q_{\wb_{ext}, \Ib_{ext}} \times Q_{\mu_{ext}, \sg_{ext}^{2}} \RC}{\EE} \mathds{1}\LB \Bigl( \langle \vb_{ext}, \bpsi \rangle + y b_{ext} \Bigr) \leq 0 \RB \textrm{.} 
\end{align*}
Hence, we obtain
\begin{equation}\label{eq:perte_Q_exp_generale_clf_gn_proof_3}
        \begin{aligned}
        \ell\Bigl[ Q, (\bpsi, \xb, y) \Bigr] &= \sum_{i=1}^{n} \LB \Theta_{k_{i}, \tau_{i}, d(\cb_{i}, \xb)} \times \Acal\LP \wb_{i}, \mu_{i}, \sg_{i} \RP \RB \\
        & \quad \quad \quad \quad \quad + \prod_{i=1}^{n} \LB 1 - \Theta_{k_{i}, \tau_{i}, d(\cb_{i}, \xb)} \RB \times \Acal\LP \wb_{ext}, \mu_{ext}, \sg_{ext} \RP \textrm{.}
    \end{aligned}
\end{equation}

Note that the expressions of $\Acal\LP \wb_{i}, \mu_{i}, \sg_{i} \RP$ and $\Acal\LP \wb_{ext}, \mu_{ext}, \sg_{ext} \RP$ differ only by the subscript notations $i$ and $ext$. Thus, it is sufficient to obtain the analytical expression of $\Acal\LP \wb_{i}, \mu_{i}, \sg_{i} \RP$ to similarly deduce that of $\Acal\LP \wb_{ext}, \mu_{ext}, \sg_{ext} \RP$. From definition, we have
\[
    \begin{aligned}
        \Acal\LP \wb_{i}, \mu_{i}, \sg_{i} \RP &\eqdef \underset{(\vb_{i}, b_{i}) \sim \LC Q_{\wb_{i}, \Ib_{i}} \times Q_{\mu_{i}, \sg_{i}^{2}} \RC}{\EE} \mathds{1}\LB \Bigl( \langle \vb_{i}, \bpsi \rangle + y b_{i} \Bigr) \leq 0 \RB \\ 
        &= \underset{b_{i} \sim Q_{\mu_{i}, \sg_{i}^{2}}}{\EE} \LB \underset{\vb_{i} \sim Q_{\wb_{i}, \Ib_{i}}}{\EE} \mathds{1}\LB \Bigl( \langle \vb_{i}, \bpsi \rangle + y b_{i} \Bigr) \leq 0 \RB \RB \textrm{.} 
    \end{aligned}  
\]
Then, as $Q_{\wb_{i}, \Ib_{i}} \eqdef \Ncal \LP \wb_{i}, \Ib_{i} \RP$, thus we have
\begin{equation}\label{eq:A_i_clf}
    \begin{aligned}
        \Acal\LP \wb_{i}, \mu_{i}, \sg_{i} \RP &= \underset{b_{i} \sim Q_{\mu_{i}, \sg_{i}^{2}}}{\EE} \LB \underset{\vb_{i} \sim \Ncal \LP \wb_{i}, \Ib_{i} \RP}{\EE} \mathds{1}\LB \Bigl( \langle \vb_{i}, \bpsi \rangle + y b_{i} \Bigr) \leq 0 \RB \RB \\ 
        &= \underset{b_{i} \sim Q_{\mu_{i}, \sg_{i}^{2}}}{\EE} \LB \int_{\mathds{R}^{d}} \LP \frac{1}{\sqrt{2\pi}} \RP^{d} e^{-\frac{1}{2} \LN \vb_{i} - \wb_{i} \RN^{2}} \mathds{1}\LB \Bigl( \langle \vb_{i}, \bpsi \rangle + y b_{i} \Bigr) \leq 0 \RB d\vb_{i}  \RB \textrm{.} 
    \end{aligned}  
\end{equation}

Now, consider any non-zero vector $\bpsi$. Let's decompose the vector $\vb_{i}$ into its part $v_{i\parallel}$ parallel to $\bpsi$ and its part $\vb_{i\perp}$ perpendicular to $\bpsi$. We then have
\[
    \vb_{i} = \LP v_{i\parallel}, \vb_{i\perp} \RP \textrm{.}
\]
Furthermore, if we define $\psi \eqdef \LN \bpsi \RN = \LN y \xb \RN = \LN \xb \RN$ we obtain the following
\[
    \langle \vb_{i}, \bpsi \rangle = v_{i\parallel} \psi \textrm{.}
\]
Similarly, let's decompose the weight vector $\wb_{i}$ into $\LP w_{i\parallel}, \wb_{i\perp} \RP$ constituted of parallel and perpendicular parts to $\bpsi$. We then get
\[
    \LN \vb_{i} - \wb_{i} \RN^{2} = \LP v_{i\parallel} - w_{i\parallel} \RP^{2} + \LN \vb_{i\perp} - \wb_{i\perp} \RN^{2} \textrm{.}
\]
With all these elements, Equation~\eqref{eq:A_i_clf} becomes 
\[
\begin{aligned}
    \Acal\LP \wb_{i}, \mu_{i}, \sg_{i} \RP &= \underset{b_{i} \sim Q_{\mu_{i}, \sg_{i}^{2}}}{\EE} \Biggl[ \int_{\mathds{R}} \frac{1}{\sqrt{2\pi}} e^{-\frac{1}{2} \LP v_{i\parallel} - w_{i\parallel} \RP^{2}} \mathds{1}\LB \Bigl( v_{i\parallel} \psi + y b_{i} \Bigr) \leq 0 \RB dv_{i\parallel} \\ 
    & \quad \quad \quad \quad \quad \quad \quad \quad \quad \quad \times \int_{\mathds{R}^{d-1}} \LP \frac{1}{ \sqrt{2\pi}} \RP^{d-1} e^{-\frac{1}{2} \LN \vb_{i\perp} - \wb_{i\perp} \RN^{2}} d\vb_{i\perp} \Biggr] \textrm{.} 
\end{aligned}
\]
Since 
\[
    \int_{\mathds{R}^{d-1}} \LP \frac{1}{ \sqrt{2\pi}} \RP^{d-1} e^{-\frac{1}{2} \LN \vb_{i\perp} - \wb_{i\perp} \RN^{2}} d\vb_{i\perp} = 1 \textrm{,}
\]
we obtain
\begin{align}
    \Acal\LP \wb_{i}, \mu_{i}, \sg_{i} \RP &= \underset{b_{i} \sim Q_{\mu_{i}, \sg_{i}^{2}}}{\EE} \Biggl[ \int_{\mathds{R}} \frac{1}{\sqrt{2\pi}} e^{-\frac{1}{2} \LP v_{i\parallel} - w_{i\parallel} \RP^{2}} \mathds{1}\LB \Bigl( v_{i\parallel} \psi + y b_{i} \Bigr) \leq 0 \RB dv_{i\parallel} \Biggr] \label{eq:Expectation_A_i_clf_over_v_i_explication_1}\\ 
    &= \underset{b_{i} \sim Q_{\mu_{i}, \sg_{i}^{2}}}{\EE} \Biggl[ \int_{\mathds{R}} \frac{1}{\sqrt{2\pi}} e^{-\frac{1}{2} t^{2}} \mathds{1}\LB \Bigl\{ (t + w_{i\parallel}) \psi + y b_{i} \Bigr\} \leq 0 \RB dt \Biggr] \label{eq:Expectation_A_i_clf_over_v_i_explication_2}\\ 
    &= \underset{b_{i} \sim Q_{\mu_{i}, \sg_{i}^{2}}}{\EE} \Biggl[ \underset{ t \sim \Ncal\LP 0, 1 \RP }{\EE} \mathds{1}\LB \Bigl( \psi t + \langle \wb_{i}, \bpsi \rangle + y b_{i} \Bigr) \leq 0 \RB \Biggr] \label{eq:Expectation_A_i_clf_over_v_i_explication_3}\\ 
    &= \underset{ t \sim \Ncal\LP 0, 1 \RP }{\EE} \Biggl[ \underset{b_{i} \sim Q_{\mu_{i}, \sg_{i}^{2}}}{\EE} \mathds{1}\LB \Bigl( \psi t + \langle \wb_{i}, \bpsi \rangle + y b_{i} \Bigr) \leq 0 \RB \Biggr] \label{eq:Expectation_A_i_clf_over_b_i_explication_1}\\ 
    &= \underset{ t \sim \Ncal\LP 0, 1 \RP }{\EE} \Biggl[ \underset{b_{i} \sim \Ncal\LP \mu_{i}, \sg_{i}^{2} \RP}{\EE} \mathds{1}\Bigl[ y b_{i} \leq - \psi t - \langle \wb_{i}, \bpsi \rangle \Bigr] \Biggr] \label{eq:Expectation_A_i_clf_over_b_i_explication_2} \textrm{.} 
\end{align}
Here, Equation~\eqref{eq:Expectation_A_i_clf_over_v_i_explication_2} is a consequence of the change of variable $t \eqdef v_{i\parallel} - w_{i\parallel}$, and~\eqref{eq:Expectation_A_i_clf_over_v_i_explication_3} is derived by replacing $w_{i\parallel}$ by $\frac{\langle \wb_{i}, \bpsi \rangle}{\psi}$. Equation~\eqref{eq:Expectation_A_i_clf_over_b_i_explication_1} results from the change in the order of expectation, and~\eqref{eq:Expectation_A_i_clf_over_b_i_explication_2} follows from the definition of $Q_{\mu_{i}, \sg_{i}^{2}} \eqdef \Ncal\LP \mu_{i}, \sg_{i}^{2} \RP$.

Taking advantage of the fact that $y \in \LC -1, +1 \RC$, Equation~\eqref{eq:Expectation_A_i_clf_over_b_i_explication_2} can be reformulated as
\begin{equation}\label{eq:A_i_clf_end}
    \begin{aligned}
        \Acal\LP \wb_{i}, \mu_{i}, \sg_{i} \RP &= \underset{t \sim \Ncal\LP 0, 1 \RP}{\EE} \Biggl[ \underset{yb_{i} \sim \Ncal\LP y \mu_{i}, \sg_{i}^{2} \RP}{\EE} \mathds{1}\Bigl[ y b_{i} \leq - \psi t - \langle \wb_{i}, \bpsi \rangle \Bigr] \Biggr] \\ 
        &= \underset{t \sim \Ncal\LP 0, 1 \RP}{\EE} \Biggl[ \Phi\LP \frac{ - \psi t - \langle \wb_{i}, \bpsi \rangle - y \mu_{i} }{ \sg_{i} } \RP \Biggr] \\ 
        &= \underset{t \sim \Ncal\LP 0, 1 \RP}{\EE} \Biggl[ \Phi\LP \frac{ -\LN \xb \RN t - y \LP \langle \wb_{i}, \xb \rangle + \mu_{i} \RP }{ \sg_{i} } \RP \Biggr] \\ 
        &= \Phi\LP \frac{ - y \LP \langle \wb_{i}, \xb \rangle + \mu_{i} \RP}{\sqrt{ \sg_{i}^{2} + \LN \xb \RN^{2} }} \RP\\ 
        &= 1 - \Phi\LP \frac{y \LP \langle \wb_{i}, \xb \rangle + \mu_{i} \RP}{\sqrt{ \sg_{i}^{2} + \LN \xb \RN^{2} }} \RP \textrm{,} 
    \end{aligned}
\end{equation}
where $\Phi$ is the cumulative distribution function (CDF) of the standard normal distribution.

Similarly, we obtain 
\begin{equation}\label{eq:A_ext_clf}
    \Acal\LP \wb_{ext}, \mu_{ext}, \sg_{ext} \RP = 1 - \Phi\LP \frac{y \LP \langle \wb_{ext}, \xb \rangle + \mu_{ext} \RP}{\sqrt{ \sg_{ext}^{2} + \LN \xb \RN^{2} }} \RP\; \textrm{.}
\end{equation}
Plugging~\eqref{eq:A_i_clf_end} and \eqref{eq:A_ext_clf} into~\eqref{eq:perte_Q_exp_generale_clf_gn_proof_3} finally gives us 
\[
    \begin{aligned}
        \ell\Bigl[ Q, (\bpsi, \xb, y) \Bigr] &= \sum_{i=1}^{n} \LB \Theta_{k_{i}, \tau_{i}, d(\cb_{i}, \xb)} \times \LC 1 - \Phi\LP \frac{y (\langle \wb_{i}, \xb \rangle + \mu_{i})}{\sqrt{\sg_{i}^{2} + \LN \xb \RN^{2}}} \RP \RC \RB \\ 
        & \quad \quad \quad + \prod_{i=1}^{n} \LB 1 - \Theta_{k_{i}, \tau_{i}, d(\cb_{i}, \xb)} \RB \times \LC 1 - \Phi\LP \frac{y (\langle \wb_{ext}, \xb \rangle + \mu_{ext})}{\sqrt{\sg_{ext}^{2} + \LN \xb \RN^{2}}} \RP \RC \textrm{.}
    \end{aligned}
\]
which completes the proof.

\section{Proof of the Expression of Loss Function in Section~\ref{Linear_Regression_reg_gn}}
\label{app:Linear_Regression_reg_gn}

The proof of the loss $\ell\Bigl[ Q, (\xb, y) \Bigr]$ of~\eqref{eq:perte_Q_exp_generale_reg_gn} in Section~\ref{Linear_Regression_reg_gn} closely follows the procedure carried out in Appendix~\ref{app:Binary_Linear_clf_gn}. The loss $\ell\Bigl[ Q, (\xb, y) \Bigr]$ of posterior $Q$ on example $(\xb,y)$, by definition, is given by
\begin{equation}\label{eq:perte_Q_exp_generale_reg_gn_proof_1}
    \begin{aligned}
        \ell\Bigl[ Q, (\xb, y) \Bigr] &\eqdef \underset{Q}{\EE} \LC \ell\Bigl[ \Bigl(\bm{c}, \bm{v}, \bm{b}, \bbeta, \vb_{ext}, b_{ext} \Bigr), (\xb, y) \Bigr] \RC \\ 
        &= \underset{Q}{\EE} \LC \sum_{i=1}^{n} \LB \Bigl( \langle \vb_{i}, \xb \rangle + b_{i} - y \Bigr)^{2} K(\cb_{i}, \xb, \be_{i}) \RB \RC \\
        & \quad \quad \quad \quad \quad + \underset{Q}{\EE} \LC \Bigl( \langle \vb_{ext}, \xb \rangle + b_{ext} - y \Bigr)^{2}  \prod_{i=1}^{n} \bar{K}(\cb_{i}, \xb, \be_{i}) \RC \textrm{.} 
    \end{aligned}
\end{equation}

Since variables $\LC \vb_{i}, b_{i}, \be_{i} \RC_{i=1}^{n}$ are all independent, therefore, Equation~\eqref{eq:perte_Q_exp_generale_reg_gn_proof_1} becomes
\begin{equation}\label{eq:perte_Q_exp_generale_reg_gn_proof_2}
    \begin{aligned}
        \ell\Bigl[ Q, (\xb, y) \Bigr] &= \sum_{i=1}^{n} \LB \underset{Q}{\EE} \LC \Bigl( \langle \vb_{i}, \xb \rangle + b_{i} - y \Bigr)^{2} K(\cb_{i}, \xb, \be_{i}) \RC \RB \\
        & \quad \quad \quad \quad \quad + \underset{Q}{\EE} \LC \Bigl( \langle \vb_{ext}, \xb \rangle + b_{ext} - y \Bigr)^{2} \prod_{i=1}^{n} \bar{K}(\cb_{i}, \xb, \be_{i}) \RC \\ 
        &= \sum_{i=1}^{n} \LB \underset{(\vb_{i}, b_{i}) \sim \LC Q_{\wb_{i}, \Ib_{i} \rho_{i}^{2}} \times Q_{\mu_{i}, \sg_{i}^{2}} \RC}{\EE} \Bigl( \langle \vb_{i}, \xb \rangle + b_{i} - y \Bigr)^{2} \times \underset{ \be_{i} \sim Q_{k_{i}, \tau_{i}} }{\EE} K(\cb_{i}, \xb, \be_{i}) \RB \\
        & \quad + \underset{(\vb_{ext}, b_{ext}) \sim \LC Q_{\wb_{ext}, \Ib_{ext} \rho_{ext}^{2}} \times Q_{\mu_{ext}, \sg_{ext}^{2}} \RC}{\EE} \Bigl( \langle \vb_{ext}, \xb \rangle + b_{ext} - y \Bigr)^{2} \\ 
        & \quad \quad \quad \quad \quad \quad \quad \quad \quad \quad \quad \quad \quad \quad \quad \quad \quad \quad \quad \times \prod_{i=1}^{n} \LB \underset{ \be_{i} \sim Q_{k_{i}, \tau_{i}} }{\EE} \bar{K}(\cb_{i}, \xb, \be_{i}) \RB \textrm{.} 
    \end{aligned}
\end{equation}
Recall that from Subsection~\ref{Binary_Linear_clf_gn}, Equation~\eqref{eq:theta}, we already have
\[
    \underset{ \be_{i} \sim Q_{k_{i}, \tau_{i}} }{\EE} K(\cb_{i}, \xb, \be_{i}) = \underset{ \be_{i} \sim \Gm\LP k_{i}, \tau_{i} \RP }{\EE} \mathds{1}\Bigl[ d(\cb_{i}, \xb) \leq \be_{i} \Bigr] \eqdef \Theta_{k_{i}, \tau_{i}, d(\cb_{i}, \xb)} \textrm{.}
\]
Then, Equation~\eqref{eq:perte_Q_exp_generale_reg_gn_proof_2} can be expressed as
\begin{equation}\label{eq:perte_Q_exp_generale_reg_gn_proof_3}
    \begin{aligned}
        \ell\Bigl[ Q, (\xb, y) \Bigr] &= \sum_{i=1}^{n} \LB \Theta_{k_{i}, \tau_{i}, d(\cb_{i}, \xb)} \times \underset{(\vb_{i}, b_{i}) \sim \LC Q_{\wb_{i}, \Ib_{i} \rho_{i}^{2}} \times Q_{\mu_{i}, \sg_{i}^{2}} \RC}{\EE} \Bigl( \langle \vb_{i}, \xb \rangle + b_{i} - y \Bigr)^{2} \RB \\
        & \quad + \prod_{i=1}^{n} \LB 1 - \Theta_{k_{i}, \tau_{i}, d(\cb_{i}, \xb)} \RB \\ 
        & \quad \quad \quad \quad \quad \quad \times \underset{(\vb_{ext}, b_{ext}) \sim \LC Q_{\wb_{ext}, \Ib_{ext} \rho_{ext}^{2}} \times Q_{\mu_{ext}, \sg_{ext}^{2}} \RC}{\EE} \Bigl( \langle \vb_{ext}, \xb \rangle + b_{ext} - y \Bigr)^{2} \\ 
        &= \sum_{i=1}^{n} \LB \Theta_{k_{i}, \tau_{i}, d(\cb_{i}, \xb)} \times \Bcal\LP \wb_{i}, \rho_{i}, \mu_{i}, \sg_{i} \RP \RB \\
        & \quad + \prod_{i=1}^{n} \LB 1 - \Theta_{k_{i}, \tau_{i}, d(\cb_{i}, \xb)} \RB  \times \Bcal\LP \wb_{ext}, \rho_{ext}, \mu_{ext}, \sg_{ext} \RP \textrm{,} 
    \end{aligned}
\end{equation}
where the estimates $\Bcal\LP \wb_{i}, \rho_{i}, \mu_{i}, \sg_{i} \RP$ and $\Bcal\LP \wb_{ext}, \rho_{ext}, \mu_{ext}, \sg_{ext} \RP$ are defined as
\begin{equation}\label{eq:def_expectation_of_loss_func_with_Bcal1}
    \Bcal\LP \wb_{i}, \rho_{i}, \mu_{i}, \sg_{i} \RP \eqdef \underset{(\vb_{i}, b_{i}) \sim \LC Q_{\wb_{i}, \Ib_{i} \rho_{i}^{2}} \times Q_{\mu_{i}, \sg_{i}^{2}} \RC}{\EE} \Bigl( \langle \vb_{i}, \xb \rangle + b_{i} - y \Bigr)^{2} \textrm{,}
\end{equation}
and
\begin{equation}\label{eq:def_expectation_of_loss_func_with_Bcal2}
    \Bcal\LP \wb_{ext}, \rho_{ext}, \mu_{ext}, \sg_{ext} \RP \eqdef \underset{(\vb_{ext}, b_{ext}) \sim \LC Q_{\wb_{ext}, \Ib_{ext} \rho_{ext}^{2}} \times Q_{\mu_{ext}, \sg_{ext}^{2}} \RC}{\EE} \Bigl( \langle \vb_{ext}, \xb \rangle + b_{ext} - y \Bigr)^{2} \textrm{.}
\end{equation}
Beyond the index notations $i$ and $ext$, the expressions of~\eqref{eq:def_expectation_of_loss_func_with_Bcal1} and~\eqref{eq:def_expectation_of_loss_func_with_Bcal2} are identical. In the following, we will find the analytical expression of~\eqref{eq:def_expectation_of_loss_func_with_Bcal1} and deduce that of~\eqref{eq:def_expectation_of_loss_func_with_Bcal2}.
Let's first rewrite ~\eqref{eq:def_expectation_of_loss_func_with_Bcal1} as follows
\begin{equation}\label{eq:B_i_reg}
    \begin{aligned}
        \Bcal\LP \wb_{i}, \rho_{i}, \mu_{i}, \sg_{i} \RP &\eqdef \underset{(\vb_{i}, b_{i}) \sim \LC Q_{\wb_{i}, \Ib_{i} \rho_{i}^{2}} \times Q_{\mu_{i}, \sg_{i}^{2}} \RC}{\EE} \Bigl( \langle \vb_{i}, \xb \rangle + b_{i} - y \Bigr)^{2} \\ 
        &= \underset{b_{i} \sim Q_{\mu_{i}, \sg_{i}^{2}}}{\EE} \LB \underset{\vb_{i} \sim Q_{\wb_{i}, \Ib_{i} \rho_{i}^{2}}}{\EE} \Bigl( \langle \vb_{i}, \xb \rangle + b_{i} - y \Bigr)^{2} \RB \\ 
        &= \underset{b_{i} \sim Q_{\mu_{i}, \sg_{i}^{2}}}{\EE} \LB \int_{\mathds{R}^{d}} \LP \frac{1}{\rho_{i} \sqrt{2\pi}} \RP^{d} e^{-\frac{1}{2 \rho_{i}^{2}} \LN \vb_{i} - \wb_{i} \RN^{2}} \Bigl( \langle \vb_{i}, \xb \rangle + b_{i} - y \Bigr)^{2} d\vb_{i} \RB \textrm{.} 
    \end{aligned}  
\end{equation}

As in previous Appendix, let's now consider any non-zero vector $\xb$ and decompose the vector $\vb_{i}$ into $\LP v_{i\parallel}, \vb_{i\perp} \RP$ constituted of parallel and perpendicular parts to $\xb$. We then have
\[
    \vb_{i} = \LP v_{i\parallel}, \vb_{i\perp} \RP \textrm{.}
\]
Besides, if we define $\phi \eqdef \LN \xb \RN$ we obtain the following
\[
    \langle \vb_{i}, \xb \rangle = v_{i\parallel} \phi \textrm{.}
\]
Similarly, let's decompose the weight vector $\wb_{i}$ into $\LP w_{i\parallel}, \wb_{i\perp} \RP$ constituted of parallel and perpendicular parts to $\xb$. We then get
\[
    \LN \vb_{i} - \wb_{i} \RN^{2} = \LP v_{i\parallel} - w_{i\parallel} \RP^{2} + \LN \vb_{i\perp} - \wb_{i\perp} \RN^{2} \textrm{.}
\]
With all these elements, Equation~\eqref{eq:B_i_reg} becomes 
\[
\begin{aligned}
    \Bcal\LP \wb_{i}, \rho_{i}, \mu_{i}, \sg_{i} \RP &= \underset{b_{i} \sim Q_{\mu_{i}, \sg_{i}^{2}}}{\EE} \Biggl[ \int_{\mathds{R}} \frac{1}{\rho_{i} \sqrt{2\pi}} e^{-\frac{1}{2 \rho_{i}^{2}} \LP v_{i\parallel} - w_{i\parallel} \RP^{2}} \LP v_{i\parallel} \phi + b_{i} - y \RP^{2} dv_{i\parallel} \\ 
    & \quad \quad \quad \quad \quad \quad \quad \times \int_{\mathds{R}^{d-1}} \LP \frac{1}{\rho_{i} \sqrt{2\pi}} \RP^{d-1} e^{-\frac{1}{2 \rho_{i}^{2}} \LN \vb_{i\perp} - \wb_{i\perp} \RN^{2}} d\vb_{i\perp} \Biggr] \\ 
    &= \underset{b_{i} \sim Q_{\mu_{i}, \sg_{i}^{2}}}{\EE} \Biggl[ \int_{\mathds{R}} \frac{1}{\rho_{i} \sqrt{2\pi}} e^{-\frac{1}{2 \rho_{i}^{2}} \LP v_{i\parallel} - w_{i\parallel} \RP^{2}} \LP v_{i\parallel} \phi + b_{i} - y \RP^{2} dv_{i\parallel} \Biggr]
    \\ 
    &= \underset{b_{i} \sim Q_{\mu_{i}, \sg_{i}^{2}}}{\EE} \Biggl[ \int_{\mathds{R}} \frac{1}{\rho_{i} \sqrt{2\pi}} e^{-\frac{1}{2 \rho_{i}^{2}} t^{2}} \LC (t + w_{i\parallel}) \phi + b_{i} - y \RC^{2} dt \Biggr]\textrm{,} 
\end{aligned}
\]
where $t \eqdef v_{i\parallel} - w_{i\parallel}$. Then, with $w_{i\parallel} = \frac{\langle \wb_{i}, \xb \rangle}{\phi}$ we obtain
\[
    \begin{aligned}
        \Bcal\LP \wb_{i}, \rho_{i}, \mu_{i}, \sg_{i} \RP &= \underset{b_{i} \sim Q_{\mu_{i}, \sg_{i}^{2}}}{\EE} \Biggl[ \underset{ t \sim \Ncal\LP 0, \rho_{i}^{2} \RP }{\EE} \Bigl( \phi t + \langle \wb_{i}, \xb \rangle + b_{i} - y \Bigr)^{2} \Biggr] \\ 
        &= \phi^{2} \underset{b_{i} \sim Q_{\mu_{i}, \sg_{i}^{2}}}{\EE} \Biggl[ \underset{t \sim \Ncal\LP 0, \rho_{i}^{2} \RP}{\EE} \LP t + \frac{\langle \wb_{i}, \xb \rangle + b_{i} - y}{\phi} \RP^{2} \Biggr] \\ 
        &= \phi^{2} \underset{b_{i} \sim Q_{\mu_{i}, \sg_{i}^{2}}}{\EE} \Biggl[  \underset{t \sim \Ncal\LP 0, \rho_{i}^{2} \RP}{\EE} \LB t^{2} \RB + \frac{2\Bigl( \langle \wb_{i}, \xb \rangle + b_{i} - y \Bigr)}{\phi} \underset{t \sim \Ncal\LP 0, \rho_{i}^{2} \RP}{\EE} \LB t \RB \\ & \quad \quad \quad \quad \quad \quad \quad \quad \quad \quad \quad \quad \quad \quad \quad \quad \quad \quad \quad \quad + \frac{\Bigl( \langle \wb_{i}, \xb \rangle + b_{i} - y \Bigr)^{2}}{\phi^{2}} \Biggr] \\ 
        &= \phi^{2} \underset{b_{i} \sim Q_{\mu_{i}, \sg_{i}^{2}}}{\EE} \Biggl[ \rho_{i}^{2} + 0 + \frac{\Bigl( \langle \wb_{i}, \xb \rangle + b_{i} - y \Bigr)^{2}}{\phi^{2}} \Biggr] \\ 
        &= \phi^{2} \rho_{i}^{2} + \underset{b_{i} \sim Q_{\mu_{i}, \sg_{i}^{2}}}{\EE} \Bigl( \langle \wb_{i}, \xb \rangle + b_{i} - y \Bigr)^{2} \\ 
        &= \phi^{2} \rho_{i}^{2} + \underset{b_{i} \sim \Ncal\LP \mu_{i}, \sg_{i}^{2} \RP}{\EE} \Bigl( \langle \wb_{i}, \xb \rangle + b_{i} - y \Bigr)^{2} \\ 
        &= \phi^{2} \rho_{i}^{2} + \underset{s \sim \Ncal\LP 0, \sg_{i}^{2} \RP}{\EE} \Bigl( s + \langle \wb_{i}, \xb \rangle + \mu_{i} - y \Bigr)^{2} \textrm{,} 
    \end{aligned}
\]
where $s \eqdef b_{i} - \mu_{i}$. Then, we have
\begin{equation}\label{eq:B_i_reg_end}
    \begin{aligned}
        \Bcal\LP \wb_{i}, \rho_{i}, \mu_{i}, \sg_{i} \RP &= \phi^{2} \rho_{i}^{2} + \underset{s \sim \Ncal\LP 0, \sg_{i}^{2} \RP}{\EE} \LB s^{2} \RB + 2\Bigl( \langle \wb_{i}, \xb \rangle + \mu_{i} - y \Bigr) \underset{s \sim \Ncal\LP 0, \sg_{i}^{2} \RP}{\EE} \LB s \RB \\ & \quad \quad \quad \quad \quad \quad \quad \quad \quad \quad \quad \quad \quad \quad \quad \quad \quad \quad \quad \quad + \Bigl( \langle \wb_{i}, \xb \rangle + \mu_{i} - y \Bigr)^{2} \\ 
        &= \phi^{2} \rho_{i}^{2} + \sg_{i}^{2} + 0 + \Bigl( \langle \wb_{i}, \xb \rangle + \mu_{i} - y \Bigr)^{2} \\ 
        &= \LN \xb \RN^{2} \rho_{i}^{2} + \sg_{i}^{2} + \Bigl( \langle \wb_{i}, \xb \rangle + \mu_{i} - y \Bigr)^{2} \textrm{.} 
    \end{aligned}
\end{equation}
Similarly, we obtain
\begin{equation}\label{eq:B_ext_reg}
    \Bcal\LP \wb_{ext}, \rho_{ext}, \mu_{ext}, \sg_{ext} \RP = \LN \xb \RN^{2} \rho_{ext}^{2} + \sg_{ext}^{2} + \Bigl( \langle \wb_{ext}, \xb \rangle + \mu_{ext} - y \Bigr)^{2} \textrm{.}
\end{equation}
Combining Equations~\eqref{eq:B_i_reg_end} and \eqref{eq:B_ext_reg}, Equation~\eqref{eq:perte_Q_exp_generale_reg_gn_proof_3} then becomes
\[
    \begin{aligned}
        \ell\Bigl[ Q, (\xb, y) \Bigr] &= \sum_{i=1}^{n} \LB \Theta_{k_{i}, \tau_{i}, d(\cb_{i}, \xb)} \times \LC \LN \xb \RN^{2} \rho_{i}^{2} + \sg_{i}^{2} + \Bigl( \langle \wb_{i}, \xb \rangle + \mu_{i} - y \Bigr)^{2}  \RC \RB \\ 
        & \quad \quad \quad + \prod_{i=1}^{n} \LB 1 - \Theta_{k_{i}, \tau_{i}, d(\cb_{i}, \xb)} \RB \\ 
        & \quad \quad \quad \quad \quad \quad \quad \quad \times \LC \LN \xb \RN^{2} \rho_{ext}^{2} + \sg_{ext}^{2} + \Bigl( \langle \wb_{ext}, \xb \rangle + \mu_{ext} - y \Bigr)^{2} \RC \textrm{,}
    \end{aligned}
\]
which completes the proof.

\textbf{Remark.} The loss $\ell\Bigl[ Q, (\xb, y) \Bigr]$ given by~\eqref{eq:perte_Q_exp_generale_reg_gn_search} in Section~\ref{Linear_Regression_reg_search_gn}, is obtained by following exactly the same steps as here. In this scenario, only the probability of belonging to the locality $\be_{i}$ is affected. Thus, instead of having $\Theta_{k_{i}, \tau_{i}, d(\cb_{i}, \xb)}$ we obtain $\Up_{d, \LN \cb_{i_{0}} - \xb \RN, \varep_{i}, k_{i}, \tau_{i}}$. 

\newpage
\section{ Quasi-Monte Carlo Approximation's of Equation~\ref{eq:upsilon_def}}
\label{app:Binary_Linear_clf_gn_search}

In this appendix, we show how we use the Quasi-Monte Carlo method to efficiently calculate $\Up_{d, \LN \cb_{i_{0}} - \xb \RN, \varep_{i}, k_{i}, \tau_{i}}$ from~\eqref{eq:upsilon_def} in Subsection~\ref{Binary_Linear_clf_gn_search}. Recall that $\Up_{d, \LN \cb_{i_{0}} - \xb \RN, \varep_{i}, k_{i}, \tau_{i}}$ represents the probability that the instance $\xb$ will be located within a ball centered at the point $\cb_{i_{0}}$ with some variance $\varep_{i}^{2}$ and the unknown ball radius is expressed in terms of $\varep_{i}$, $k_{i}$ and $\tau_{i}$. The expression of $\Up_{d, \LN \cb_{i_{0}} - \xb \RN, \varep_{i}, k_{i}, \tau_{i}}$ given by~\eqref{eq:upsilon_def} is as follows
\[
    \Up_{d, \LN \cb_{i_{0}} - \xb \RN, \varep_{i}, k_{i}, \tau_{i}} \eqdef \underset{ \be_{i} \sim Q_{k_{i}, \tau_{i}} }{\EE} \LB P \LP \frac{\be_{i}^{2}}{\varep_{i}^{2}}; d, \frac{\LN \cb_{i_{0}} - \xb \RN^{2}}{\varep_{i}^{2}} \RP \RB \textrm{,}
\]
where $P \LP \frac{\be_{i}^{2}}{\varep_{i}^{2}}; d, \frac{\LN \cb_{i_{0}} - \xb \RN^{2}}{\varep_{i}^{2}} \RP$ is the CDF of the noncentral chi-squared distribution $\chi^{2}$ with $d$ degrees of freedom, and non-centrality parameter $\frac{\LN \cb_{i_{0}} - \xb \RN^{2}}{\varep_{i}^{2}}$.

The noncentral $\chi^{2}$ is a well-known distribution and easy to compute. So, to numerically calculate $\Up_{d, \LN \cb_{i_{0}} - \xb \RN, \varep_{i}, k_{i}, \tau_{i}}$, we just need to randomly draw some sample $\beta_{i}$ from the $\Gamma\LP k_{i}, \tau_{i} \RP$ distribution and compute the mean of $P \LP \frac{\be_{i}^{2}}{\varep_{i}^{2}}; d, \frac{\LN \cb_{i_{0}} - \xb \RN^{2}}{\varep_{i}^{2}} \RP$. Note that the process of randomly drawing $\beta_{i} \sim \Gamma\LP k_{i}, \tau_{i} \RP$ is realized by Quasi-Monte Carlo package in python. Thus, the main question that arises here is how many samples of $\beta_{i}$ randomly drawn from the $\Gamma\LP k_{i}, \tau_{i} \RP$ distribution would be sufficient to obtain an accurate approximation to $\Up_{d, \LN \cb_{i_{0}} - \xb \RN, \varep_{i}, k_{i}, \tau_{i}}$?

To answer this question, let's examine the value of $\Up_{d, \LN \cb_{i_{0}} - \xb \RN, \varep_{i}, k_{i}, \tau_{i}}$ for each sample size of $\beta_{i}$ belonging to the set $\{10, 20, 30, 40, 50, 60, 70, 80, 90, 100\}$, while giving some numeric values to the parameters $\LP d, \LN \cb_{i_{0}} - \xb \RN, \varep_{i}, k_{i}, \tau_{i} \RP$. 

\begin{figure}[H]
    \begin{subfigure}[b]{0.32\textwidth}
        \includegraphics[scale=0.8]{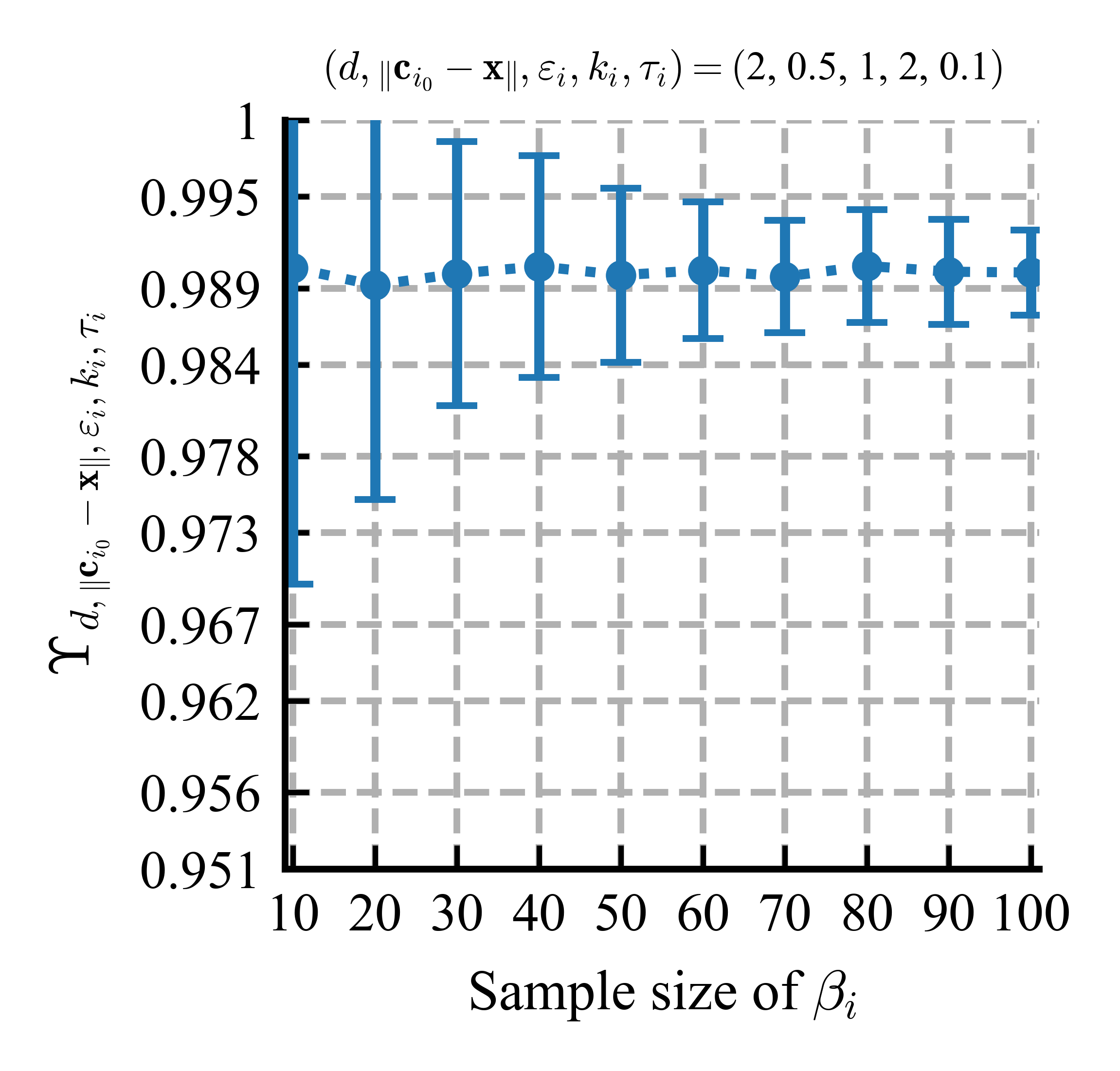}
    \end{subfigure}
    \hfill
    \begin{subfigure}[b]{0.32\textwidth}
        \includegraphics[scale=0.8]{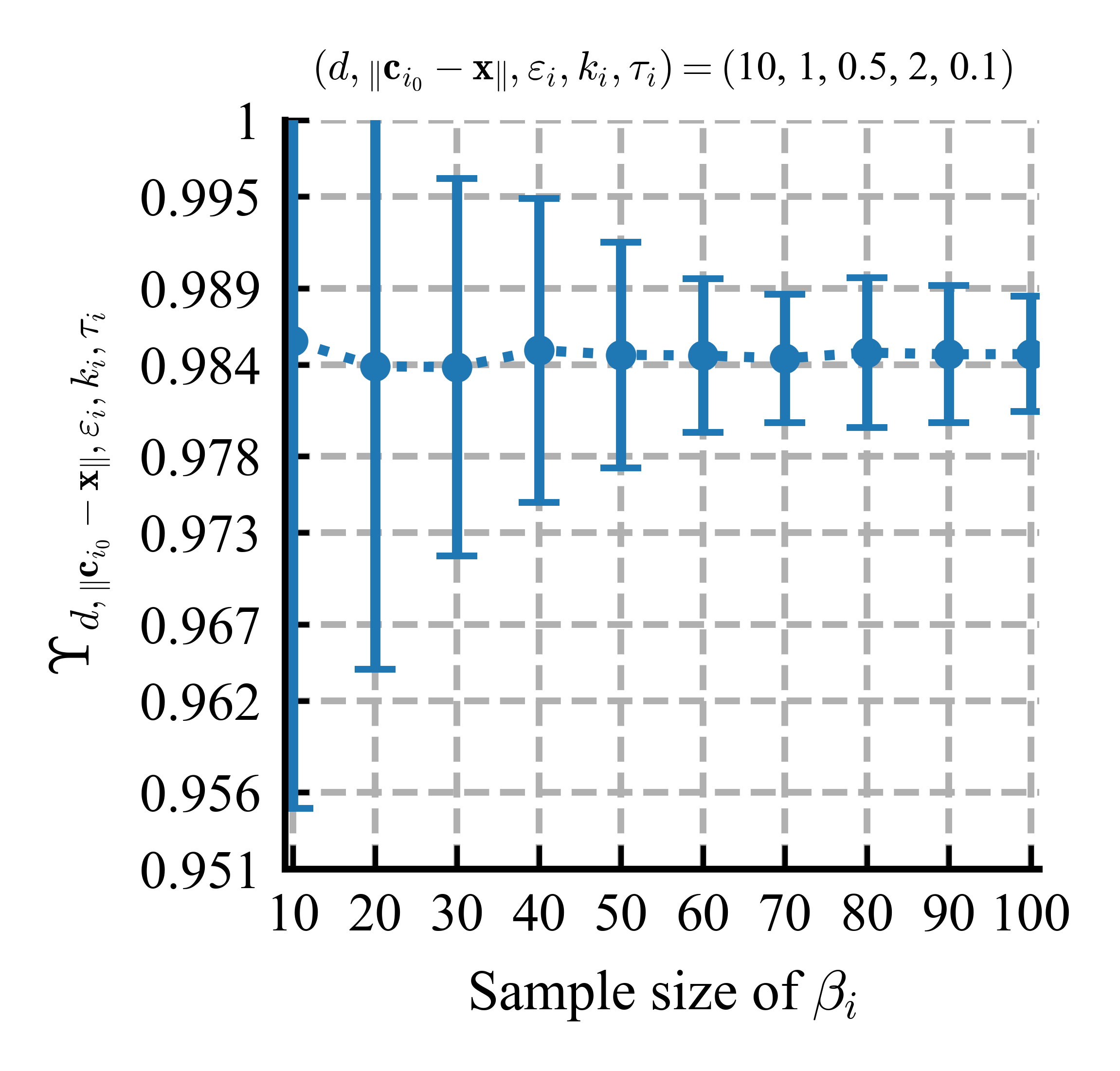}
    \end{subfigure}
    \hfill
    \begin{subfigure}[b]{0.32\textwidth}
        \includegraphics[scale=0.8]{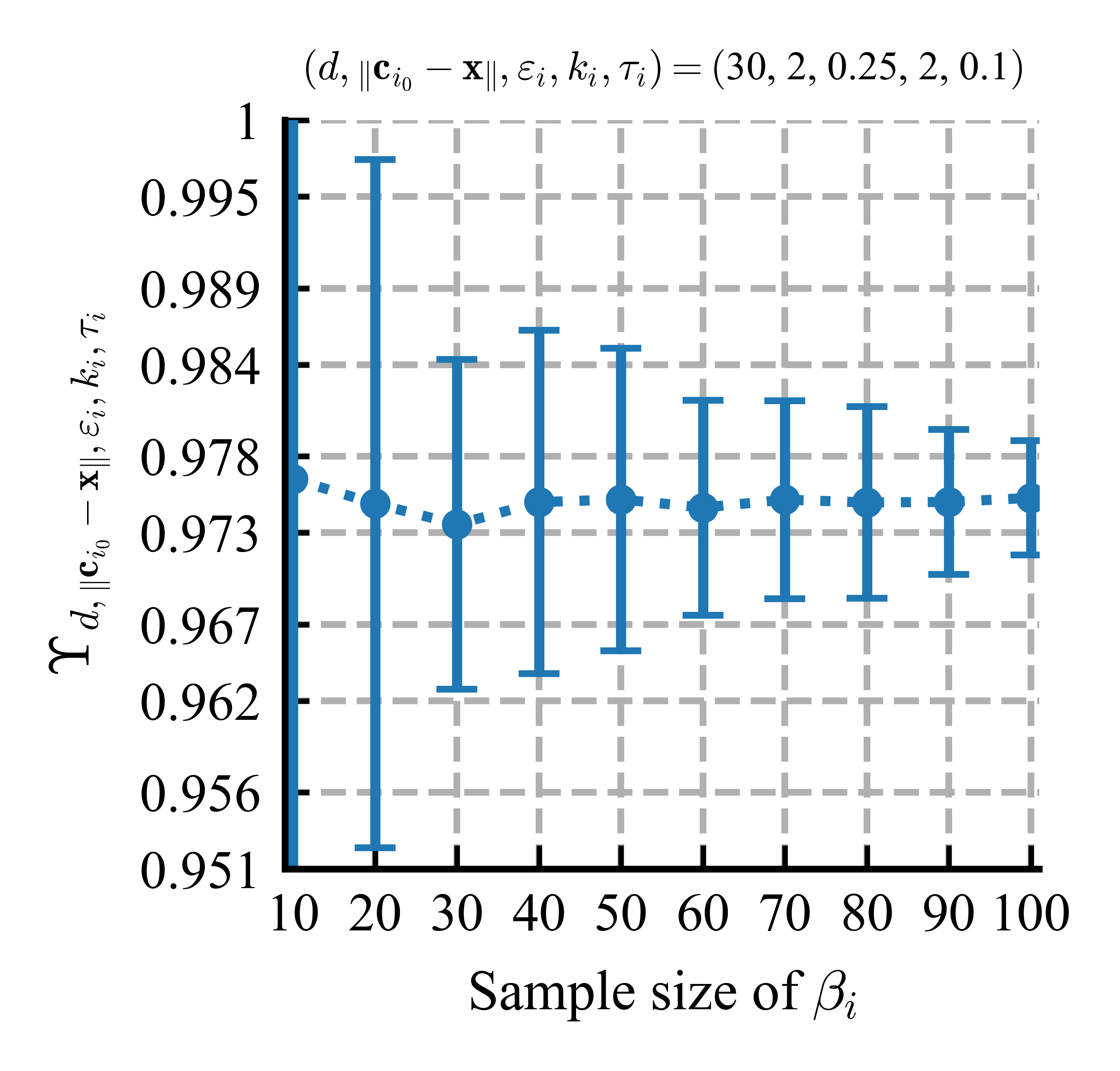}
    \end{subfigure}
    \begin{subfigure}[b]{0.32\textwidth}
        \includegraphics[scale=0.8]{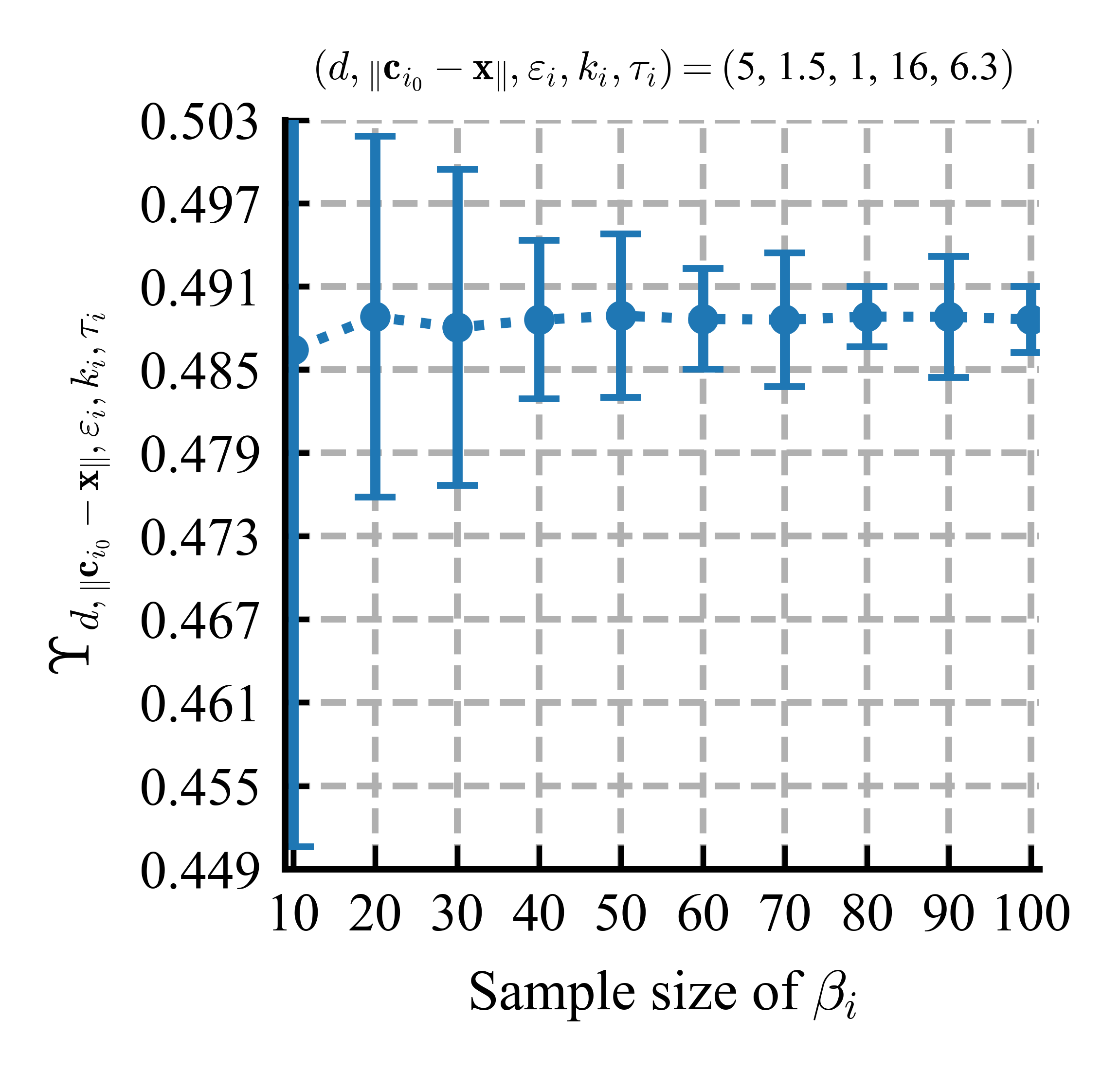}
    \end{subfigure}
    \hfill
    \begin{subfigure}[b]{0.32\textwidth}
        \includegraphics[scale=0.8]{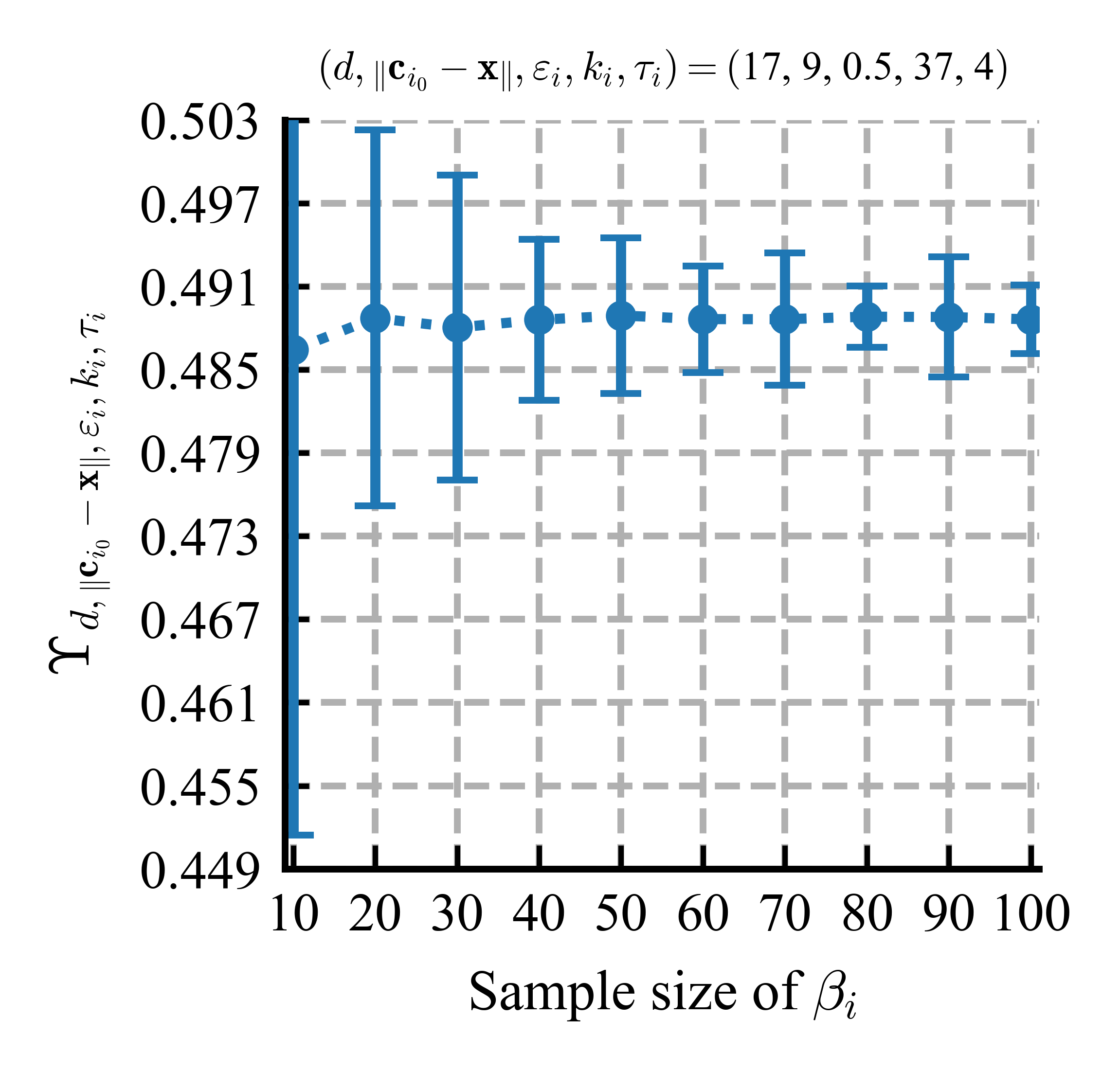}
    \end{subfigure}
    \hfill
    \begin{subfigure}[b]{0.32\textwidth}
        \includegraphics[scale=0.8]{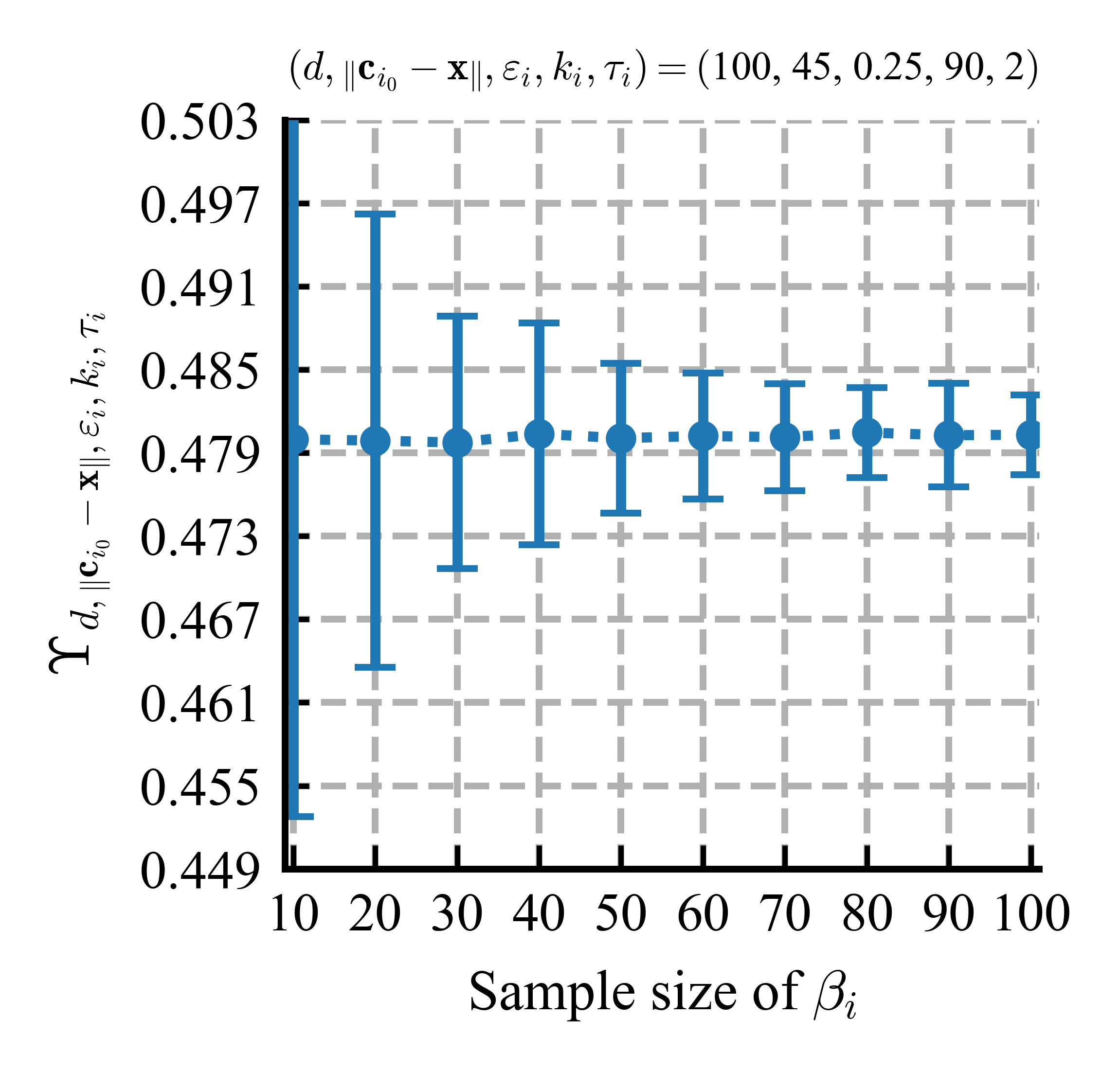}
    \end{subfigure}
    \begin{subfigure}[b]{0.32\textwidth}
        \includegraphics[scale=0.8]{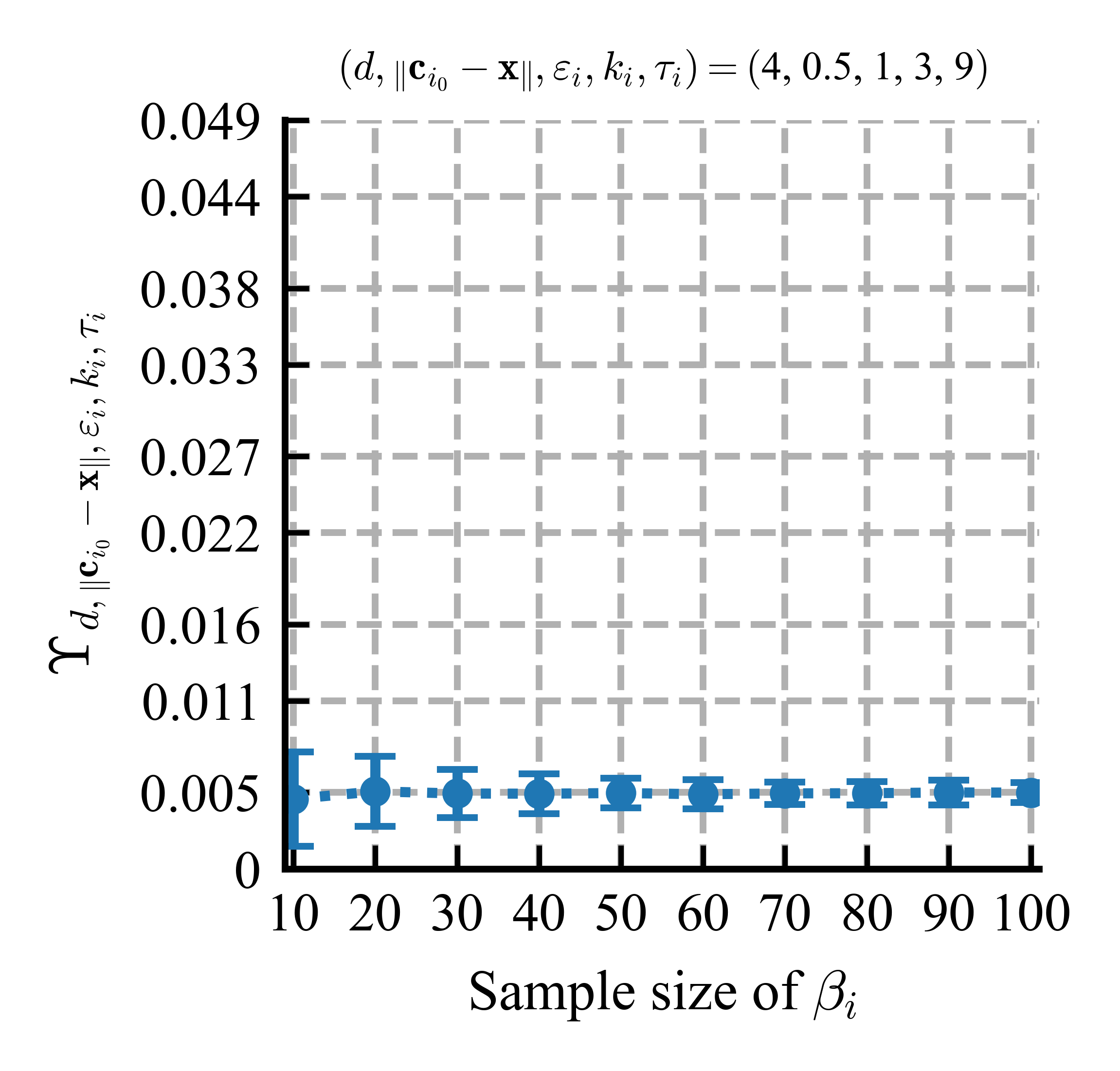}
    \end{subfigure}
    \hfill
    \begin{subfigure}[b]{0.32\textwidth}
        \includegraphics[scale=0.8]{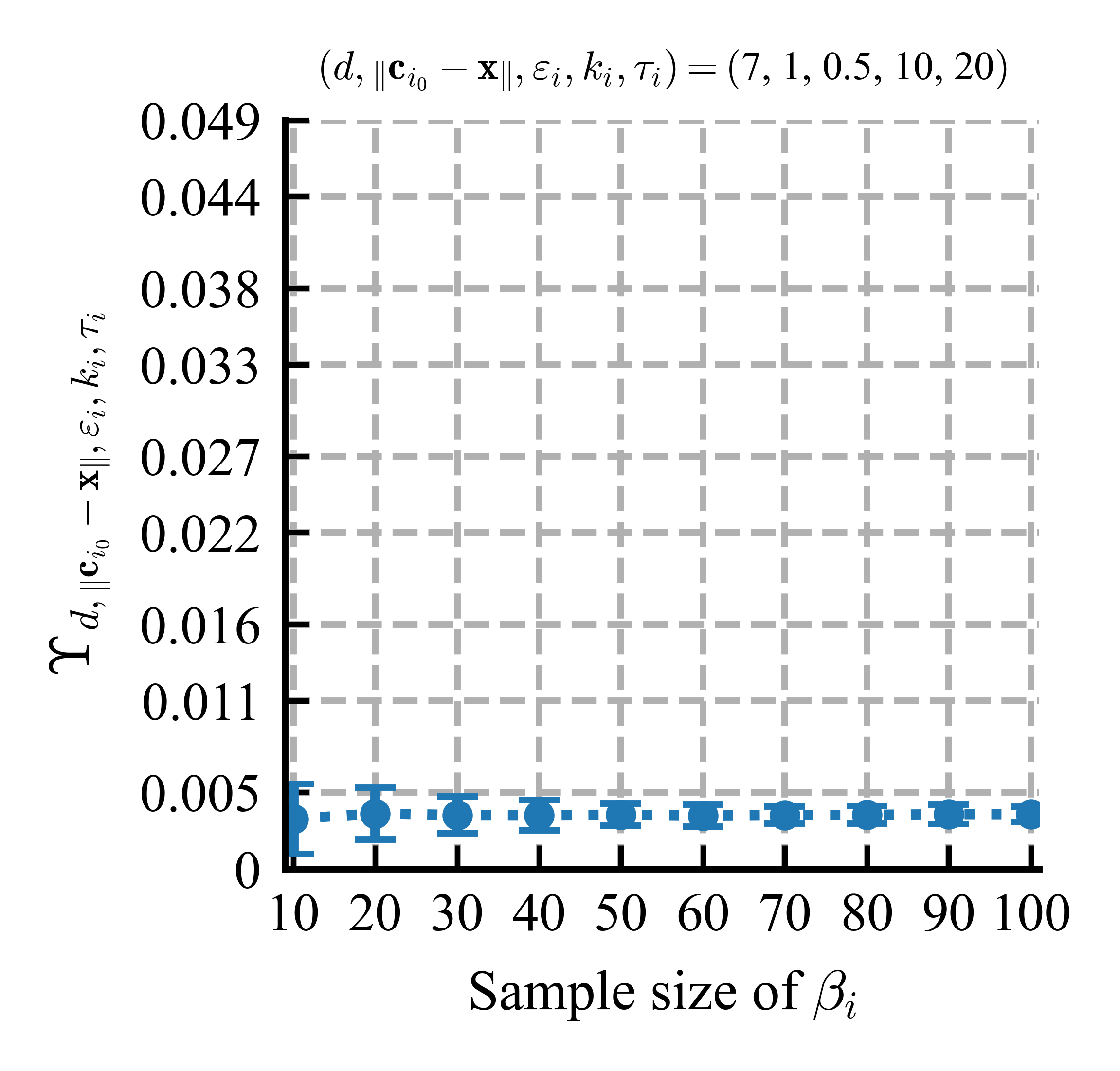}
    \end{subfigure}
    \hfill
    \begin{subfigure}[b]{0.32\textwidth}
        \includegraphics[scale=0.8]{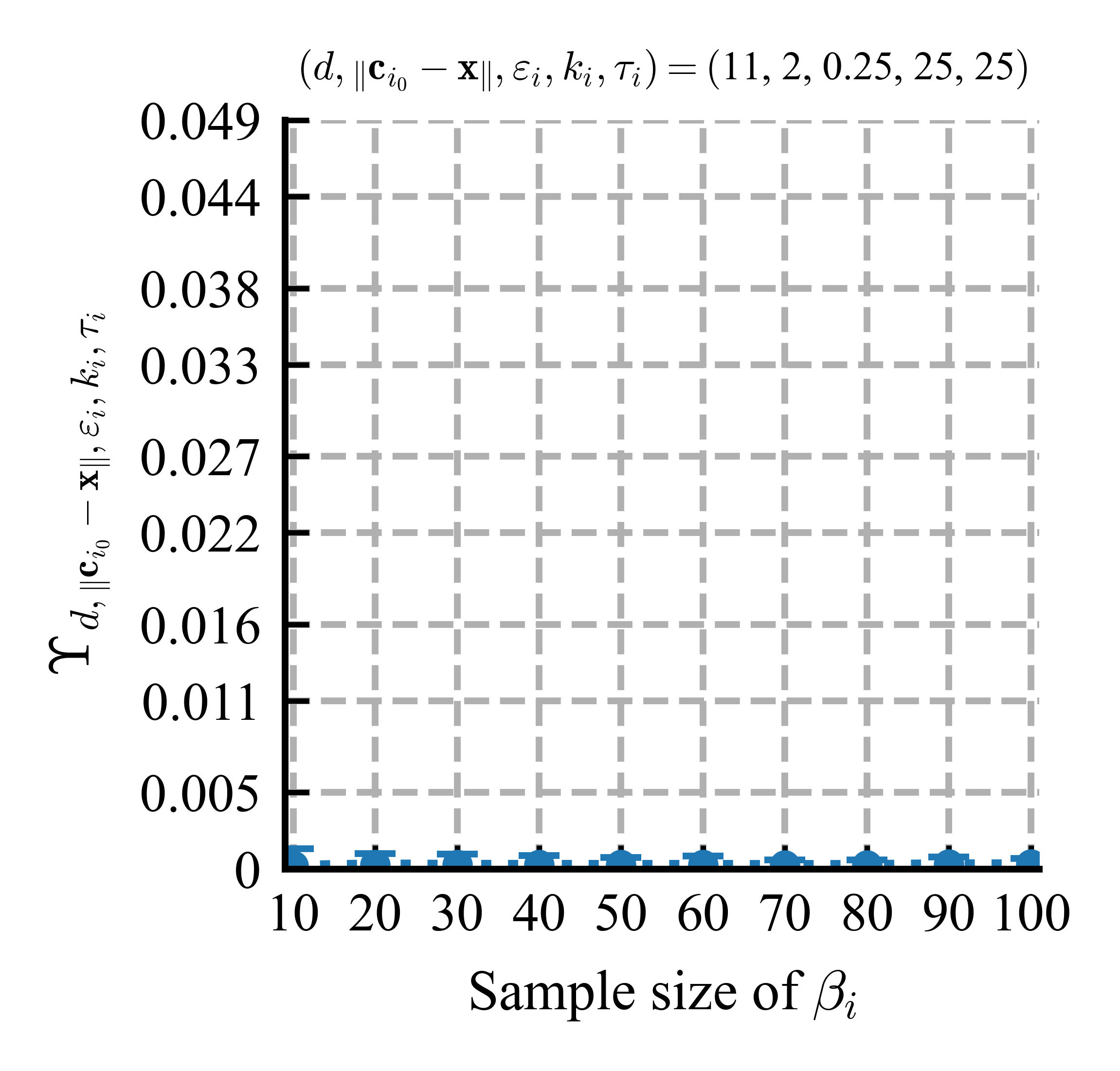}
    \end{subfigure}
    \caption{Mean value of $\Up_{d, \LN \cb_{i_{0}} - \xb \RN, \varep_{i}, k_{i}, \tau_{i}}$ with its error bar over $500$ randomly draw $\beta_{i} \sim \Gamma\LP k_{i}, \tau_{i} \RP$.}
    \label{fig:Quasi_Monte_Carlo_method_1}
\end{figure}

Figure \ref{fig:Quasi_Monte_Carlo_method_1} shows nine subfigures on which we have the mean value of $\Up_{d, \LN \cb_{i_{0}} - \xb \RN, \varep_{i}, k_{i}, \tau_{i}}$ with its error bar over $500$ randomly draws $\beta_{i} \sim \Gamma\LP k_{i}, \tau_{i} \RP$. It shows us that in general, with only $60$ $\beta_{i}$ drawn from the $\Gamma\LP k_{i}, \tau_{i} \RP$ distribution, we are able to achieve an accurate approximation to $\Up_{d, \LN \cb_{i_{0}} - \xb \RN, \varep_{i}, k_{i}, \tau_{i}}$.
 
\vskip 0.2in
\bibliography{sample}

\end{document}